\documentclass[final]{cvpr}

\usepackage{times}
\usepackage{epsfig}
\usepackage{graphicx}
\usepackage{amsmath}
\usepackage{amssymb}
\usepackage{import}
\usepackage[ruled,vlined]{algorithm2e}
\usepackage{color}
\usepackage{xcolor}
\usepackage{subfigure}
\usepackage{afterpage}
\usepackage{longtable}

\usepackage{booktabs} 
\usepackage{tabularx}

\usepackage[pagebackref=true,breaklinks=true,colorlinks,bookmarks=false]{hyperref}

\begin{document}

\title{IIRC: Incremental Implicitly-Refined Classification}

\author{%
  Mohamed Abdelsalam$^{1,2}$, Mojtaba Faramarzi$^{1,2}$, Shagun Sodhani$^{3}$, Sarath Chandar$^{1,4,5}$\\~\\
  $^1$\small{Mila - Quebec AI Institute,  $^2$University of Montreal, $^3$Facebook AI Research,}\\
  $^4$\small{\'Ecole Polytechnique de Montréal, $^5$Canada CIFAR AI Chair}\\
   \small{\texttt{\{abdelsam, faramarm, sarath.chandar\}@mila.quebec, sshagunsodhani@gmail.com}}
}

\maketitle

\begin{abstract}
We introduce the ``Incremental Implicitly-Refined Classification (IIRC)'' setup, an extension to the class incremental learning setup where the incoming batches of classes have two granularity levels. i.e., each sample could have a high-level (coarse) label like ``bear'' and a low-level (fine) label like ``polar bear''. Only one label is provided at a time, and the model has to figure out the other label if it has already learned it. This setup is more aligned with real-life scenarios, where a learner usually interacts with the same family of entities multiple times, discovers more granularity about them, while still trying not to forget previous knowledge. Moreover, this setup enables evaluating models for some important lifelong learning challenges that cannot be easily addressed under the existing setups. These challenges can be motivated by the example ''if a model was trained on the class \textit{bear} in one task and on \textit{polar bear} in another task, will it forget the concept of \textit{bear}, will it rightfully infer that a \textit{polar bear} is still a \textit{bear}? and will it wrongfully associate the label of \textit{polar bear} to other breeds of \textit{bear}?''. We develop a standardized benchmark that enables evaluating models on the IIRC setup. We evaluate several state-of-the-art lifelong learning algorithms and highlight their strengths and limitations. For example, distillation-based methods perform relatively well but are prone to incorrectly predicting too many labels per image. We hope that the proposed setup, along with the benchmark, would provide a meaningful problem setting to the practitioners.
\end{abstract}

\section{Introduction}
\label{sec:introduction}

\begin{figure}[tb]
  \centering\includegraphics[width=0.98\linewidth]{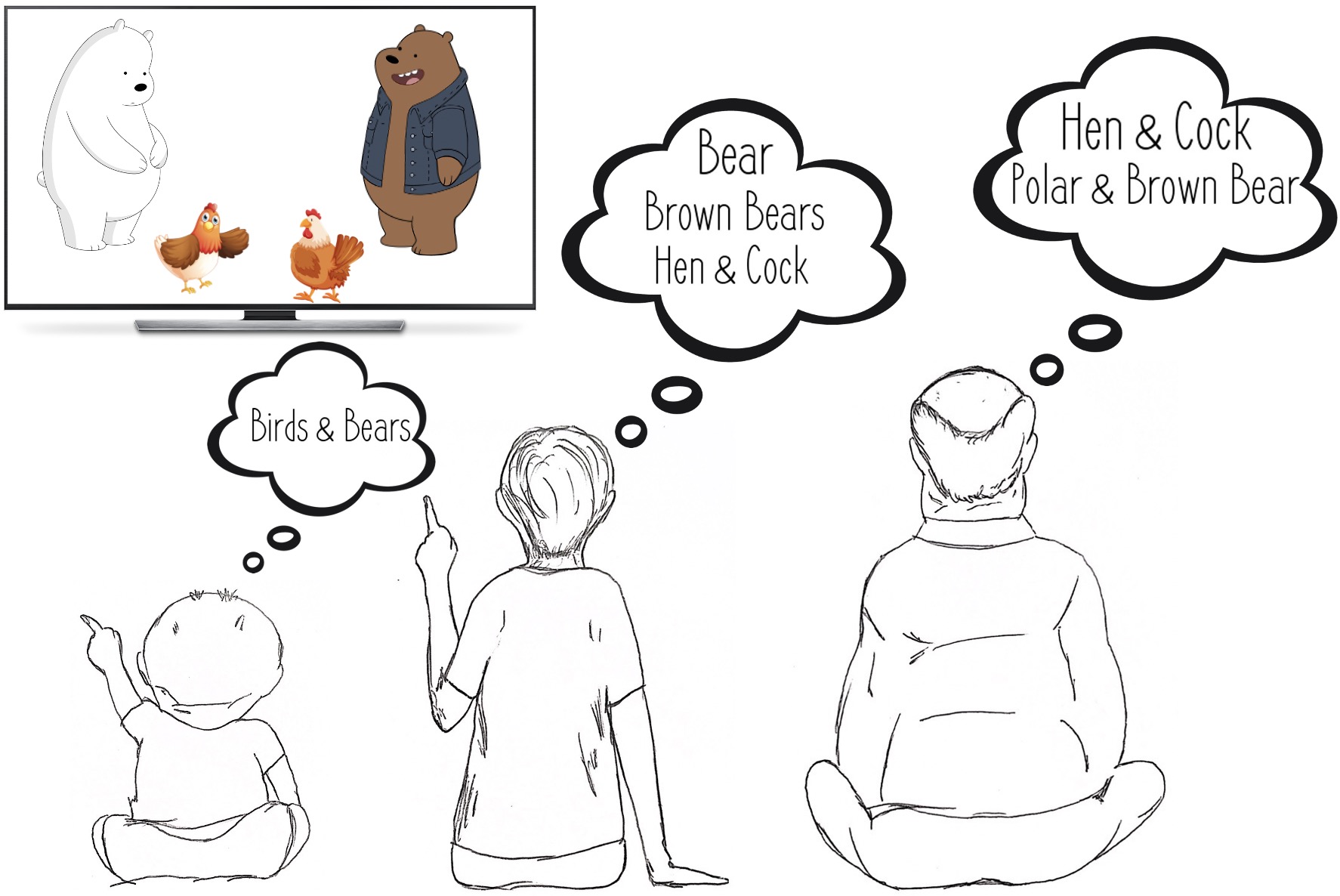}
  \caption{Humans incrementally accumulate knowledge over time. They encounter new entities and discover new information about existing entities. In this process, they associate new \textit{labels} with entities and refine or update their existing \textit{labels}, while ensuring the accumulated knowledge is coherent.}
  \label{fig:setup_summary}
\end{figure}

\begin{figure*}[t]
  \centering\includegraphics[width=0.98\textwidth]{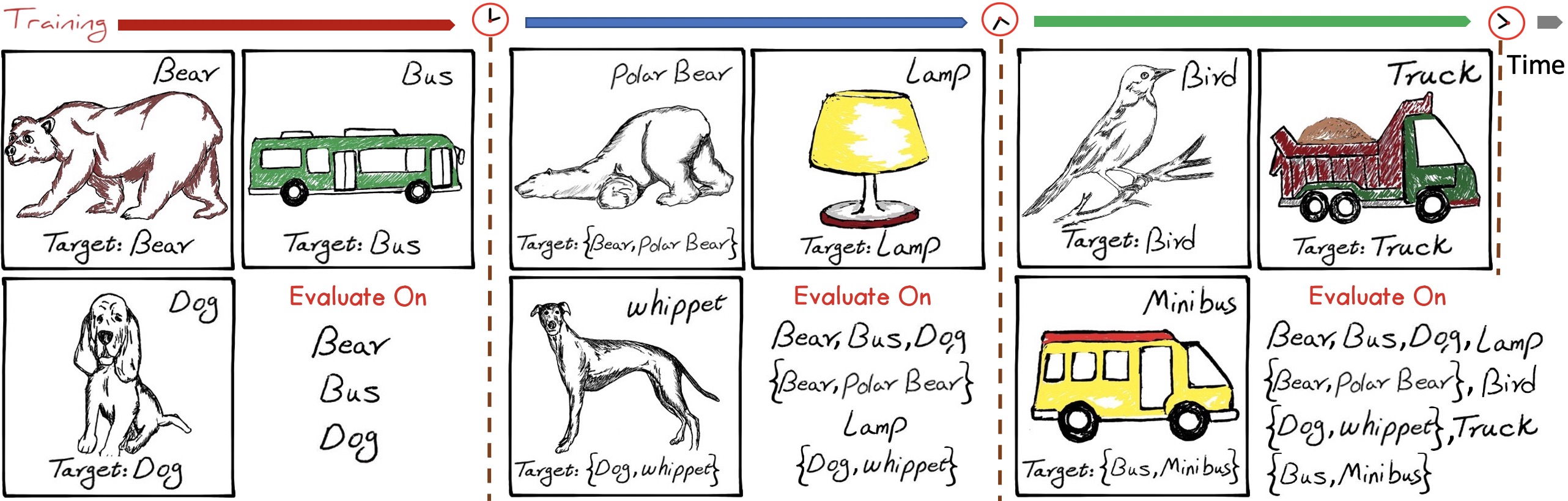}
  \caption{IIRC setup showing how the model expands its knowledge and associates and re-associates labels over time. The top right label shows the label model sees during training, and the bottom label (annotated as ``Target'') is the one that model should predict during evaluation.  The right bottom panel for each task shows the set classes that model is evaluated on and the dashed line shows different tasks.}
  \label{fig:setup_visualization}
\end{figure*}

Deep learning algorithms have led to transformational breakthroughs in computer vision~\cite{imagenet,resnet}, natural language processing~\cite{lstm, transformer}, speech processing~\cite{amodei2016deep, baevski2020wav2vec}, reinforcement learning~\cite{nature_dqn, alpha_zero}, robotics~\cite{hafner2019learning, akkaya2019solving}, recommendation systems~\cite{cheng2016wide, he2017neural} etc. On several tasks, deep learning models have either matched or surpassed human performance. However, such \textit{super-human} performance is still limited to some very narrow and well-defined setups. Moreover, humans can continually learn and accumulate knowledge over their lifetime, while the current learning algorithms are known to suffer from several challenges when training over a sequence of tasks~\cite{mccloskey1989catastrophic, goodfellow2013empirical, chaudhry2018riemannian, sodhani2020toward}. These challenges are broadly studied under the domain of Lifelong Learning~\cite{thrunlifelong}, also called Incremental Learning~\cite{schlimmer1986incremental}, Continual Learning~\cite{ring1997}, and Never Ending Learning~\cite{mitchell2018never}. In the general lifelong learning setup, the model experiences new knowledge, in terms of new tasks, from the same domain or different domains. The model is expected to learn and solve new tasks while retaining useful knowledge from previous tasks.

There are two popular paradigms in lifelong learning~\cite{scenarios}: \textbf{i)} \textit{task incremental} learning, where the model has access to a task delimiter (say a \textit{task id}), which distinguish between tasks. Models for this setup are generally multi-headed, where there exists a separate classification layer for each task. \textbf{ii)} \textit{class incremental learning}, where the model does not have access to a task delimiter, so it needs to discriminate between all classes from all tasks at inference time. Therefore, models developed for this paradigm are generally single-headed. The class incremental setup is more closely aligned with the real-life scenarios and is more challenging than the task incremental scenario.

Several useful benchmarks have been proposed for evaluating models in the lifelong learning setting~\cite{antoniou2020defining,lomonaco2017core50}. While useful for measuring high-level aggregate quantities, these benchmarks take a narrow and limited view on the broad problem of lifelong learning. One common assumption that many class incremental setups make is ``information about a given sample (say label) can not change across tasks''. For example, an image of a bear is always labeled as ``bear'', no matter how much knowledge the model has acquired.

While this assumption appears to be ``obviously correct'' in the context of the supervised learning paradigm (where each sample generally has a fixed label), the assumption is not always satisfied in real-life scenarios. We often interact with the same entities multiple times and discover new information about them. Instead of invalidating the previous knowledge or outright rejecting the new information, we refine our previous knowledge using the new information. Figure~\ref{fig:setup_summary} illustrates an example where a child may recognize all bears as ``bear'' (and hence label them as ``bear''). However, while growing up, they may hear different kinds of bear being called by different names, and so they update their knowledge as: ``Some bears are brown bears, some bears are polar bears, and other bears are just bears. Brown bears and polar bears are both still bears but they are distinct''. This does not mean that their previous knowledge was wrong (or that previous label ``bear'' was ``incorrect''), but they have discovered new information about an entity and have coherently updated their knowledge. This is the general scheme of learning in humans.  

A concrete instantiation of this learning problem is that two similar or even identical input samples have two different labels across two different tasks. We would want the model to learn the new label, associate it with the old label without forgetting the old label. Evaluating lifelong learning models for these capabilities is generally outside the scope of existing benchmarks. We propose the \textit{Incremental Implicitly-Refined Classification (IIRC)} setup to fulfill this gap. We adapt the publicly available CIFAR100 and ImageNet datasets to create a benchmark instance for the IIRC setup and evaluate several well-known algorithms on the benchmark. Our goal is not to develop a new state-of-the-art model but to surface the challenges posed by the IIRC setup.
    
The main contributions of our work are as follows:
    \begin{enumerate}
        \item We propose the \textit{Incremental Implicitly-Refined Classification (IIRC)} setup, where the model starts training with some coarse, high-level classes and observes new, fine-grained classes as it trains over new tasks. During the lifetime of the model, it may encounter a new sample or an old sample with a fine-grained label.
        \item We provide a standardized benchmark to evaluate a lifelong model in the IIRC setup. We adapt the commonly used ImageNet and CIFAR datasets, and provide the benchmark setup compatible with several major deep learning frameworks (PyTorch, Tensorflow, and Jax)\footnote{\url{https://chandar-lab.github.io/IIRC/}}.
        \item We evaluate well-known lifelong learning algorithms on the benchmark and highlight their strengths and limitations, while ensuring that the models are compared in a fair and standardized setup.
    \end{enumerate}

\section{Incremental Implicitly-Refined Classification (IIRC)}
\label{sec:task_setup}
While class incremental learning is a challenging and close-to-real-life formulation of the lifelong learning setup, most existing benchmarks do not explore the full breadth of the complexity. They tend to over-focus on catastrophic forgetting (which is indeed an essential aspect) at the expense of several other unique challenges to the class incremental learning. In this work, we highlight those challenges and propose the \textit{Incremental Implicitly-Refined Classification (IIRC)} setting, an extension of the class incremental learning setting, that enables us to study these under-explored challenges, along with the other well-known challenges like catastrophic forgetting. We provide an instantiation of the setup, in the form of a benchmark, and evaluate several well-known lifelong learning algorithms on it.
\subsection{Under-explored challenges in class incremental learning setting}
In class incremental learning, the model encounters new classes as it trains over new tasks. The nature of the class distributions and the relationship between classes (across tasks) can lead to several interesting challenges for the learning model: If the model is trained on a \textit{high-level} label (say ``bear'') in the initial task and then trained on a \textit{low-level} label, which is a refined category of the previous label (say ``polar bear''), what kind of associations will the model learn and what associations will it forget? Will the model generalize and label the images of polar bear as both ``bear'' and ``polar bear''? Will the model catastrophically forget the concept of ``bear''? Will the model infer the spurious correlation: ``all bears are polar bears''? What happens if the model sees different labels (at different levels of granularity) for the \textit{same sample} (across different tasks)? Does the model remember the latest label or the oldest label or does it remember all the labels? These challenges can not be trivially overcome by removing restrictions on memory or replay buffer capacity (as we show in Section~\ref{sec:experiments}).
\subsection{Terminology}
We describe the terminology used in the paper with the help of an example. As shown in Figure~\ref{fig:setup_visualization}, at the start, the model trains on data corresponding to classes ``bear'', ``bus'' and ``dog''. Training the model on data corresponding to these three classes is the first \textbf{task}. After some time, a new set of classes (``polar bear'', ``lamp'' and ``whippet'') is encountered, forming the second task. Since ``whippet'' is a type of ``dog'', it is referred as the \textbf{subclass}, while ``dog'' is refereed as the \textbf{superclass}. The ``dog-whippet'' pair is referred to as the superclass-subclass pair. Note that not all classes have a superclass (example ``lamp''). We refer to these classes as subclasses as well, though they do not have any superclasses. When training the model on an example of a ``whippet'', we may provide only ``whippet'' as the supervised learning label. This setup is referred to as the \textbf{incomplete information} setup, where if a task sample has two labels, only the label that belongs to the current task is provided. Alternatively, we may provide both ``whippet'' and ``dog'' as the supervised learning labels. This setup is referred as the \textbf{complete information} setup, where if a task sample has two labels, labels that belong to the current and previous tasks are provided. Note that majority of our experiments are performed in the incomplete information setup as that is closer to the real life setup, requiring the model to recall the previous knowledge when it encounters some new information about a known entity. We want to emphasize that the use of the word \textbf{task} in our setup refers to the arrival of a new batch of classes for the model to train on in a single-head setting, and so it is different from it's use to indicate a distinct classification head in the task incremental learning.

As the model is usually trained in an \textit{incomplete information} setup, it would need access to a validation set to monitor the progress in training that is still an \textit{incomplete information} set, otherwise there would be some sort of labels leakage. On the other hand, after training on a specific task, the model has to be be evaluated on a \textit{complete information} set, hence a \textit{complete information} validation set is needed to be used during the process of model development and tweaking, so as to not overfit on the test set. We provide both in the benchmark, where we call the first one the \textbf{in-task validation set}, while the latter one the \textbf{post-task validation set}.
\subsection{Setup}
We describe the high-level design of the IIRC setup (for a visual illustration, see Figure~\ref{fig:setup_visualization}). We have access to a series of $N$ tasks denoted as $T_1, \cdots, T_N$. Each task comprises of three collections of datasets, for training, validation and testing. Each sample can have one or two labels associated with it. In the case of two labels, one label is a subclass and the other label is a superclass. For any superclass-subclass pair, the superclass is always introduced in an earlier task, with the intuition that a high-level label should be relatively easier to learn. Moreover, the number of samples for a superclass is always more than the number of samples for a subclass (it increases with the number of subclasses, up to a limit). During training, we always follow the incomplete information setup. During the first task, only a subset of superclasses (and no sublcasses) are used to train the model. The first task has more classes (and samples), as compared to the other tasks and it can be seen as a kind of pretraining task. The subsequent tasks have a mix of superclasses and subclasses. During the training phase, the model is evaluated on the in-task validation set (with incomplete information), and during the evaluation phase, the model is evaluated on the post-task validation set and the test set (both with complete information).

\section{Related Work}
\label{sec:related_word}
Lifelong Learning is a broad, multi-disciplinary, and expansive research domain with several synonyms:
Incremental Learning~\cite{schlimmer1986incremental}, Continual Learning~\cite{ring1997}, and Never Ending Learning~\cite{mitchell2018never}. One dimension for organizing the existing literature is whether the model has access to explicit task delimiters or not, where the former case is referred to as task incremental learning, and the latter case, which is closely related to our setup IIRC, is referred to as class incremental learning. 

In terms of learning methods, there are three main approaches~\cite{lange2019continual}: \textbf{i)} replay based, \textbf{ii)} regularization based, and  \textbf{iii)} parameter isolation methods. Parameter isolation methods tend to be computationally expensive and require access to a task identifier, making them a good fit for the task incremental setup. Prominent works that follow this approach include Piggyback~\cite{mallya2018piggyback}, PackNet~\cite{mallya2018packnet}, HAT~\cite{serra2018overcoming}, TFM~\cite{masana2020ternary}, DAN~\cite{rosenfeld2018incremental}, PathNet~\cite{fernando2017pathnet}. The replay and regularization based approaches can be used with both task and class incremental setups, however, replay based approaches usually perform better in the class incremental setup~\cite{masana2020class}. Among the regularization based approaches, LwF~\cite{li2017learning} uses finetuning with distillation. LwM~\cite{dhar2019learning} improves LwF by adding an attention loss. MAS~\cite{aljundi2018memory}, EWC~\cite{kirkpatrick2017overcoming}, SI~\cite{si2017} and RWalk~\cite{chaudhry2018riemannian} estimate the importance of network parameters, and penalize changes to important ones. As for the replay based approaches, iCaRL~\cite{rebuffi2017icarl} is considered an important baseline in the field. iCaRL selects exemplars for the replay buffer using herding strategy, and alleviates catastrophic forgetting by using distillation loss during training, and using a nearest-mean-of-exemplars classifier during inference. EEIL~\cite{castro2018end} modifies iCaRL by learning the feature extractor and the classifier jointly in an end to end manner. LUCIR~\cite{hou2019learning} applies the distillation loss on the normalized latent space rather than the output space, proposes to replace the standard softmax layer with a cosine normalization layer, and uses a margin ranking loss to ensure a large margin between the old and new classes. Other works include LGM~\cite{ramapuram2017lifelong}, IL2M~\cite{belouadah2019il2m}, BIC~\cite{wu2019large}, and ER~\cite{rolnick2019experience}. GEM is another replay-based method, which solves a constrained optimization problem. It uses the replay buffer to constrain the gradients on the current task so that the loss on the previous tasks does not increase. A-GEM~\cite{chaudhry2019efficient} improves over GEM by relaxing some of the constraints, and hence increasing the efficiency, while retaining the performance. Finally, \cite{chaudhry2019tiny} shows that vanilla experience replay, where the model simply trains on the replay buffer along with the new task data, is by itself a very strong baseline. In this work, we include variants of iCaRL, LUCIR, A-Gem, and vanilla experience replay as baselines.

We propose a benchmark for evaluating a model's performance in the IIRC setup, as having a realistic, standardized, and large-scale benchmark helps provide a fair and reproducible comparison for the different approaches. Existing efforts for benchmarking the existing lifelong learning setups include CORe50 benchmark~\cite{lomonaco2017core50}, and ~\cite{antoniou2020defining} that proposes a benchmark for continual few-shot learning.

Our work is also related to knowledge (or concept) drift, where the statistical properties of the data changes over time and old knowledge can become ``irrelevant'' ~\cite{lu2014concept,Lu_2018}. Unlike those works, we focus on learning new associations and updating existing associations as new tasks are learnt. As the model acquires new knowledge, the old knowledge does not become `irrelevant''. Recently, BREEDS~\cite{breeds} proposed a benchmark to evaluate model's generalization capabilities in the context of subpopulation shift. Specifically, they define a hierarchy and train the model on samples corresponding to some subpopulations (e.g. ``poodles'' and ``terriers'' are subpopulations of ``dogs''). The model is then evaluated on samples from an unseen subpopulation. e.g. it should label ``dalmatians'' as ``dogs''. While at a quick glance, IIRC might appear similar to BREEDS, there are several differences. IIRC focuses on the lifelong learning paradigm while BREEDS focuses on generalization. Moreover, the training and evaluation setups are also different. If we were to extend the dogs example to IIRC, the model may first train on some examples of ``poodles`` and ``terriers'' (labeled as ``dogs''). In the next task, it may train on some exampled of ``poodles'' (labeled as ``poodles''). When the model is evaluated on both tasks, it should predict both labels (``poodles'' and ``dogs'') for the images of poodles.

\section{Benchmark}
\label{sec:benchmark}
\subsection{Dataset}

We use two popular computer vision datasets in our benchmark - ImageNet~\cite{imagenet} and CIFAR100~\cite{krizhevsky2009cifar}. For both the datasets, we create a two-level hierarchy of class labels, where each label starts as a leaf-node and similar labels are assigned a common parent. The leaf-nodes are the \textit{subclasses} and the parent-nodes are the \textit{super-classes}. Some of the subclasses do not have a corresponding superclass, so as to enrich the setup and make it more realistic. While the datasets come with a pre-defined hierarchy (e.g. ImageNet follow the WordNet hierarchy), we develop a new hierarchy as the existing hierarchy focuses more on the semantics of the labels and less on the visual similarity (e.g, ``sliding door'' and ``fence'' are both grouped under ``barriers''). We refer to these adapted datasets as IIRC-ImageNet and IIRC-CIFAR.

\begin{figure}[t]
  \centering\includegraphics[width=0.98\linewidth]{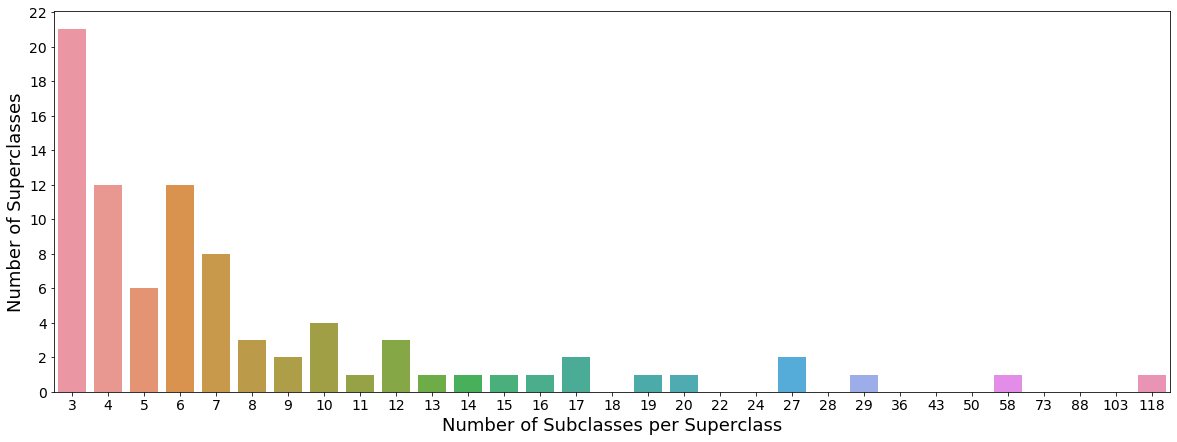}
  \caption{The distribution of the number of subclasses per superclass on IIRC-Imagenet.}
  \label{fig:imagenet_distribution}
\end{figure}

In IIRC-CIFAR, each superclass has similar number of subclasses (four to eight). However, the sub-class distribution for IIRC-ImageNet is very skewed (Figure~\ref{fig:imagenet_distribution}) and number of sublcasses varies from 3 to  118. We explicitly decided not to \textit{fix} this imbalance to ensure that visually similar classes are grouped together. Moreover, in the real life, not all classes are observed at the same frequency, making our setup more realistic. More statistics and the full class hierarchies for both IIRC-ImageNet and IIRC-CIFAR are provided in Appendix-\ref{appendix:dataset_statistics} and \ref{appendix:clusters}.

As mentioned in Section~\ref{sec:task_setup}, we use two validation sets - one with incomplete information (for model selection and monitoring per-task performance) and one with complete information (for the model evaluation after each task). Each validation dataset comprises 10\% of the training data for CIFAR, and 4\% of the training data for ImageNet, and is fixed through all the runs. Some aggregate information about the splits is provided in Table~\ref{table:splits} in Appendix.

Since we are creating the class hierarchy, superclasses do not have any samples assigned to them. For the training set and the in-task validation set, we assign 40\% of samples from each subclass to its superclass, while retaining 80\% of the samples for the subclass. This means that subclass-superclass pairs share about 20\% of the samples or, for 20\% of the cases, the model observes the same sample with different labels (across different tasks). Since some superclasses have an extremely large number of subclasses, we limit the total number of samples in a superclass. A superclass with more than eight subclasses, uses $\frac{8}{\text{number of subclasses}} \times 40\%$ of samples from its subclasses. We provide the pseudo code for the dataloader in Appendix~\ref{appendix:pseudo_codes}.

Now that we have a dataset with superclasses and subclasses, and with samples for both kind of classes, the tasks are created as follows: The first task is always the largest task with 63 superclasses for IIRC-ImageNet and 10 superclasses for IIRC-CIFAR. In the supsequent tasks, each new task introduces 30 classes for IIRC-ImageNet and 5 classes for IIRC-CIFAR. Recall that each task introduces a mix of superclasses and subclasses. IIRC-ImageNet has a total of 35 tasks, while IIRC-CIFAR has a total of 22 tasks. Since the order of classes can have a bearing on the models' evaluation, we create 5 preset class orders (called task configurations) for IIRC-ImageNet and 10 task configurations for IIRC-CIFAR, and report the average (and standard deviation) of the performance on these configurations.  

Finally, we acknowledge that while IIRC-ImageNet provides interesting challenges in terms of data diversity, training on the dataset could be difficult and time consuming. Hence, we provide a shorter, lighter version which has just ten tasks (with five tasks configurations). We shall call the original version IIRC-ImageNet-full, and the lighter version IIRC-ImageNet-lite, while referring to both collectively as IIRC-ImageNet. Although we do not recommend the use of this lighter version for benchmarking the model performance, we hope that it will make it easier for others to perform quick, debugging experiments. We report all the metrics on IIRC-ImageNet-lite as well. 

\subsection{Metrics}

\begin{figure*}[t]
\begin{center}
\subfigure{
\includegraphics[clip,trim=3.5cm 1.5cm 4.5cm 2.75cm, width=0.48\linewidth]{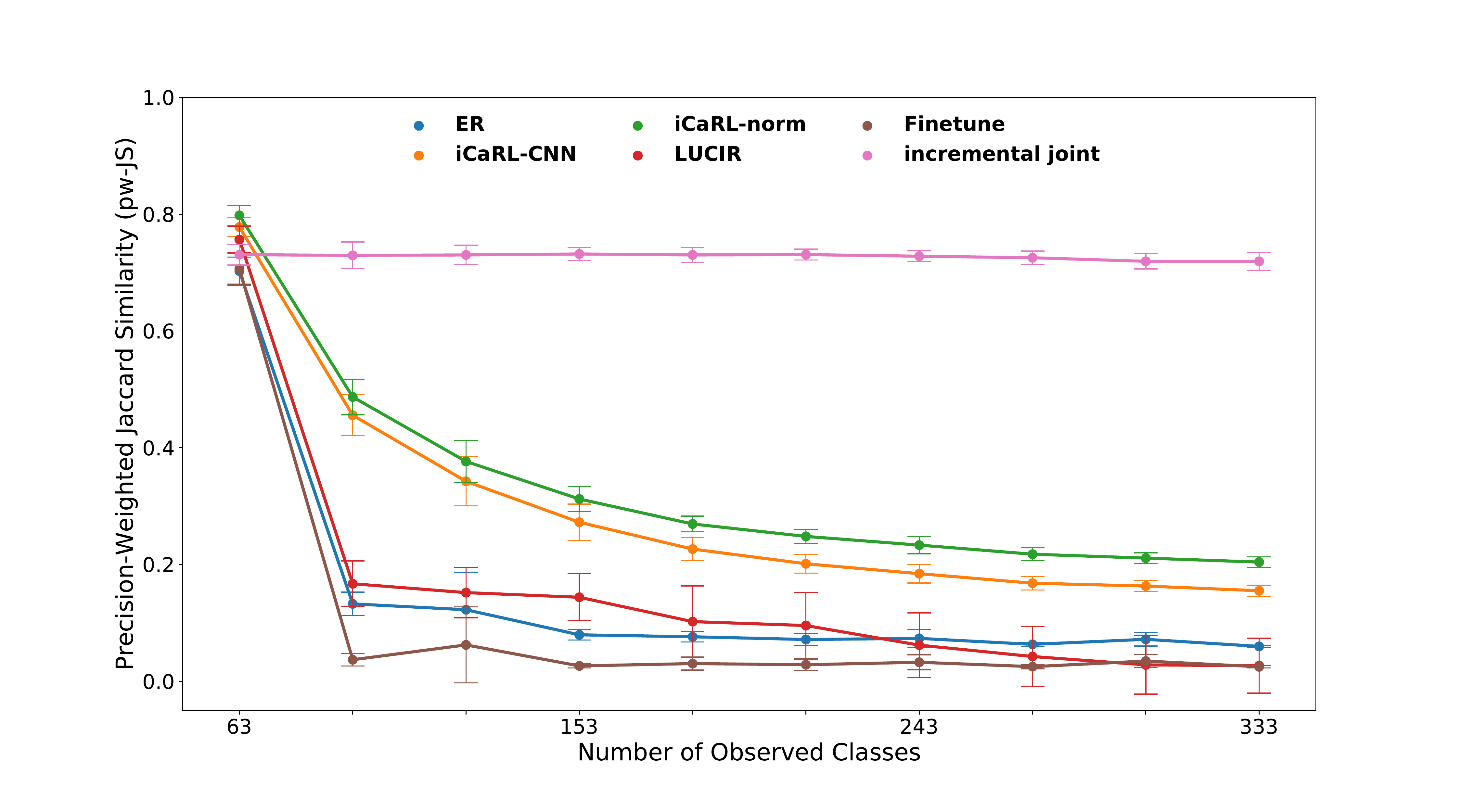}
\label{fig:imagenet_lite_incremental}
}
\subfigure{
\includegraphics[clip,trim=3.5cm 1.5cm 4.5cm 2.75cm, width=0.48\linewidth]{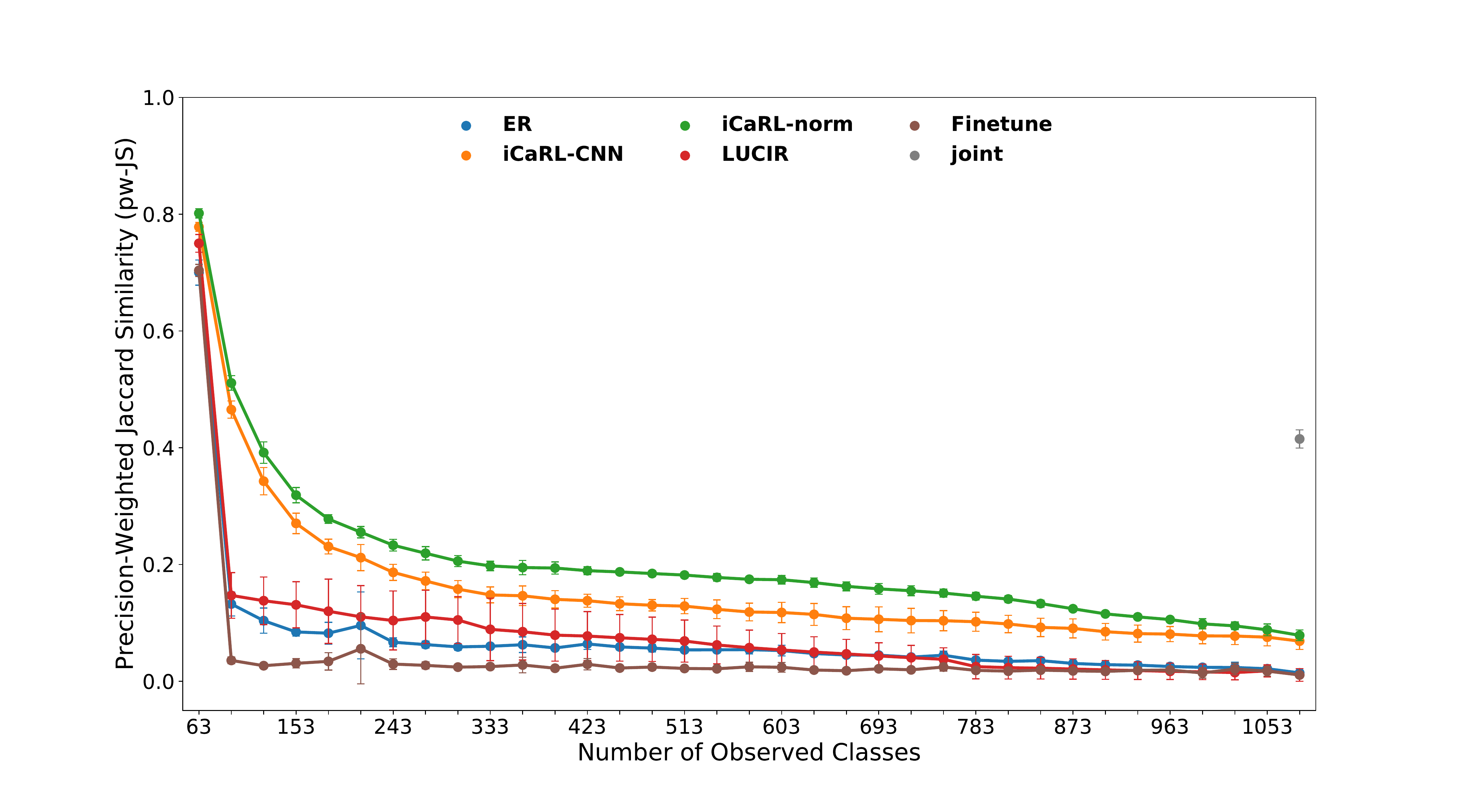}
}
\label{fig:imagenet_full_incremental}
\end{center}
\caption{Average performance using the precision-weighted Jaccard Similarity. (left) IIRC-Imagenet-lite and (right) IIRC-Imagenet-full. We run experiments on five different task configurations and report the mean and standard deviation.}
\label{fig:imagenet_incremental}
\end{figure*}

\begin{figure}[b]
\begin{center}
\includegraphics[clip,trim=3.5cm 1.5cm 4.5cm 2.75cm, width=0.98\linewidth]{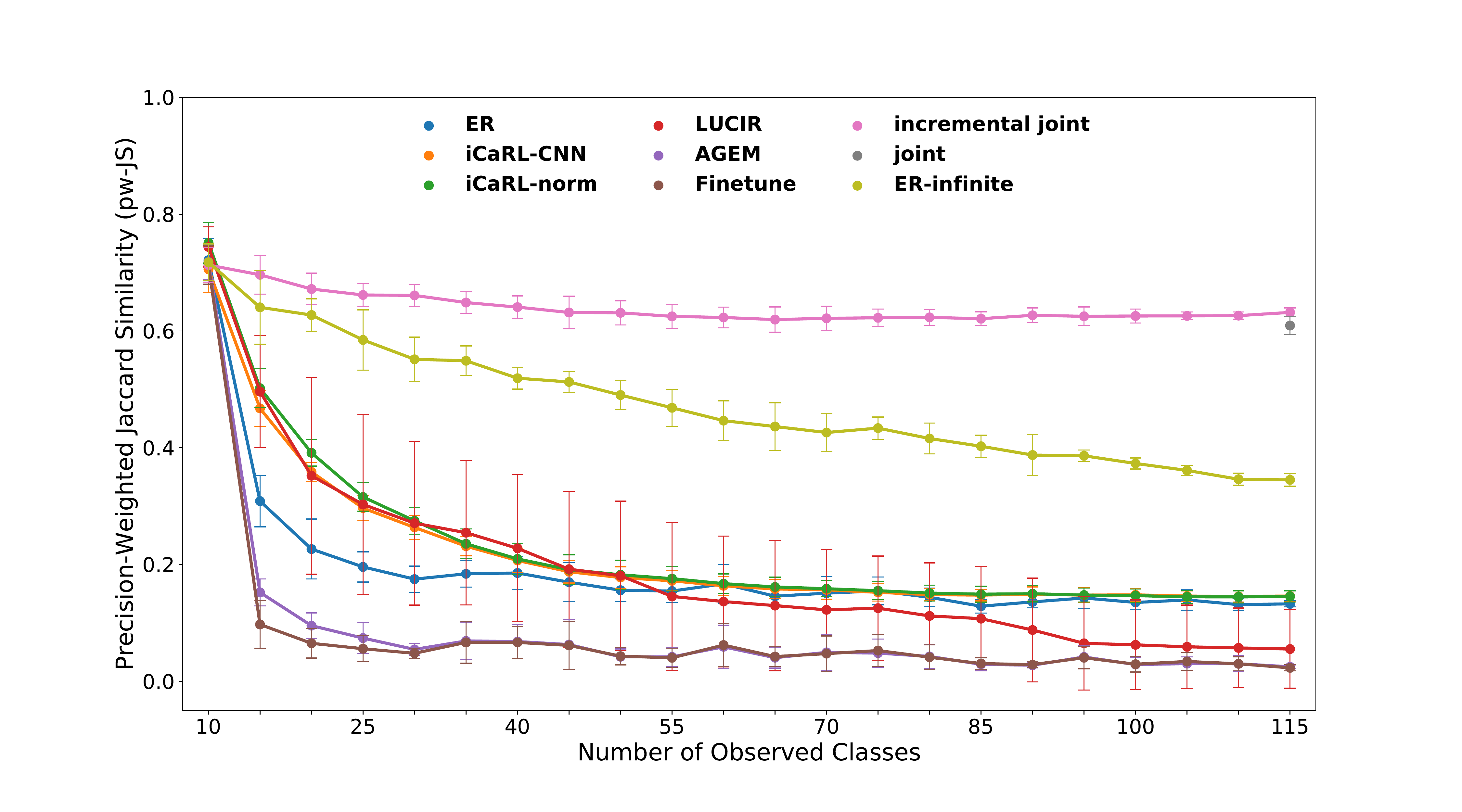}
\end{center}
\caption{Average performance on IIRC-CIFAR. We run experiments on ten different task configurations and report the mean and standard deviation.}
\label{fig:cifar_incremental_mjaccard}
\end{figure}

Most lifelong learning benchmarks operate in the single-label classification setup, making accuracy the appropriate metric. In our setup, the model should be able to predict multiple labels for each sample, even if those labels are seen across different tasks. We considered using the \textit{Exact-Match Ratio} ($MR$) metric \cite{multilabelsurvey}, a multi-label extension of the accuracy metric. $MR$ is defined as $\frac{1}{n} \sum_{i=1}^n I(Y_i == \hat{Y}_i)$ where $I$ is the indicator function, $\hat{Y}_i$ are the set of (model) predictions for the $i_{th}$ sample, $Y_i$ are the ground truth labels, and $n$ is the total number of samples. One limitation is that it does not differentiate between partially incorrect predictions and completely incorrect predictions.

Another popular metric (for multi-label classification) is the Jaccard similarity($JS$), also called ``intersection over union''\cite{multilabelsurvey}. $JS$ is defined as $\frac{1}{n} \sum_{i=1}^n \frac{|Y_i \cap \hat{Y}_i|}{|Y_i \cup \hat{Y}_i|}$. 
To further penalize the imprecise models, we \textit{weight} the Jaccard similarity by the per sample precision (i.e., the ratio of true positives over the sum of true positives and false positives). We refer to this metric as the \textit{precision-weighted Jaccard similarity (pw-JS)}.

We measure the performance of a model on task $k$ after training on task $j$ using the precision-weighted Jaccard similarity, denoted $R_{jk}$, as follow:
\begin{equation}
    R_{jk} = \frac{1}{n_k}\sum_{i=1}^{n_k } \frac{|Y_{ki} \cap \Hat{Y}_{ki}|}{|Y_{ki} \cup \Hat{Y}_{ki}|} \times \frac{|Y_{ki} \cap \Hat{Y}_{ki}|}{|\Hat{Y}_{ki}|}\,, 
\label{equ:r_metric}
\end{equation}
where ($j \geq k$), $\hat{Y}_{ki}$ is the set of (model) predictions for the $i_{th}$ sample in the $k_{th}$ task, $Y_{ki}$ are the ground truth labels, and $n_k$ is number of samples in the task.
$R_{jk}$ can be used as a proxy for the model's performance on the $k^{th}$ task as it trains on more tasks (i.e. as the $j$ increases). 

We evaluate the overall performance of the model after training till the task $j$, as the average \textit{precision-weighted Jaccard similarity} over all the classes that the model has encountered so far. Note that during this evaluation, the model has to predict all the correct labels for a given sample, even if the labels were seen across different tasks (i.e. the evaluation is performed in the complete information setup). We denote this metric as $R_j$ and computed it as follow:
\begin{equation}
    R_{j} = \frac{1}{n}\sum_{i=1}^{n} \frac{|Y_{i} \cap \Hat{Y}_{i}|}{|Y_{i} \cup \Hat{Y}_{i}|} \times \frac{|Y_{i} \cap \Hat{Y}_{i}|}{|\Hat{Y}_{i}|}\,,
\label{equ:r_metric_avg}
\end{equation}
where $n$ is the total number of evaluation samples for all the tasks seen so far.

\section{Baselines}
\label{sec:baselines}

\begin{figure*}[t]
\begin{center}
\subfigure{
\includegraphics[clip,trim=3.5cm 1.5cm 4.5cm 2.75cm, width=0.48\linewidth]{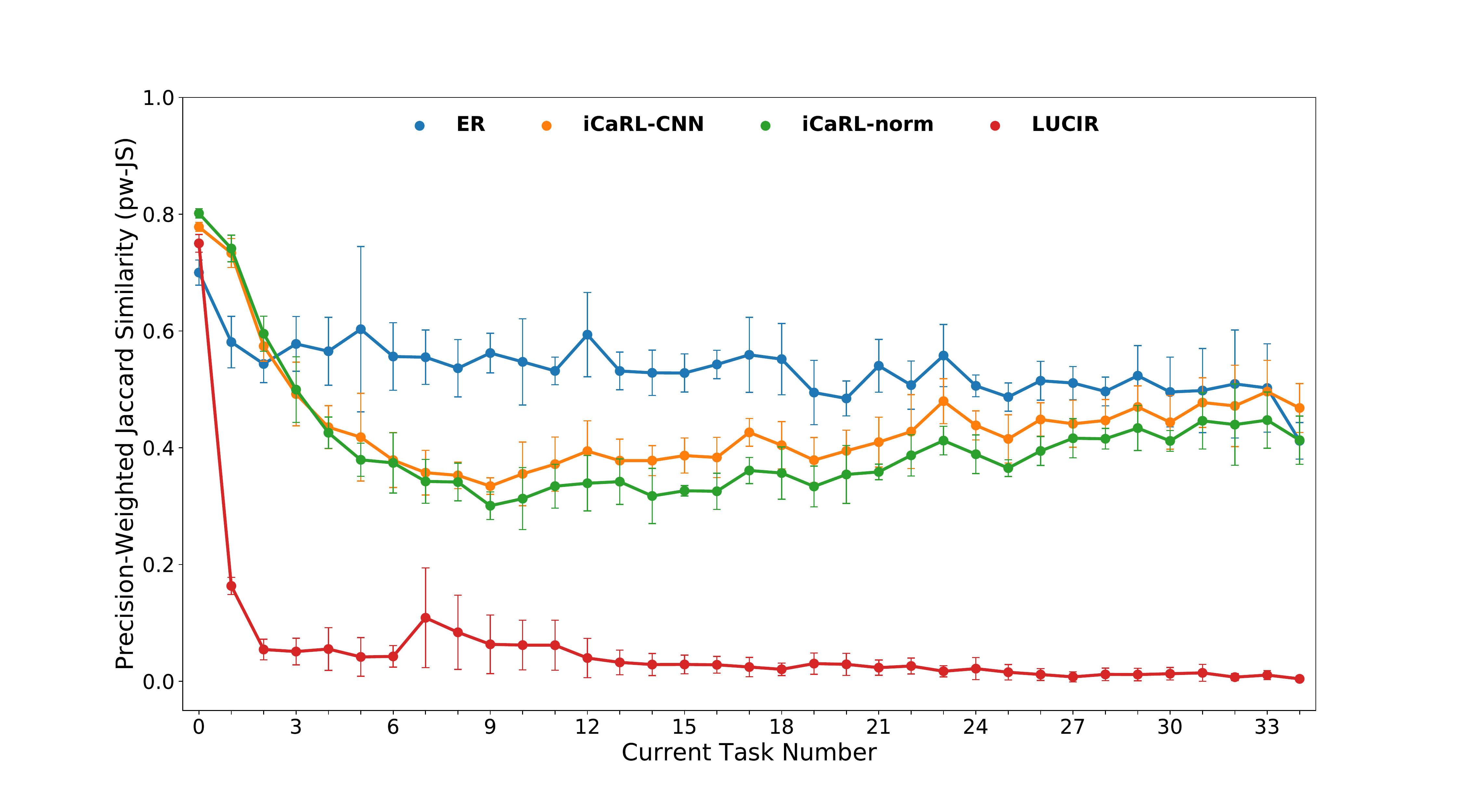}
\label{fig:lite_per_task}
}
\subfigure{
\includegraphics[clip,trim=3.5cm 1.5cm 4.5cm 2.75cm, width=0.48\linewidth]{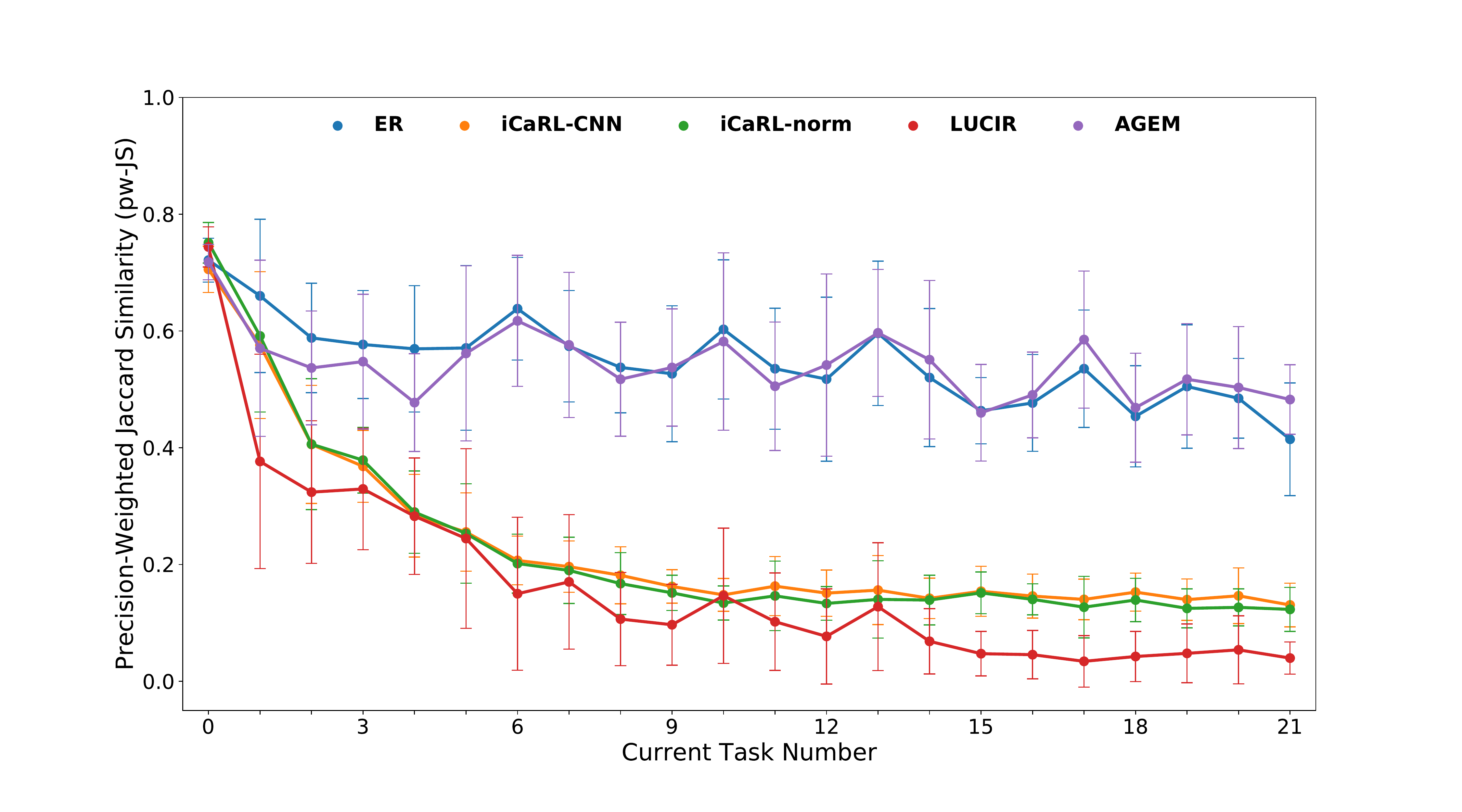}
\label{fig:cifar_per_task}
}
\end{center}
\caption{Per task performance over the test samples of a specific task $j$, after training on that task ($R_{jj}$ using Equation~\ref{equ:r_metric}). (left) IIRC-Imagenet-full and (right) IIRC-CIFAR.}
\label{fig:performance_per_task}
\end{figure*}

We evaluate several well-known lifelong learning baselines. We also consider two training setups where the model has access to all the labels for a given sample (complete information setup): \textbf{i)} \textit{joint} where the model is jointly trained on all the classes/tasks at once and \textbf{ii)} \textit{incremental joint} where as the model trains across tasks, it has access to all the data from the previous tasks in a \textit{complete information} setup. In the \textbf{Finetune} baseline, the model continues training on new batches of classes without using any replay buffer. Vanilla \textbf{Experience Replay (ER)} method finetunes the model on new classes, while keeping some older samples in the replay buffer and rehearsing on them. \textbf{Experience Replay with infinite buffer (ER-infinite)} is similar to \textit{incremental joint}, but in \textit{incomplete information} setup as in ER. This means that if a new label is introduced that applies to an old sample, the target for that sample will be updated with that new label in the incremental joint baseline but not in the ER-infinte baseline . We also have \textbf{A-GEM}~\cite{chaudhry2019efficient} that is a constrained optimization method in the replay-based methods category. It provides an efficient version of GEM~\cite{lopez2017gradient} by minimizing the average memory loss over the previous tasks at every training step. Another baseline is \textbf{iCaRL}~\cite{rebuffi2017icarl} that proposed using the exemplar rehearsal along with a distillation loss. \textbf{LUCIR}~\cite{hou2019learning} is a replay-based class incremental method that alleviates the catastrophic forgetting and the negative effect of the imbalance between the older and newer classes. More details about the baselines can be found in Appendix-\ref{appendix:baselines_details}. 

\subsection{Model Adaptations}
The earlier-stated baselines were proposed for the single label class incremental setup, while IIRC setup requires the model to be able to make multi-label predictions. Therefore, some changes have to be applied to the different models to make them applicable in the IIRC setup. To this end, we use the binary cross-entropy loss (BCE) as the classification loss. This loss is averaged by the number of observed classes so that it doesn't increase as the number of classes increases during training. During prediction, a sigmoid activation is used and classes with values above 0.5 are considered the predicted labels. Using the nearest-mean-classifier strategy for classifying samples in iCaRL is not feasible for our setting, as the model should be able to predict a variable number of labels. To overcome this issue, we use the output of the classification layer, which was used during training, and call this variant as iCaRL-CNN. 
We further consider a variant of iCaRL-CNN, called iCaRL-norm, which uses cosine normalization in the last layer. ~\cite{hou2019learning} suggests that using this normalization improves the performance in the context of incremental learning. Hence the classification score is calculated as:
\begin{equation}
    p_i(x) = \sigma(\eta \langle \bar{\theta}_i\,, \bar{f}(x) \rangle)\,,
\end{equation}
where $\sigma$ is the sigmoid function, $\bar{\theta}_i$ are the normalized weights of the last layer that correspond to label $i$, and $\bar{f}(x)$ is the output of the last hidden layer for sample $x$. $\eta$ is a learnable scalar that controls the peakiness of the sigmoid. It is important to have $\eta$ since $\langle \bar{\theta}_i\,, \bar{f}(x) \rangle$ is restricted to $[-1,1]$. We can either fix the $\eta$ or consider it as a learnable parameter. We observed that learning  $\eta$ works better in practice.

\begin{figure*}[t]
\begin{center}
\subfigure[ground truth]{
\includegraphics[width=0.23\linewidth]{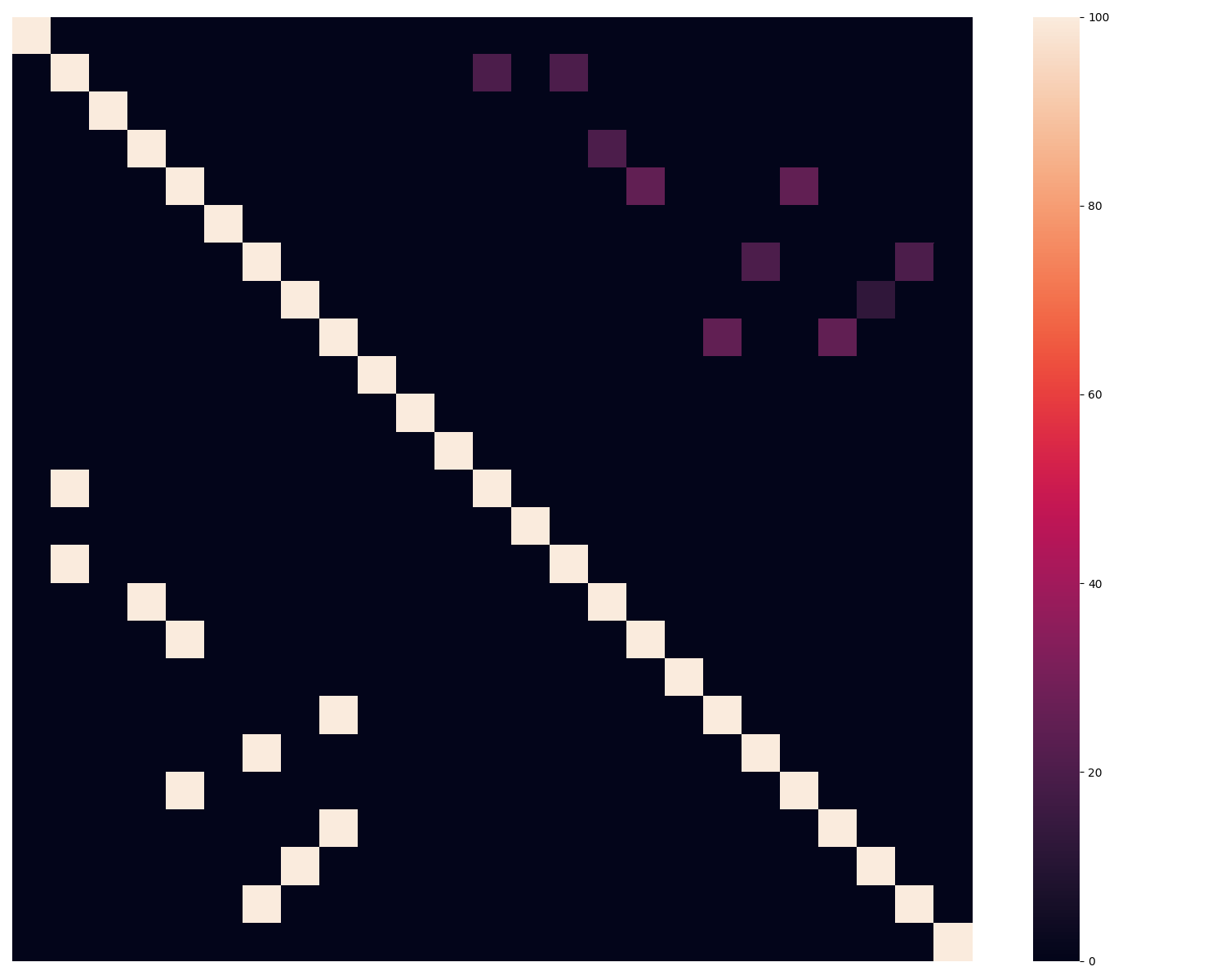}
\label{fig:cifar_confusion_gt}
}
\subfigure[ER]{
\includegraphics[width=0.23\linewidth]{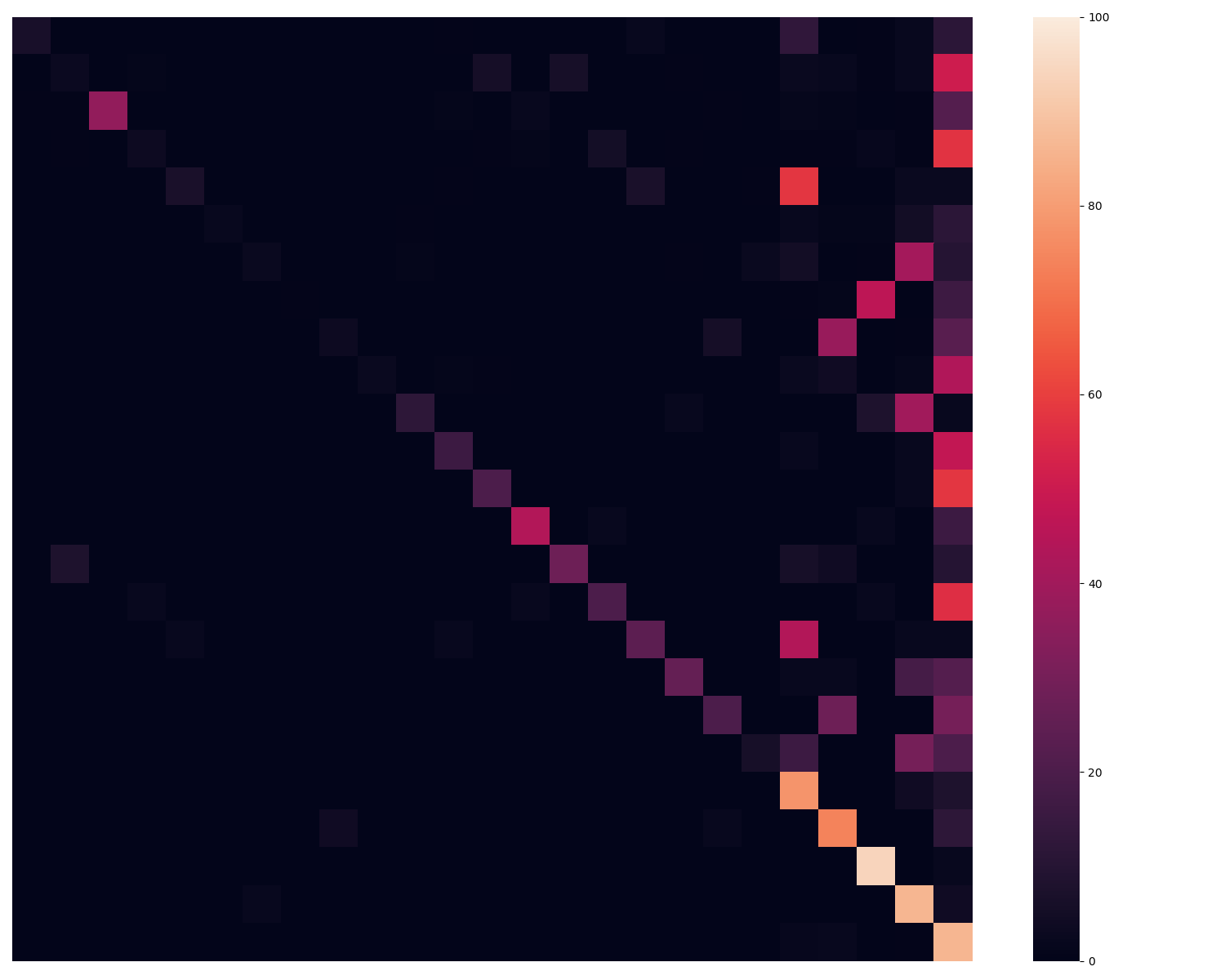}
\label{fig:cifar_confusion_replay}
}
\subfigure[iCaRL-norm]{
\includegraphics[width=0.23\linewidth]{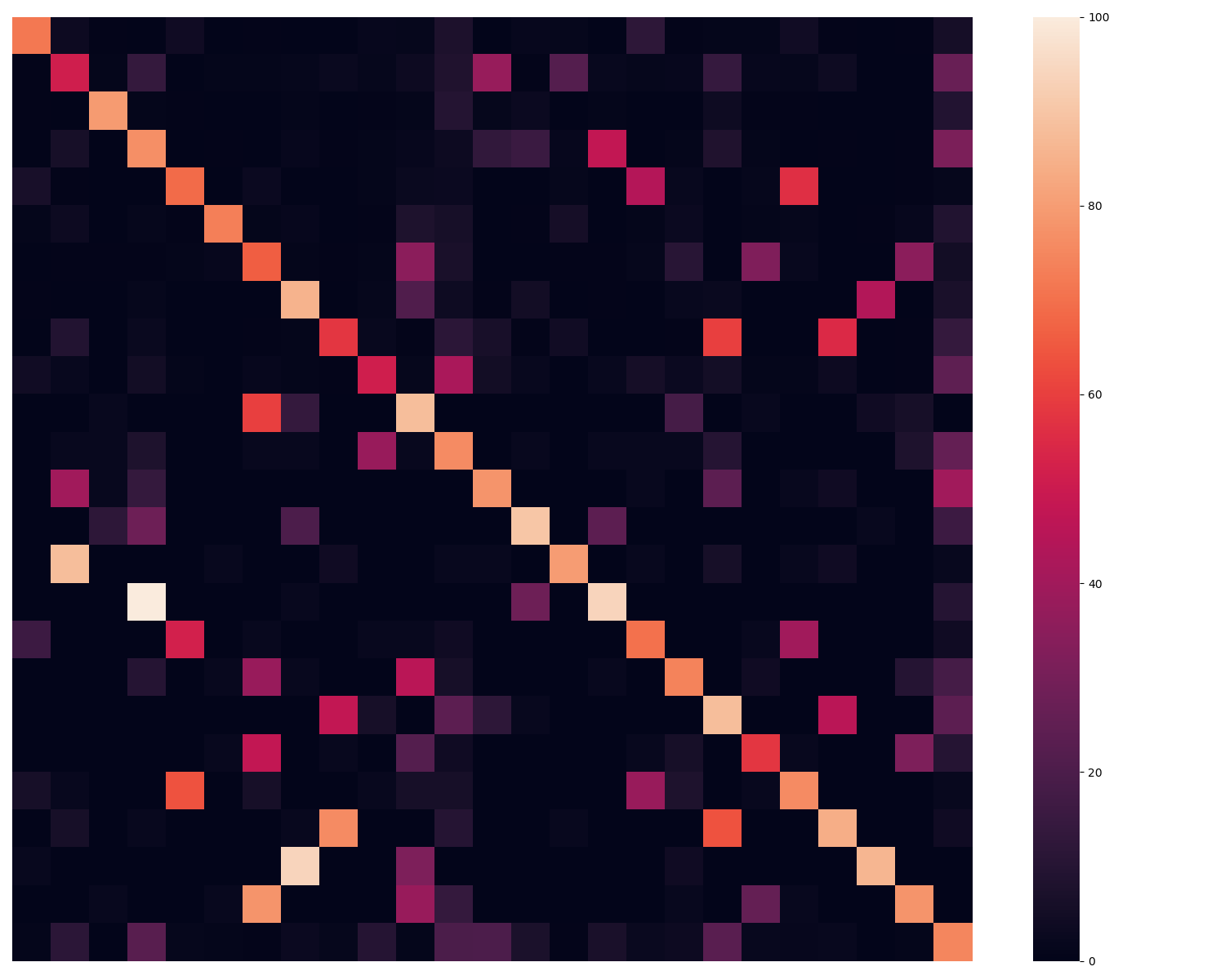}
\label{fig:cifar_confusion_icarl}
}
\subfigure[LUCIR]{
\includegraphics[width=0.23\linewidth]{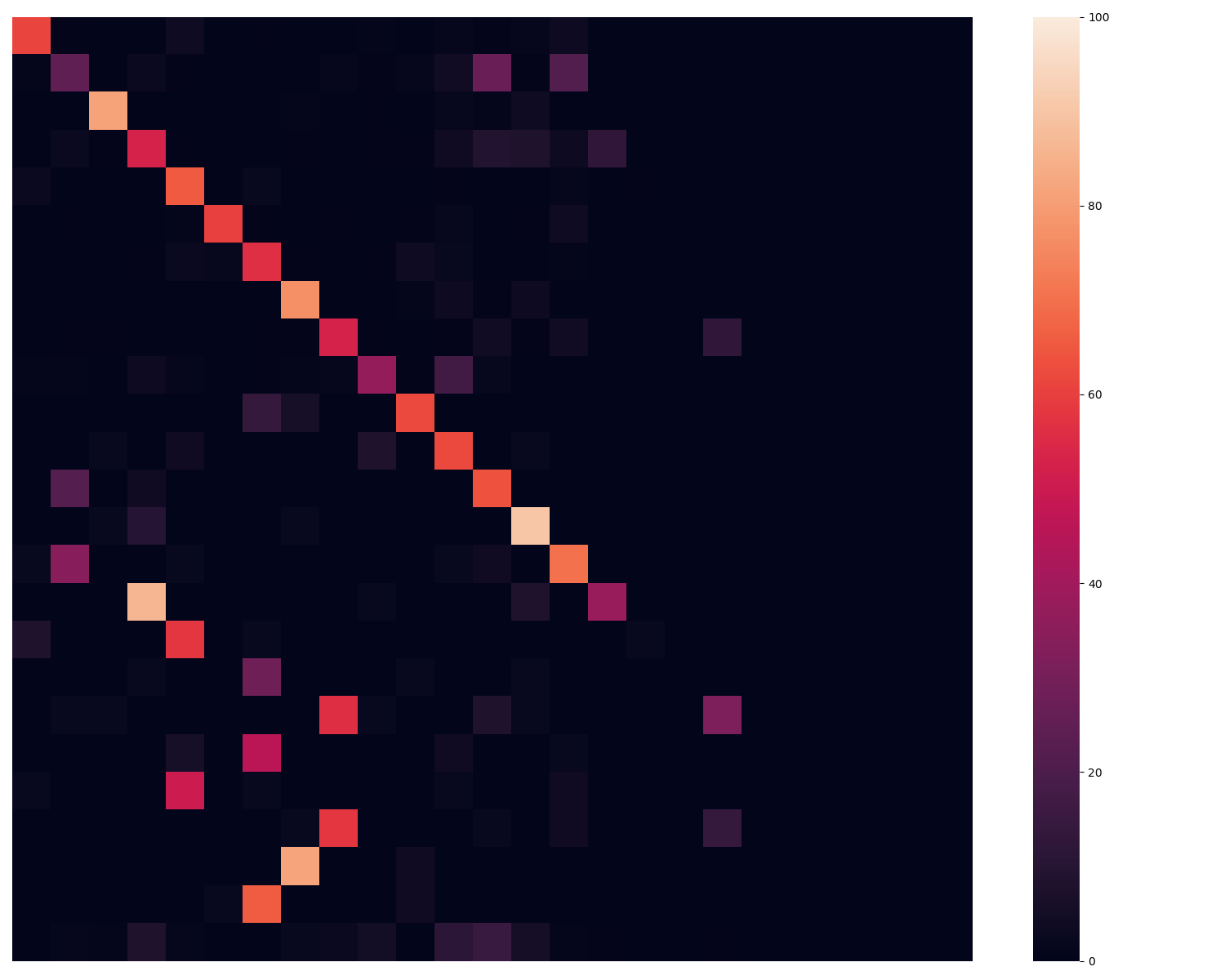}
\label{fig:cifar_confusion_unified}
}
\end{center}
\caption{Confusion matrix after training on task 10 of IIRC-CIFAR. The y-axis is the correct label (or one of the correct labels). The x-axis is the model predicted labels. Labels are arranged by their order of introduction. Only 25 labels are shown for better visibility. See Appendix~\ref{confusion_full_resolution} for the full resolution figures with labels.
}
\label{fig:cifar_confusion}
\end{figure*}

\section{Experiments}
\label{sec:experiments}
We design our experimental setup to surface challenges that lifelong learning algorithms face when operating in the IIRC setup. Our goal is neither to develop a new state-of-the-art model nor to rank existing models. We aim to highlight the strengths and weakness of the dominant lifelong learning algorithms, with the hope that this analysis will spur new research directions in the field. We use the ResNet architecture~\cite{resnet}, with ResNet-50 for IIRC-ImageNet and reduced ResNet-32 for IIRC-CIFAR. Additional implementation details and hyperparameters can be found in Section~\ref{appendix:hyperparameter_details} in the Appendix. Data used to plot the figures is provided in Appendix~\ref{appendix:results_tables} for easier future comparisons.
\subsection{Results and Discussion}
We start by analyzing how well does the model perform over all the observed classes as it encounters new classes. Specifically, as the model finishes training on the $j^{th}$ task, we report the average performance $R_j$, as measured by the \textit{pw-JS} metric using Equation~\ref{equ:r_metric_avg}, over the evaluation set of all the tasks the model has seen so far (Figures~\ref{fig:imagenet_incremental} and \ref{fig:cifar_incremental_mjaccard}). Recall that when computing $R_j$, the model has to predict all the correct labels for a given sample, even if the labels were seen across different tasks. This makes $R_j$ a challenging metric as the model can not achieve a good performance just by memorizing the older labels, but it has to learn the relationship between labels. 

In Figures~\ref{fig:imagenet_incremental} and \ref{fig:cifar_incremental_mjaccard}, we observe that the iCaRL-CNN and iCaRL-norm models perform relatively better than the other methods, with iCaRL-norm having the edge in the case of IIRC-ImageNet. However, this trend does not describe the full picture, as the iCaRL family of models is usually predicting more labels (some of which are incorrect). This behaviour can be observed for the IIRC-CIFAR setup in Figure~\ref{fig:cifar_confusion_icarl} where they tend to predict too many labels incorrectly, which penalize their performance with respect to the PW-JS metric as opposed to the JS metric (see Figure~\ref{fig:cifar_incremental} in the Appendix). We also note that A-GEM model performs poorly in the case of IIRC-CIFAR, even when compared to vanilla ER, and hence we didn't run A-GEM on IIRC-Imagenet.

One thing to notice in Figure~\ref{fig:cifar_incremental_mjaccard}, is the discrepancy between the performance of the ER-infinite baseline and the incremental joint baseline. Recall from section \ref{sec:baselines} that although both baselines don't discard previous tasks samples, incremental joint is using the \textit{complete information} setup, and hence it updates the older samples with the newly learned labels if applicable, while ER-infinite is using the \textit{incomplete information} setup. This result tells us that dealing with the memory constraint is not sufficient by itself for a model to be able to perform well in the IIRC setup.

In lifelong learning setups, the model should retain the previous knowledge as it learns new tasks. Our setup is even more challenging because the model should not only retain previous knowledge, but it should incorporate the new labels as well in this previous knowledge. In Figure~\ref{fig:lite_forgetting} and \ref{fig:cifar_forgetting}, we track how well the model performs on a specific task, as it is trained on subsequent tasks. Unlike the standard class incremental setup, the model should be able to re-associate labels across different tasks to keep performing well on a previous task. The key takeaway is that, while the baselines are generally expected to reasonably alleviate catastrophic forgetting, their performance degrades rapidly as the model trains on more tasks. ER's poor performance may be accounted for by two hypothesis: \textbf{i)} The model is trained on a higher fraction of samples per class for classes that belong to the current task, than those of previous tasks, causing bias towards newer classes. \textbf{ii)} The model sometimes gets conflicting supervising signal, as the model might observe samples that belong to the same subclass (ex. ``polar bear''), once with the superclass label from the buffer (``bear''), and another with the subclass label from the current task data (``polar bear'), and it doesn't connect these two pieces of information together. In the case of LUCIR, we hypothesize that the model's performance deteriorates because the model fails to learn new class labels. We confirm this hypothesis in Figure~\ref{fig:performance_per_task} and Figure~\ref{fig:unified_confusion_matrix_elaborate}, where we observe that while the model is able to retain the labels encountered in the previous tasks, it is not able to learn the labels it encounters during the new tasks. We can see as well in Figure~\ref{fig:performance_per_task} the performance of each model on the current task $j$, after training on that task ($R_{jj}$ using Equation~\ref{equ:r_metric}). The general trend is that the less a model is regularized, the higher it can perform on the current task, which is intuitive.

Some other important questions are whether the model correctly associates the newly learned subclass labels to their previously learned superclass, and whether it incorrectly associates the newly learned subclass label with other previously learned subclasses (that have the same superclass). We dig deeper into the confusion matrix (Figure~\ref{fig:cifar_confusion}) for the predictions of the different models after training on ten tasks of IIRC-CIFAR. Note that in Figure~\ref{fig:cifar_confusion}, the lower triangular matrix shows the percentage the model predicts older labels for the newly introduced classes, while the upper triangular matrix represents the percentage the model predict newer labels to older classes, with the ground truth being Figure~\ref{fig:cifar_confusion_gt}.
The ER method predictions always lie within the newly learned labels (last five classes), as shown in Figure~\ref{fig:cifar_confusion_replay})
The iCaRL-norm model, as shown in Figure~\ref{fig:cifar_confusion_icarl}, performs relatively well in terms of associating (previosuly learned) superclasses to (newly learned) subclasses. For example, whales are always correctly labeled as aquatic mammals, and pickup trucks are correctly labeled as vehicles 94\% of the time. However, these models learn some spurious associations as well. For instance, ``television'' is often mislabeled as ``food containers''. Similarly, the model in general correctly associates newer subclasses with older superclasses, but many times it incorrectly associates the subclasses (eg associating `` aquatic mammals'' with ``whales'' 48\% of the time and ``vehicles'' with ``pickup trucks'' 44\% of the time, while by looking at figure \ref{fig:cifar_confusion_gt}, we see that they only represent 20\% and 12.5\% of their superclasses respectively)
The LUCIR model provides accurate superclass labels to the subclasses. This is shown in Figure~\ref{fig:cifar_confusion_unified} where LUCIR follows the trends of the ground truth more closely than iCaRL-norm in the lower trianglular part of the confusion matrix. However, it fails to learn new associations.
We provide more instances of such plots in the Appendix~\ref{cifar_confusion_over_time}, which shows that the observed trends are quite general. The full resolution figures for Figure~\ref{fig:cifar_confusion} are provided in Appendix~\ref{confusion_full_resolution}.

Finally, we provide some ablations for the effect of the buffer size using ER in Appendix~\ref{buffer_results}. We can see that using ER even with a buffer size of 100 samples per class gives very poor performance in the case of IIRC-ImageNet, and hence a smarter strategy is needed for this setup.

\section{Conclusion}
\label{sec:discussion}
We introduced the ``Incremental Implicitly-Refined Classification (IIRC)'' setup, a novel extension for the class incremental learning setup where incoming batches of classes have labels at different granularity. Our setup enables studying different challenges in the lifelong learning setup that are difficult to study in the existing setups. Moreover, we proposed a standardized benchmark for evaluating the different models on the IIRC setup. We analyze the performance of several well-known lifelong learning models to give a frame of reference for future works and to bring out the strengths and limitations of different approaches. We hope this work will provide a useful benchmark for the community to focus on some important but under-studied problems in lifelong learning. 

\section*{Acknowledgments}
We would like to thank Louis Clouatre and Sanket Vaibhav Mehta for reviewing the paper and for their meaningful feedback. We would like also to thank Louis for suggesting the task name. SC is supported by a Canada CIFAR AI Chair and an NSERC Discovery Grant.

{\small
\bibliographystyle{ieee_fullname}
\bibliography{egbib}
}

\clearpage
\renewcommand\thefigure{A.\arabic{figure}} 
\appendix
\onecolumn
\section*{Appendix}
\newcolumntype{L}{>{\centering\arraybackslash}m{12cm}}

\section{Models Hyperparameter Details}
\label{appendix:hyperparameter_details}
We use SGD as optimizer, as it performs better in the continual learning setups~\cite{mirzadeh2020understanding}, with a momentum value of $0.9$. For the IIRC-CIFAR experiments, The learning rate starts with $1.0$ and is decayed by a factor of $10$ on plateau of the performance of the peri-task validation subset that corresponds to the current task. For the IIRC-ImageNet experiments, the learning rate starts with $0.5$ in the case of iCaRL-CNN and iCaRL-norm, and $0.1$ in the case of finetune, ER and LUCIR, and is decayed by a factor of $10$ on plateau. The number of training epochs per task is $140$ for IIRC-CIFAR and $100$ for IIRC-ImageNet, with the first task always trained for double the number of epochs (due to its larger size). We set the batch size to $128$ and the weight decay parameter to $1e-5$. Moreover, We set the A-GEM memory batch size, which is used to calculate the reference gradient, to $128$. For LUCIR, the margin threshold $m$ is set to $0.5$, and $\lambda_\text{base}$ is set to 5. All the hyperparameters were tuned based on the validation performance in experiments that include only the first four tasks.

During training, data augmentations are applied as follows: for IIRC-ImageNet, a random crop of size ($224\times224$) is sampled from an image, a random horizontal flip is applied, then the pixels are normalized by a pre-calculated per-channel mean and standard deviation. In IIRC-CIFAR, a padding of size $4$ is added to each size, then a random crop of size ($32\times32$) is sampled, a random horizontal flip is applied, then the pixels are normalized.

We keep a fixed number of samples per class in the replay buffer (20, except otherwise indicated). Hence, the capacity increases linearly as the model learns more classes. These samples are chosen randomly, except for iCaRL and LUCIR which use the herding approach. IIRC-CIFAR experiments are averaged over ten task configurations and the each version of IIRC-ImageNet is averaged over $5$ task configurations (see \ref{sec:benchmark} for details).


\section{Baselines Details}
\label{appendix:baselines_details}
Following are three well known baselines that we used to evaluate in the IIRC setup along other baselines including finetune, joint and incremental joint, Vanilla Experience Re-play (ER), and Experience  Replay with infinite buffer (ER-infinite). 
\paragraph{iCaRL:}
iCaRL~\cite{rebuffi2017icarl} was among the first deep learning methods to use exemplar rehearsal in order to alleviate the catastrophic forgetting in the class incremental learning setup. iCaRL model updates the model parameters using the distillation loss, where the outputs of the previous network are used as soft labels for the current network. Moreover, it uses the nearest-mean-of-exemplars classifications (NMC) strategy to classify test samples during inference. Since it is difficult to use NMC when the number of labels is variable (not a single label setup), we use the classification layer used during training during inference as well.
\paragraph{Unified learning via rebalancing (LUCIR):}
LUCIR~\cite{hou2019learning} is a class incremental method that exploits three components to alleviate the catastrophic forgetting and reduce the negative effect of the imbalance between the old and new classes, since the number of samples in the replay buffer is much less than the current task samples. LUCIR uses the cosine normalization to get balanced magnitudes for classes seen so far. It also uses the less forget constraint, where the distillation loss is applied in the feature space instead of the output space, and a margin ranking loss to ensure interclass separation.

\paragraph{A-GEM:} 
A-GEM~\cite{chaudhry2019efficient} is an improved version of GEM that is a constrained optimization method in the Replay-based approach. GEM uses memory to constrain gradients so as to update the model parameters to not interfere with previous tasks. GEM is a very computationally expensive approach that is not applicable to the large-scale setup. Hence, Averaged GEM (A-GEM) provides an efficient version of GEM, where it only requires computing the gradient on a random subset of memory examples, and it does not need as well to solve any quadratic program but just an inner product. Since A-GEM is a very well known constrained optimization method that has reasonable guarantees in terms of average accuracy in comparison to GEM, we selected it as a candidate to evaluate its performance in our more realistic large scale setting.

\clearpage

\section{IIRC Dataset Statistics}
\label{appendix:dataset_statistics}

\begin{table}[!ht]
\centering
\begin{tabular}{||c | c | c | c | c | c | c||} 
 \hline
  & \multicolumn{2}{c|}{with duplicates} & \multicolumn{2}{c|}{without duplicates} &  &  \\
  \hline
 dataset & train & in-task validation & train & in-task validation & post-task validation & test \\ [0.5ex] 
 \hline\hline
 IIRC-CIFAR & 46,160 & 5,770 & 40,000 & 5,000 & 5,000 & 10,000 \\ [0.5ex] 
 \hline
 IIRC-Imagenet-full & 1,215,123 & 51,873 & 1,131,966 & 48,802 & 51,095 & 49,900 \\ [1ex] 
 \hline
\end{tabular}
\caption{The number of samples for each split of the training set. with duplicates represents the number of samples including the dublicates between some superclasses and their subclasses (the samples that the model see two times with two different labels). This doesn't happen for the post-task validation set and test set as they are in the complete information setup}
\label{table:splits}
\end{table}

\begin{table}[!ht]
\centering
\begin{tabular}{||c | c | c | c | c||} 
 \hline
 dataset & superclasses & subclasses (under superclasses) & subclasses (with no superclasses) & total \\ [0.5ex] 
 \hline\hline
 IIRC-CIFAR & 15 & 77 & 23 & 115\\ [0.5ex] 
 \hline
 IIRC-Imagenet-full & 85 & 788 & 210 & 1083\\ [1.0ex] 
 \hline
\end{tabular}
\caption{For each dataset, these are the number of superclasses, the number of subclasses that belong to these superclasses, the number of subclasses that don't have a superclass, and the total number of superclasses and subclasses}
\label{table:superclass_subclass_stats}
\end{table}

\begin{table}[!ht]
\centering
\begin{tabular}{||c | c c c | c c||} 
 \hline
 dataset & superclass & num of subclasses & superclass size & subclass & subclass size \\ [0.5ex] 
 \hline\hline
 IIRC-CIFAR & vehicles & 8 & 1,280 & bus & 320 \\ [0.5ex] 
 IIRC-CIFAR & small mammals & 5 & 800 & squirrel & 320 \\ [0.5ex] 
 IIRC-CIFAR & - & - & - & mushroom & 400 \\ [0.5ex] 
 \hline
 IIRC-ImageNet & bird & 58 & 3,762 & ostrich & 956 \\ [0.5ex] 
 IIRC-ImageNet & big cat & 6 & 2,868 & leopard & 956 \\ [0.5ex] 
 IIRC-ImageNet & keyboard instrument & 4 & 1,912 & grand piano & 956 \\ [0.5ex] 
 IIRC-ImageNet & - & - & - & wooden spoon & 1,196 \\ [1.0ex] 
 \hline
\end{tabular}
\caption{Several examples for classes and the number of samples they have in the training set. The subclass on the right is a subclass that belongs to the superclass on the left. The left side is blank for subclasses that have no superclasses.}
\label{table:samples_per_class_example}
\end{table}

\clearpage

\section{More Figures}
\begin{figure}[!htb]
\begin{center}
\subfigure{
\includegraphics[clip,trim=3.5cm 1.5cm 4.5cm 2.75cm, width=0.48\linewidth]{latex/figures/cifar/average_modified_jaccard.pdf}
\label{fig:cifar_incremental_mjaccard2}
}
\subfigure{
\includegraphics[clip, trim=3.5cm 1.5cm 4.5cm 2.75cm, width=0.48\linewidth]{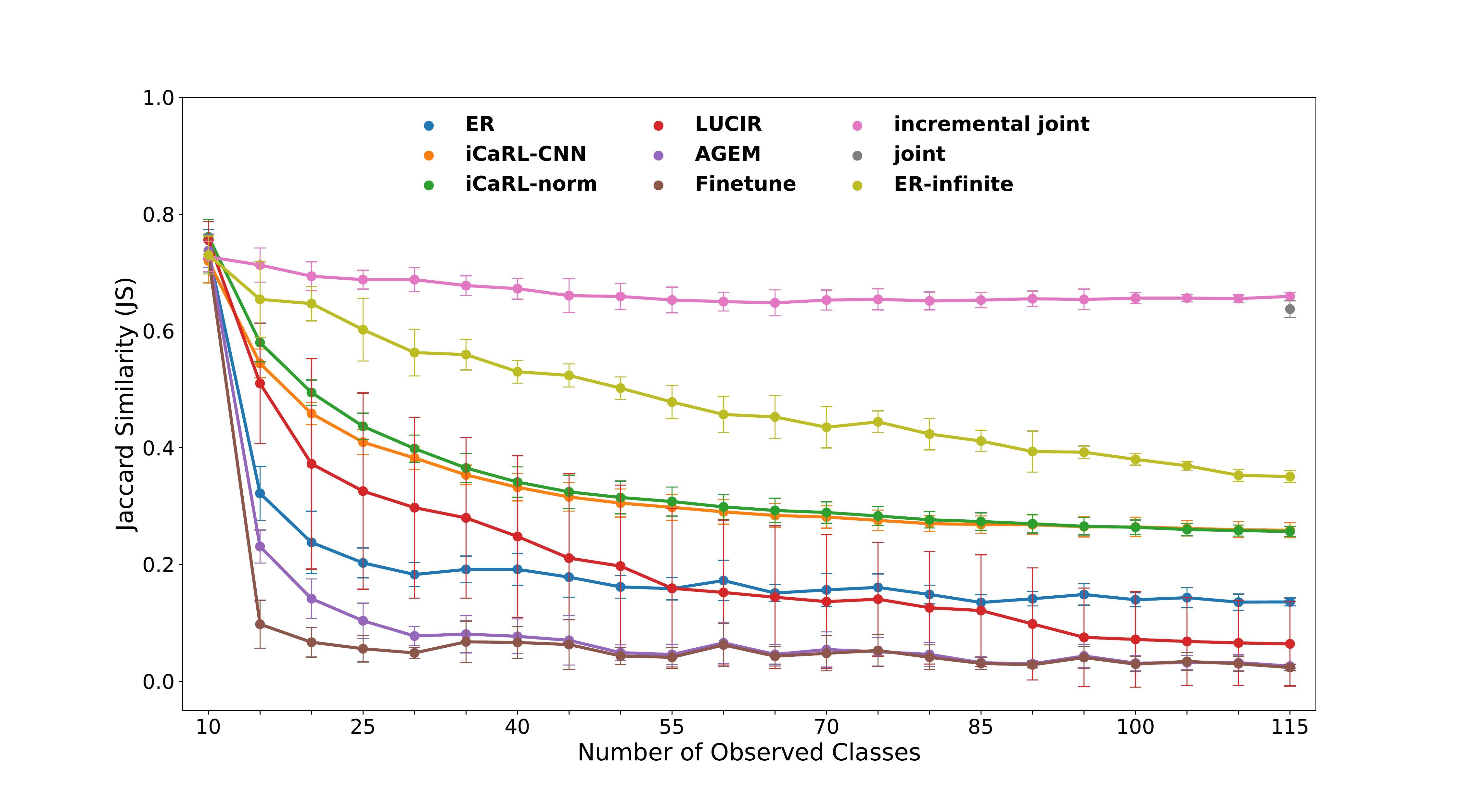}
\label{fig:cifar_incremental_jaccard2}
}
\end{center}
\caption{Average performance on IIRC-CIFAR. (left) the precision-weighted Jaccard Similarity and (right) the Jaccard Similarity. We run experiments ten times using ten different task configurations to report the performance with the mean and the standard deviation.}
\label{fig:cifar_incremental}
\end{figure}

\begin{figure}[!htb]
\begin{center}
\subfigure[Task 0]{
\includegraphics[clip,trim=3.5cm 1.5cm 4.5cm 2.75cm, width=0.48\linewidth]{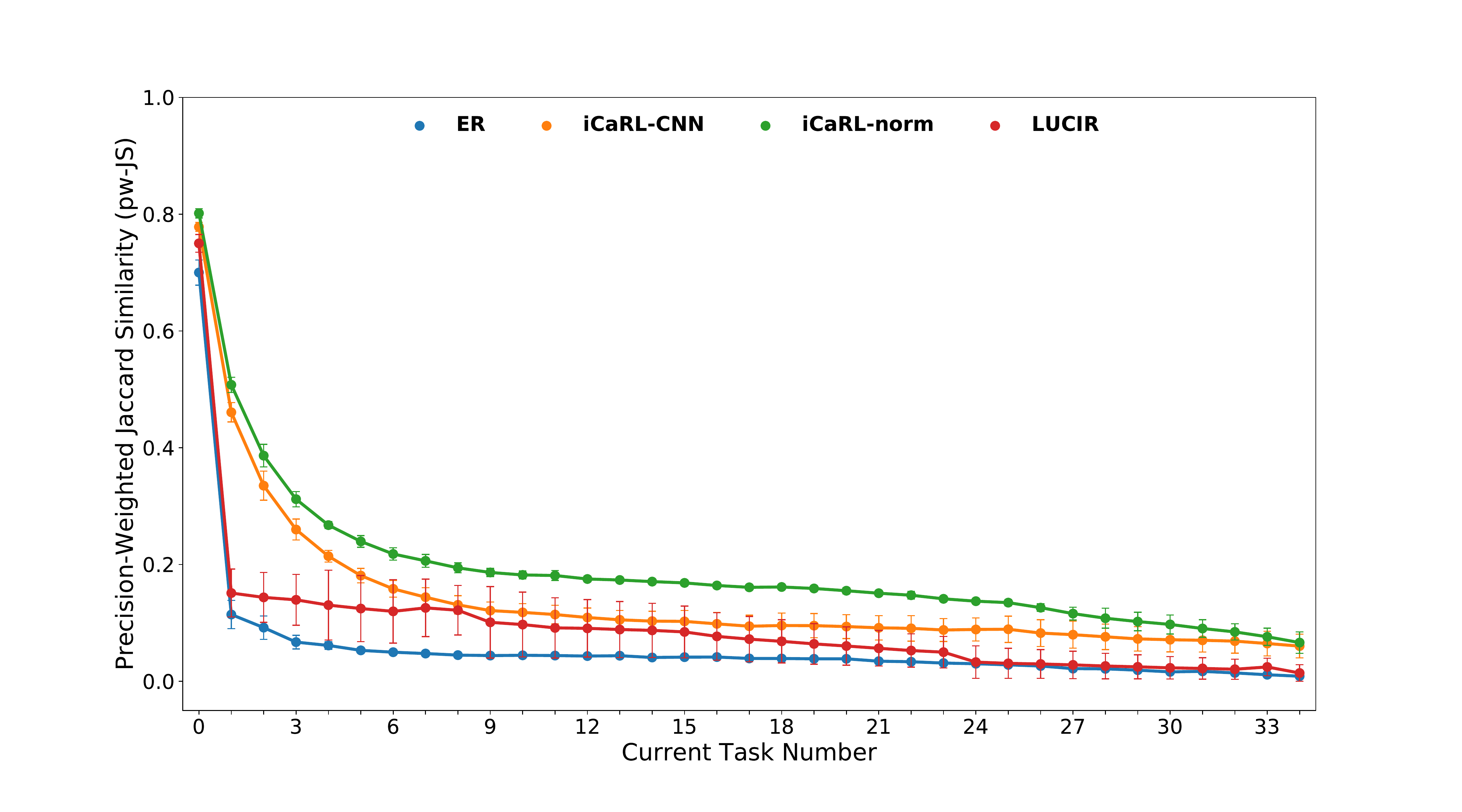}
\label{fig:lite_task0_forgetting}
}
\subfigure[Task 5]{
\includegraphics[clip,trim=3.5cm 1.5cm 4.5cm 2.75cm, width=0.48\linewidth]{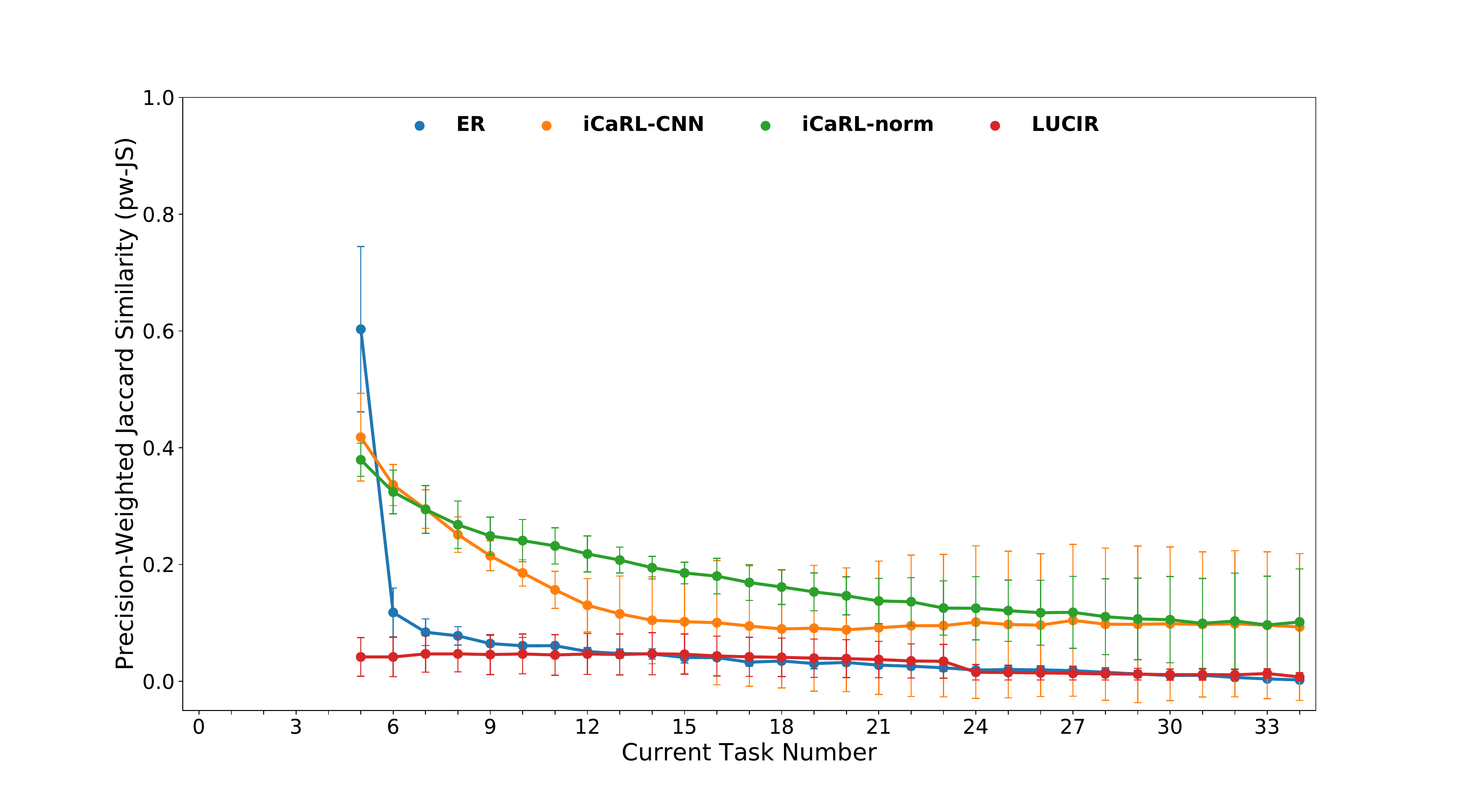}
}
\subfigure[Task 10]{
\includegraphics[clip,trim=3.5cm 1.5cm 4.5cm 2.75cm, width=0.48\linewidth]{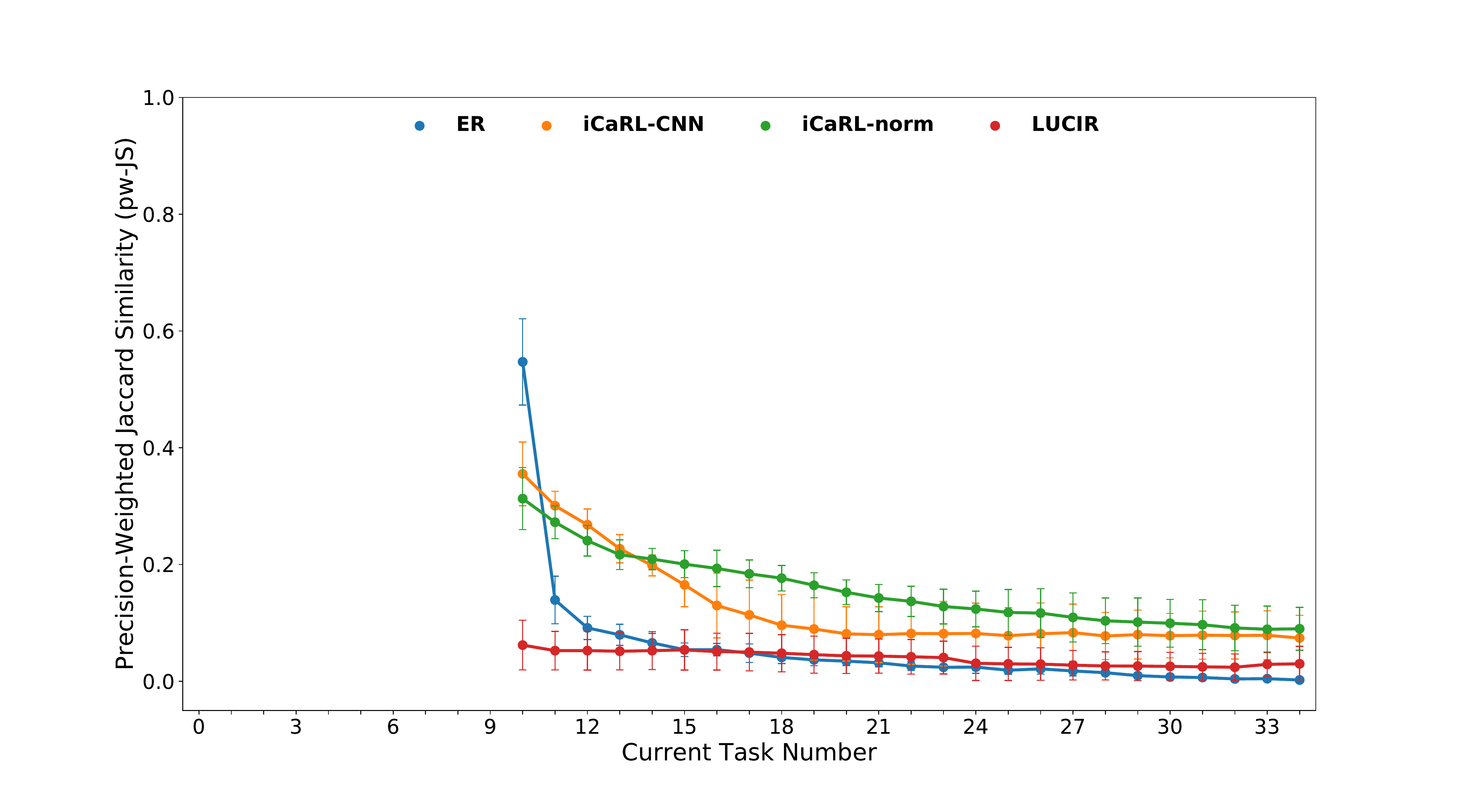}
}
\subfigure[Task 15]{
\includegraphics[clip,trim=3.5cm 1.5cm 4.5cm 2.75cm, width=0.48\linewidth]{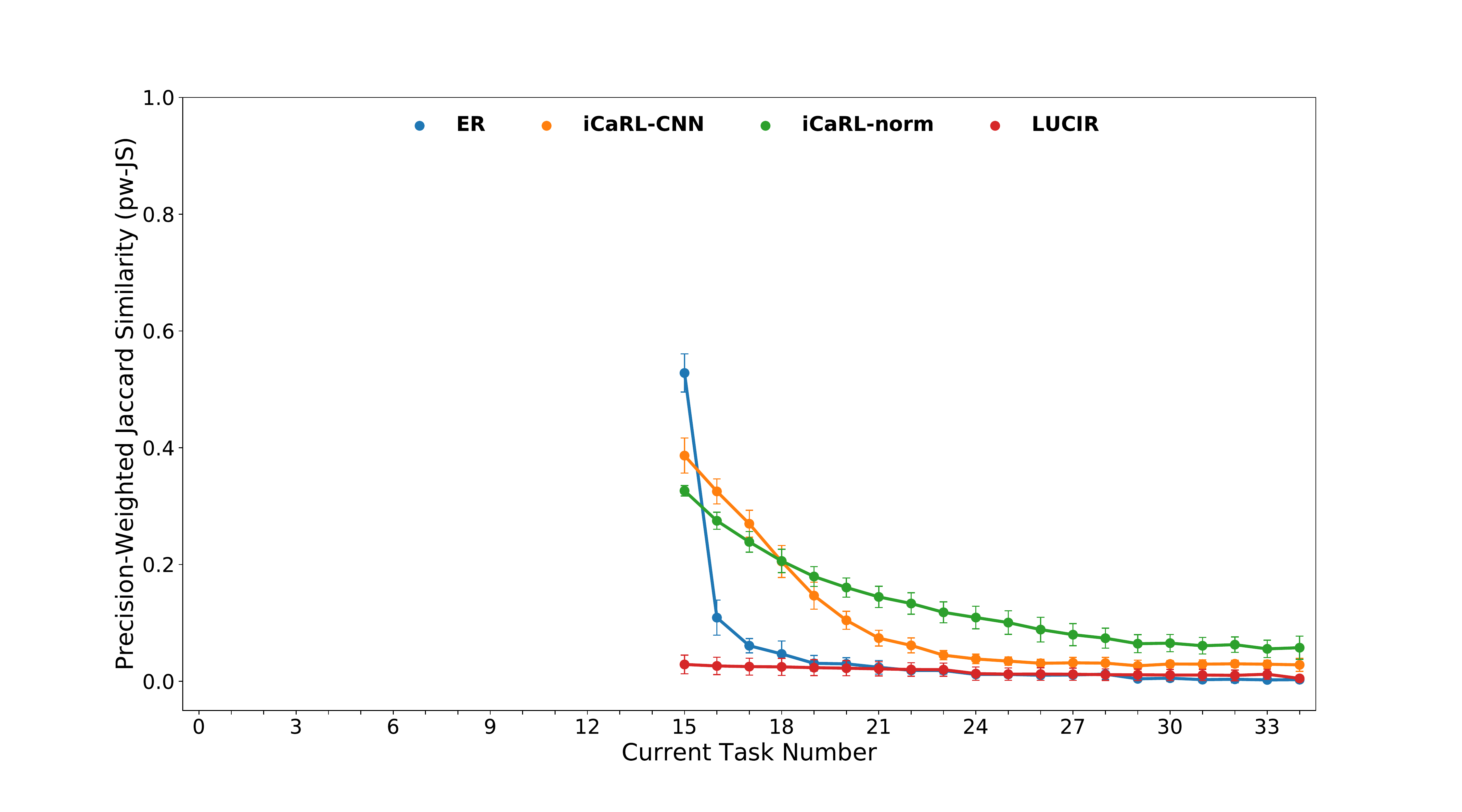}
}
\end{center}
\caption{IIRC-ImageNet-full Performance on four middle tasks throughout the whole training process, to measure their catastrophic forgetting and backward transfer.  Note that a degradation in performance is not necessarily caused by catastrophic forgetting, as a new subclass of a previously observed superclass might be introduced and the model would be penalized for not applying that label retroactively. We run experiments on ten different task configurations and report the performance with the mean and the standard deviation.}
\label{fig:lite_forgetting}
\end{figure}

\begin{figure}[!htb]
\begin{center}
\subfigure[Task 0]{
\includegraphics[clip,trim=3.5cm 1.5cm 4.5cm 2.75cm, width=0.48\linewidth]{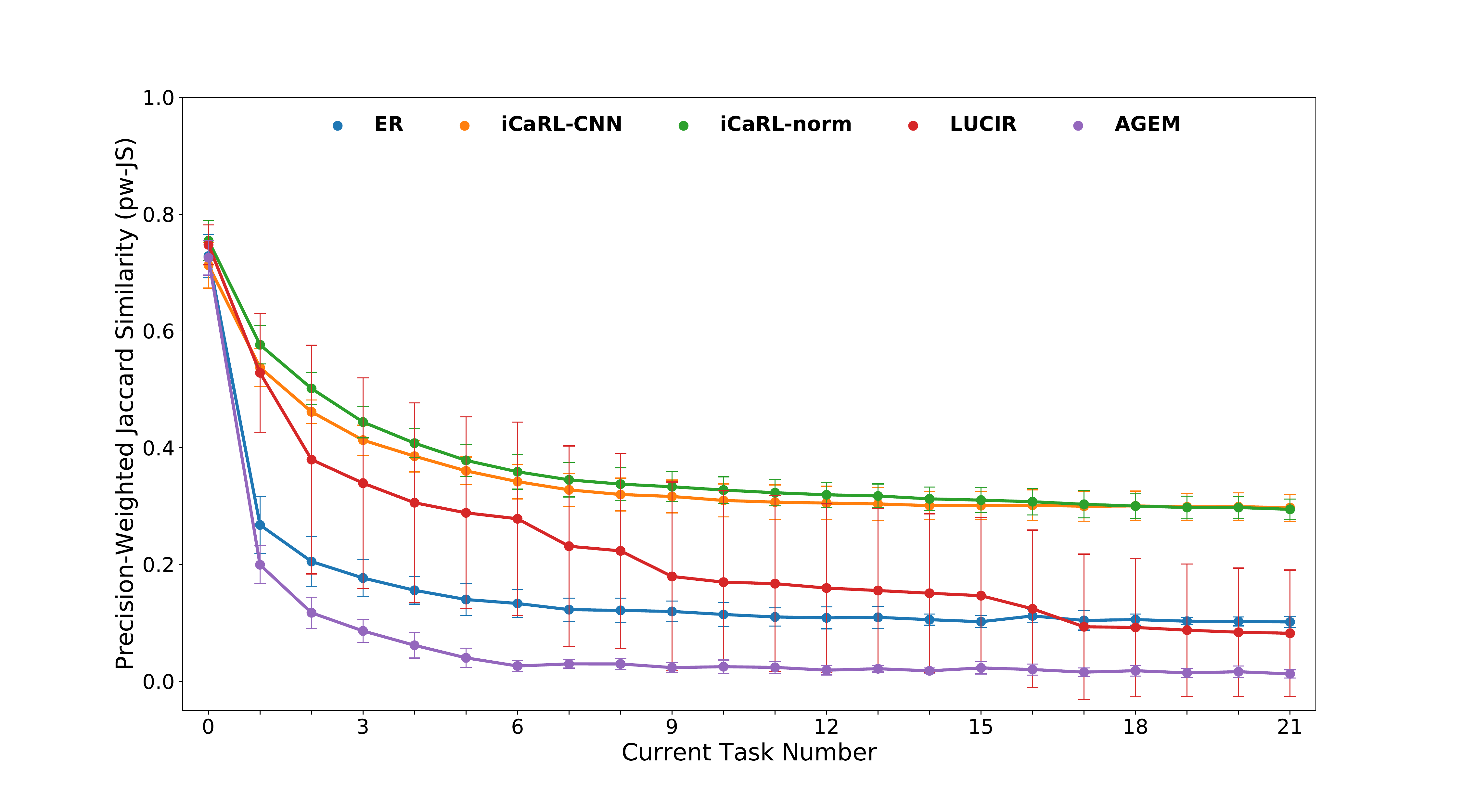}
\label{fig:cifar_task0_forgetting}
}
\subfigure[Task 5]{
\includegraphics[clip,trim=3.5cm 1.5cm 4.5cm 2.75cm, width=0.48\linewidth]{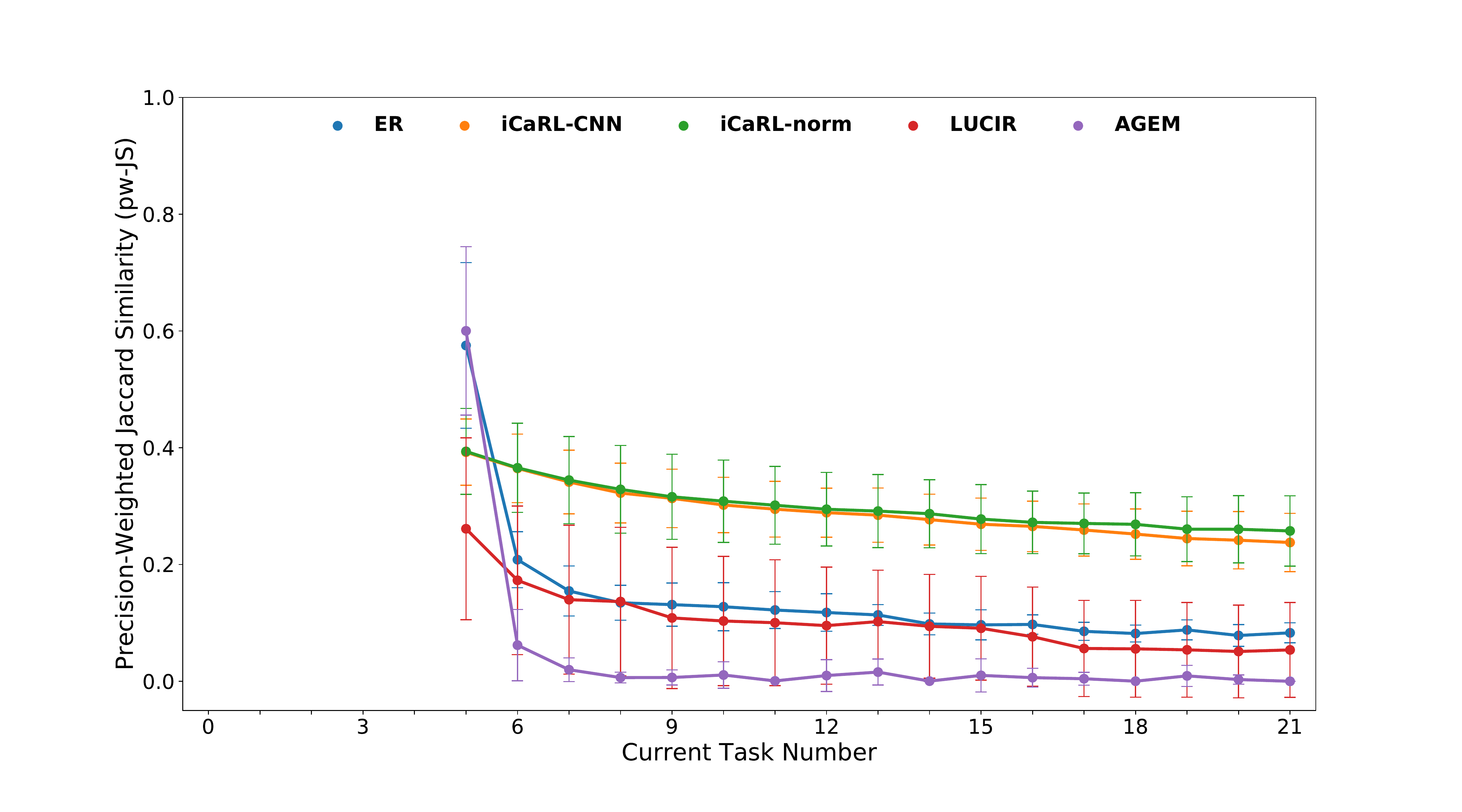}
}
\subfigure[Task 10]{
\includegraphics[clip,trim=3.5cm 1.5cm 4.5cm 2.75cm, width=0.48\linewidth]{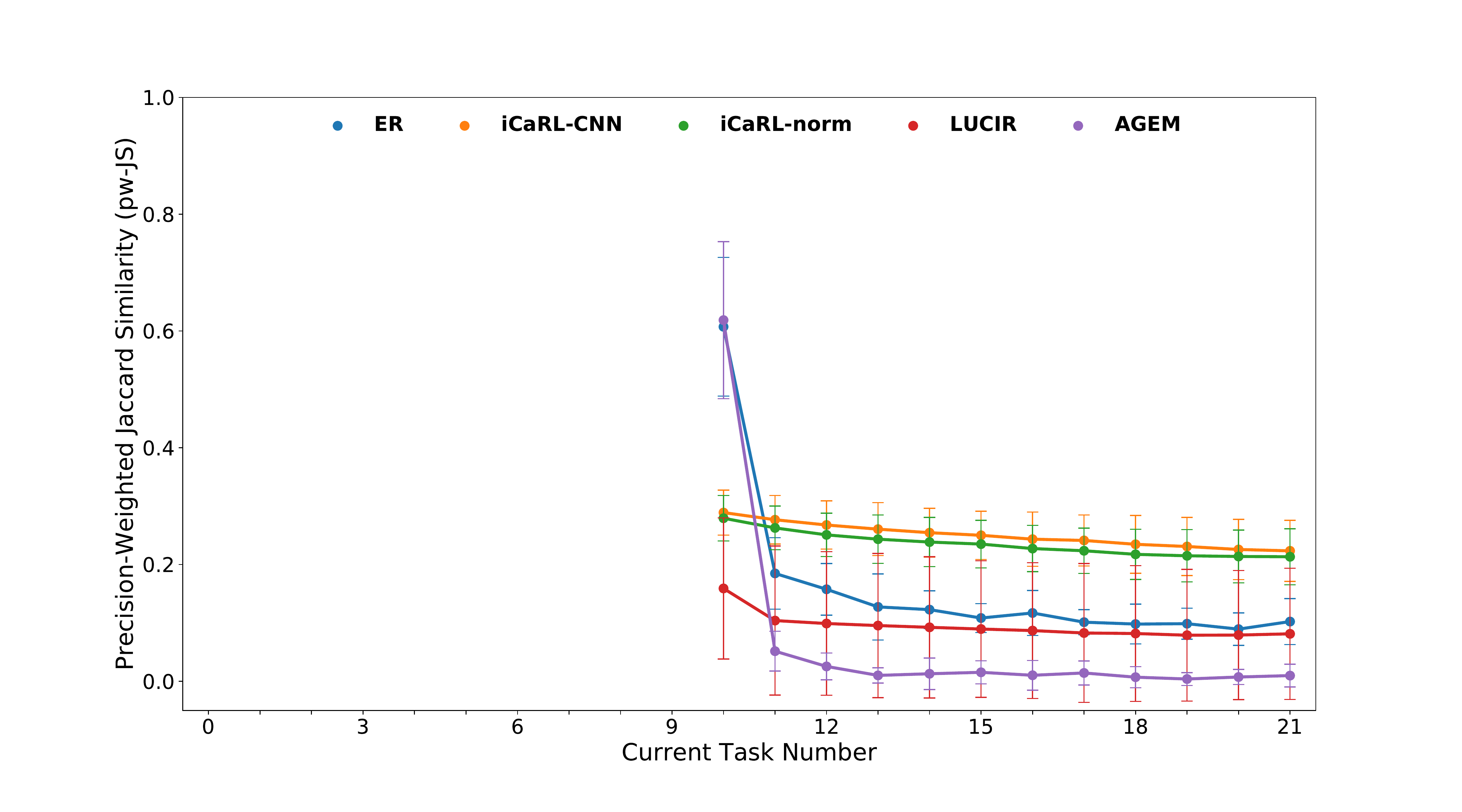}
}
\subfigure[Task 15]{
\includegraphics[clip,trim=3.5cm 1.5cm 4.5cm 2.75cm, width=0.48\linewidth]{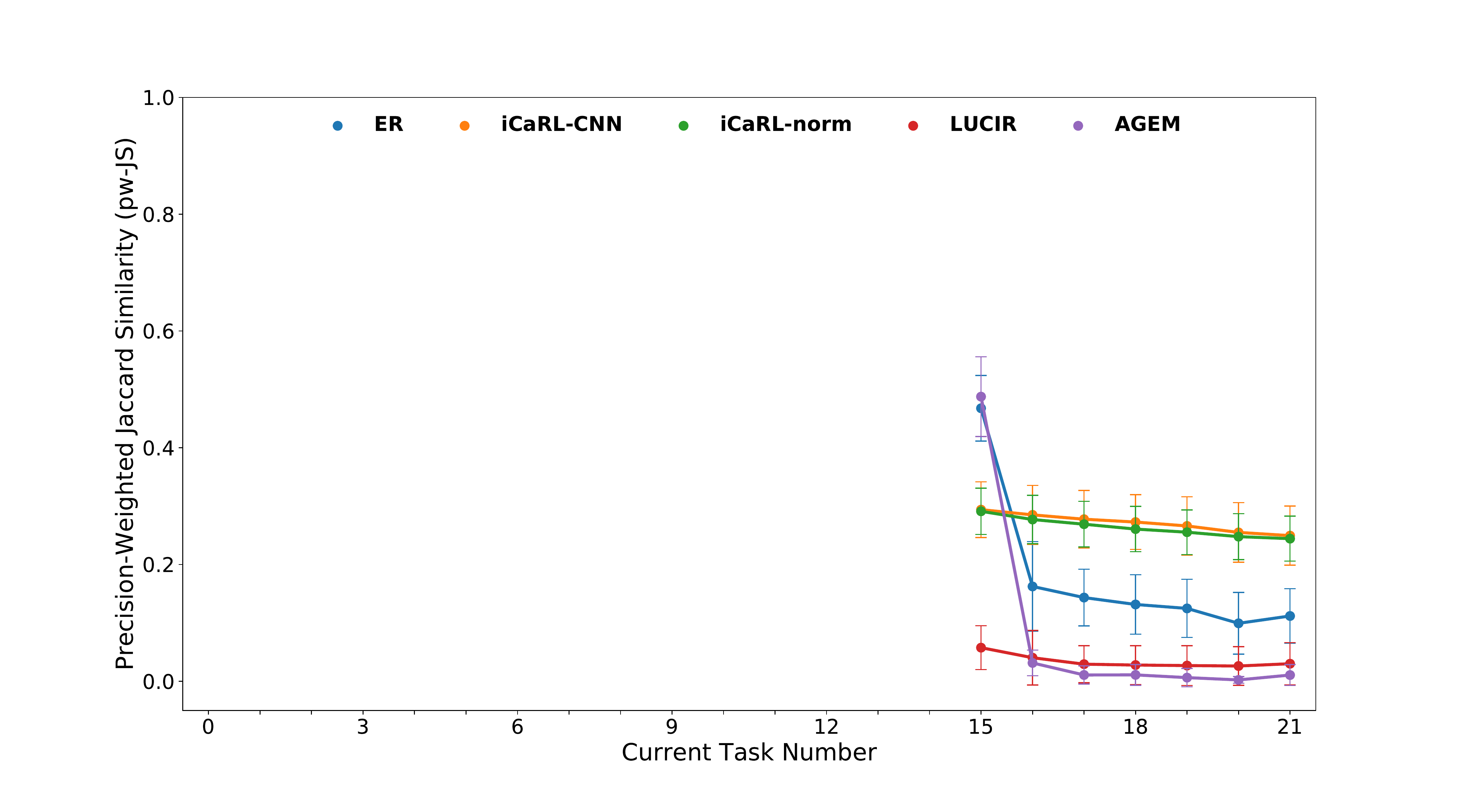}
}
\end{center}
\caption{IIRC-CIFAR Performance on four middle tasks throughout the whole training process, to measure their catastrophic forgetting and backward transfer.  Note that a degradation in performance is not necessarily caused by catastrophic forgetting, as a new subclass of a previously observed superclass might be introduced and the model would be penalized for not applying that label retroactively. We run experiments on ten different task configurations and report the performance with the mean and the standard deviation.}
\label{fig:cifar_forgetting}
\end{figure}
\clearpage
\subsection{Confusion Matrix Over time}
\label{cifar_confusion_over_time}

\begin{figure}[!ht]
\begin{center}
\subfigure[After task 0]{
\includegraphics[width=0.23\linewidth]{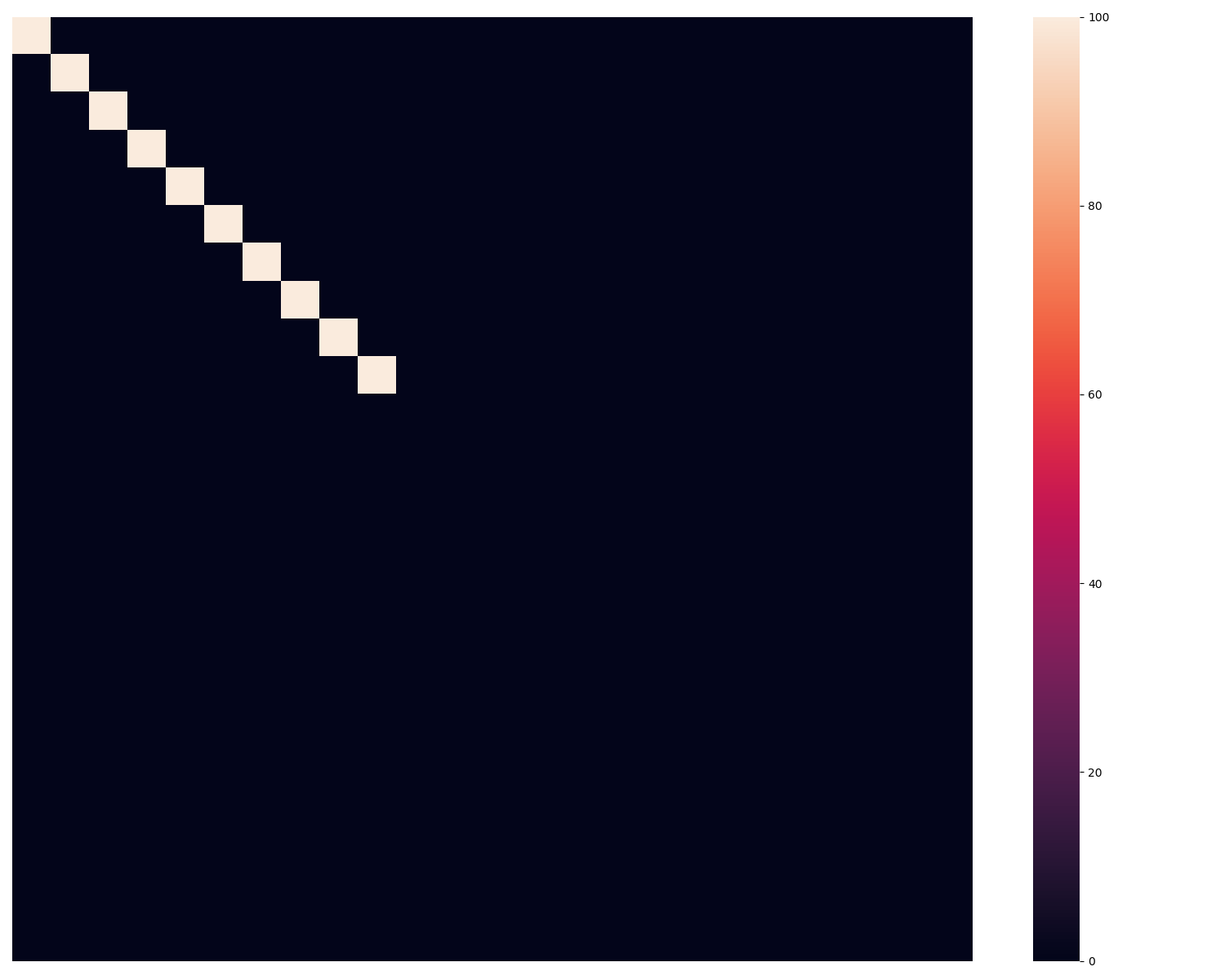}
}
\subfigure[After task 1]{
\includegraphics[width=0.23\linewidth]{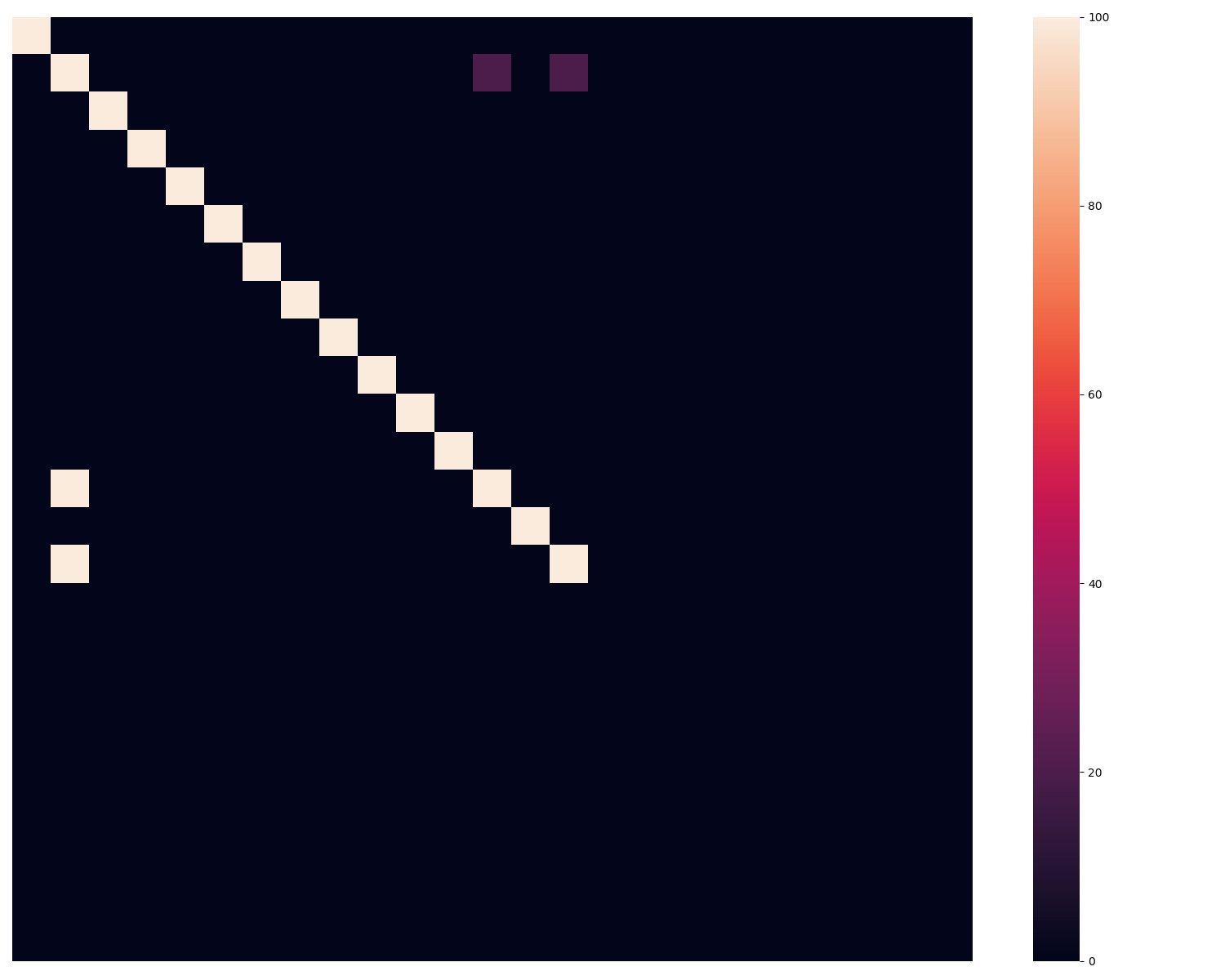}
}
\subfigure[After task 5]{
\includegraphics[width=0.23\linewidth]{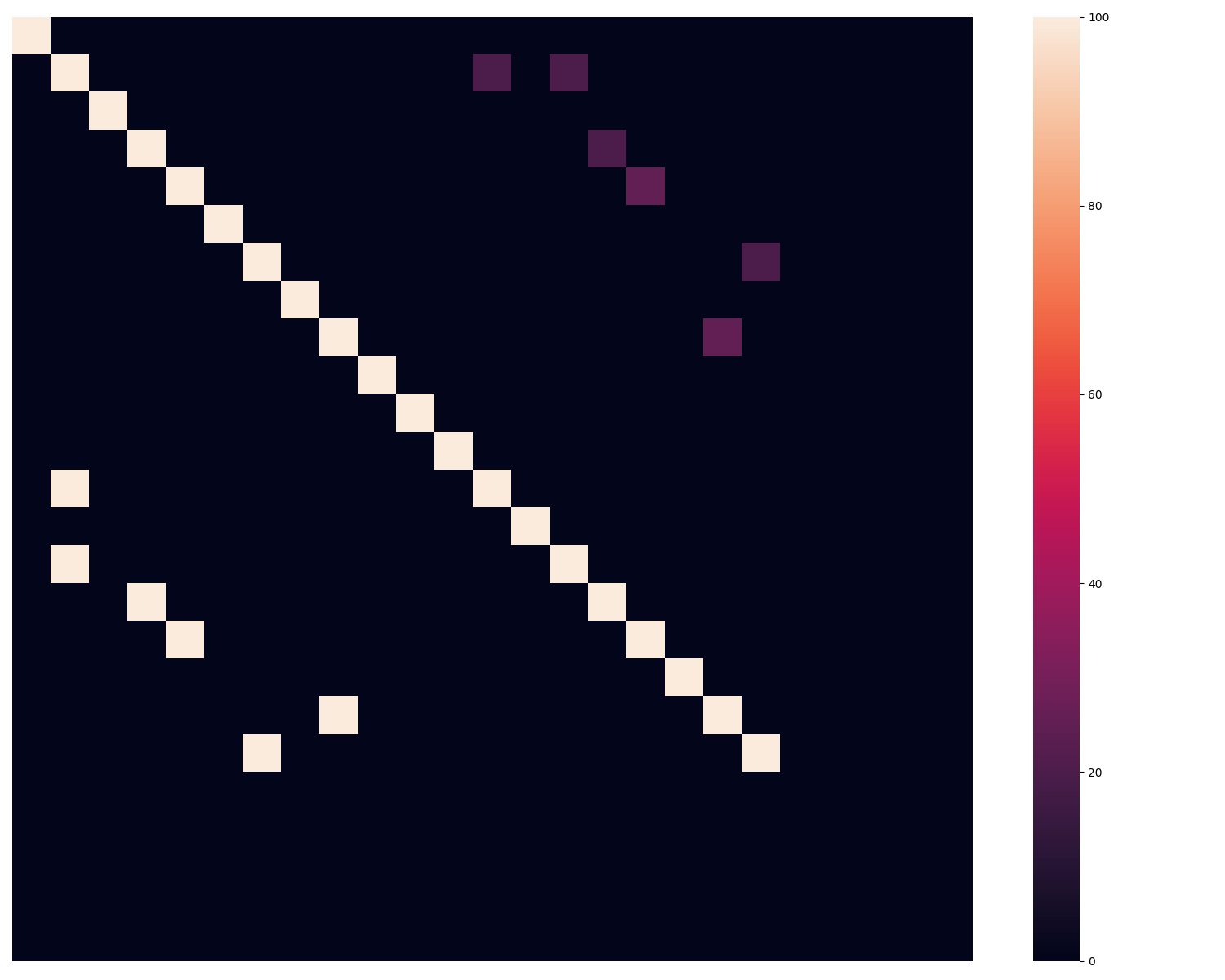}
}
\subfigure[After task 10]{
\includegraphics[width=0.23\linewidth]{latex/figures/confusion_matrices/ground_truth/task_10_confusion.png}
}
\end{center}
\caption{Ground Truth confusion matrix after introducing tasks 0, 1, 5, 10 of IIRC-CIFAR respectively. The y-axis is the correct label (or one of the correct labels). The x-axis is the model predicted labels. Labels are arranged by their order of introduction. Only 25 labels are shown for better visibility.}
\label{fig:gt_confusion_matrix_elaborate}
\end{figure}

\begin{figure}[!ht]
\begin{center}
\subfigure[After task 0]{
\includegraphics[width=0.23\linewidth]{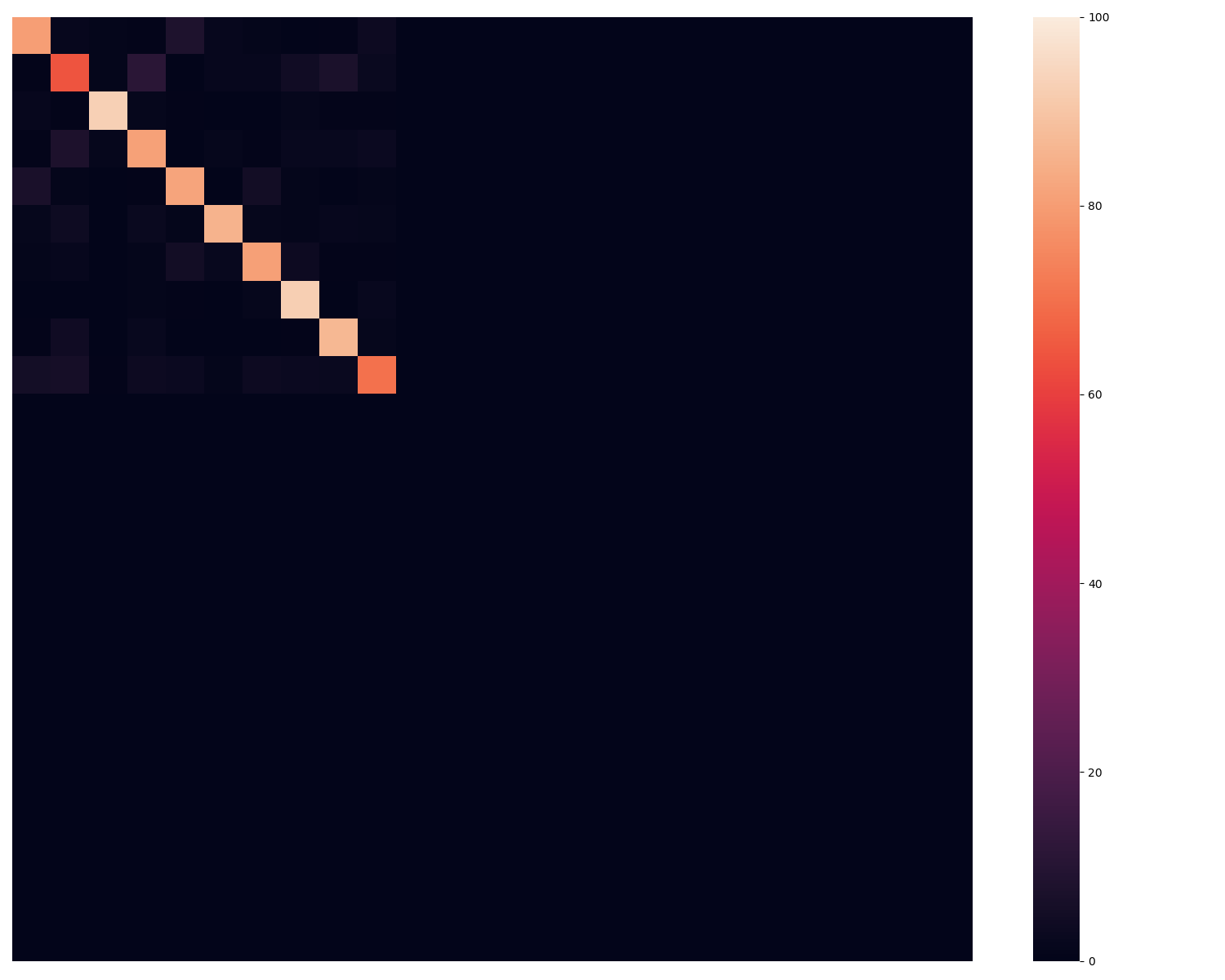}
}
\subfigure[After task 1]{
\includegraphics[width=0.23\linewidth]{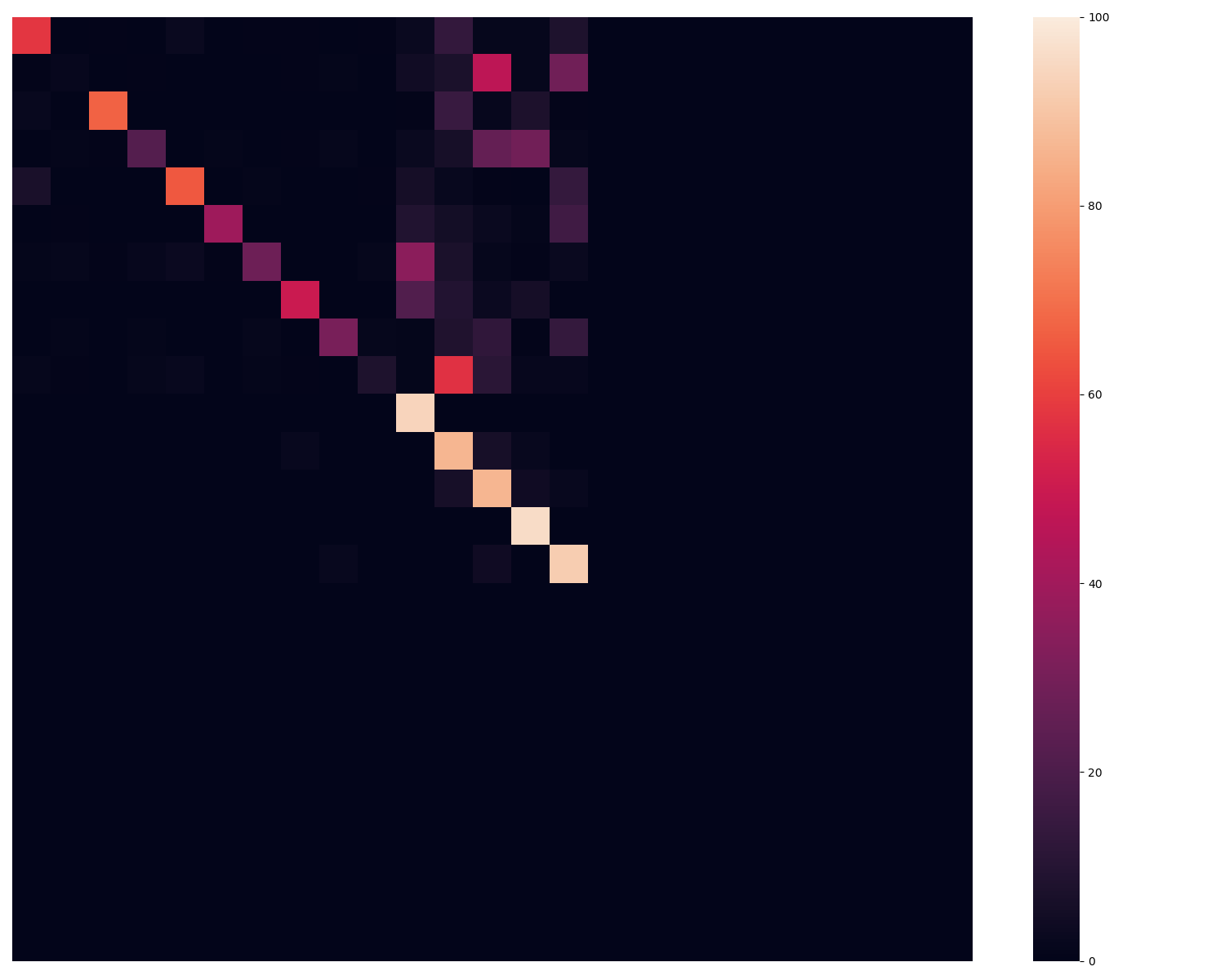}
}
\subfigure[After task 5]{
\includegraphics[width=0.23\linewidth]{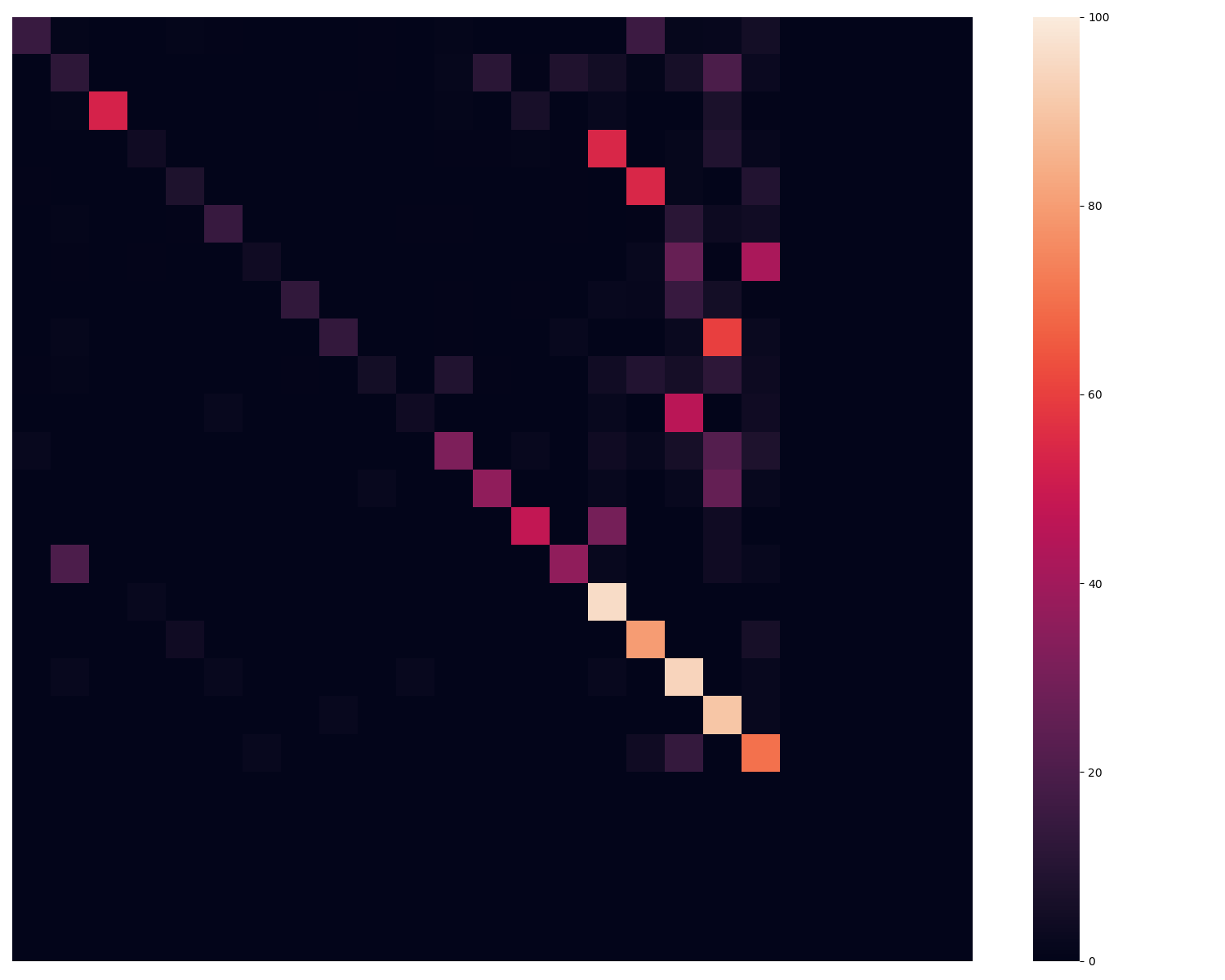}
}
\subfigure[After task 10]{
\includegraphics[width=0.23\linewidth]{latex/figures/confusion_matrices/finetune_20/task_10_confusion.png}
}
\end{center}
\caption{ER confusion matrix after introducing tasks 0, 1, 5, 10 of IIRC-CIFAR respectively. The y-axis is the correct label (or one of the correct labels). The x-axis is the model predicted labels. Labels are arranged by their order of introduction. Only 25 labels are shown for better visibility.}
\label{fig:finetune_confusion_matrix_elaborate}
\end{figure}

\begin{figure}[!ht]
\begin{center}
\subfigure[After task 0]{
\includegraphics[width=0.23\linewidth]{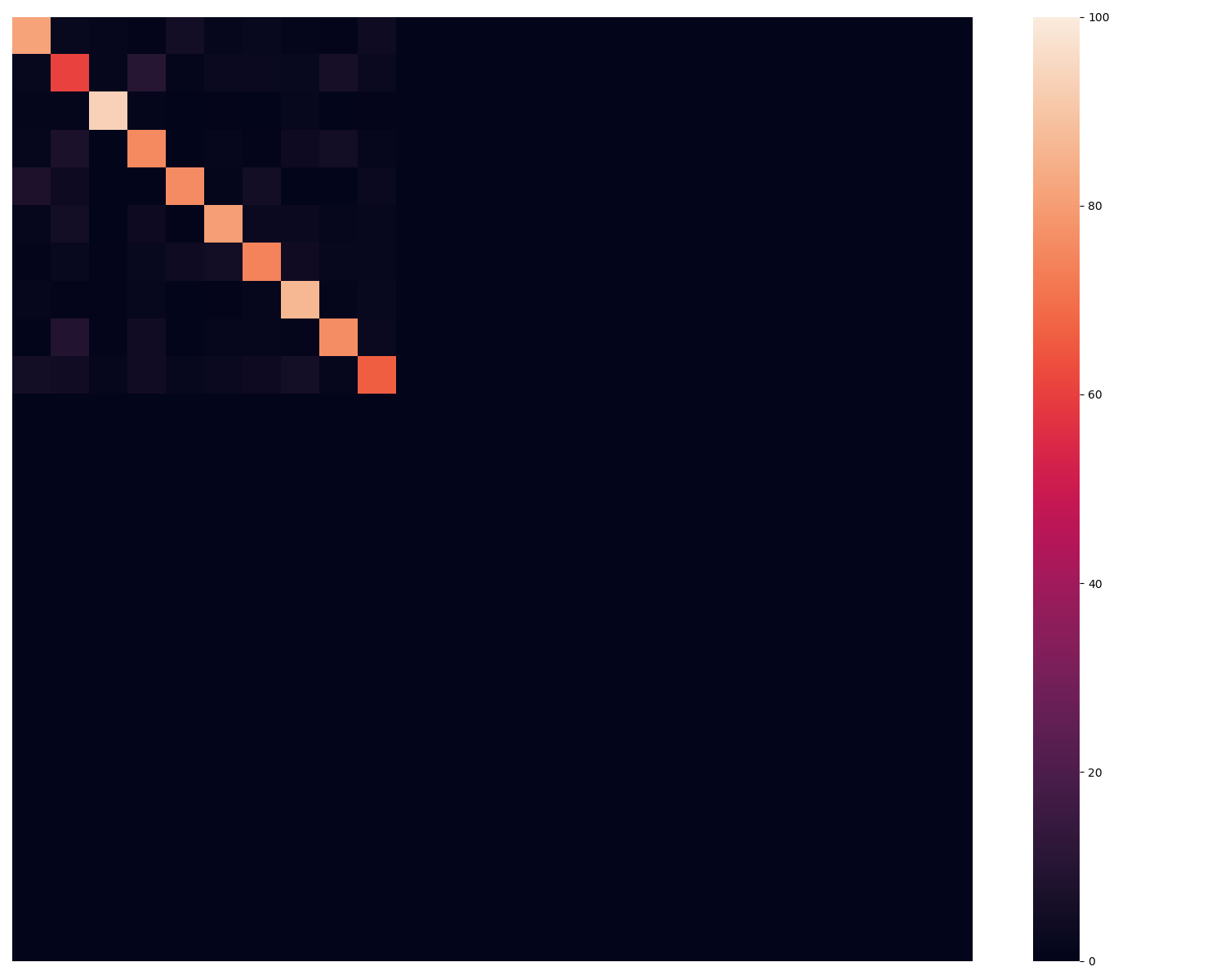}
}
\subfigure[After task 1]{
\includegraphics[width=0.23\linewidth]{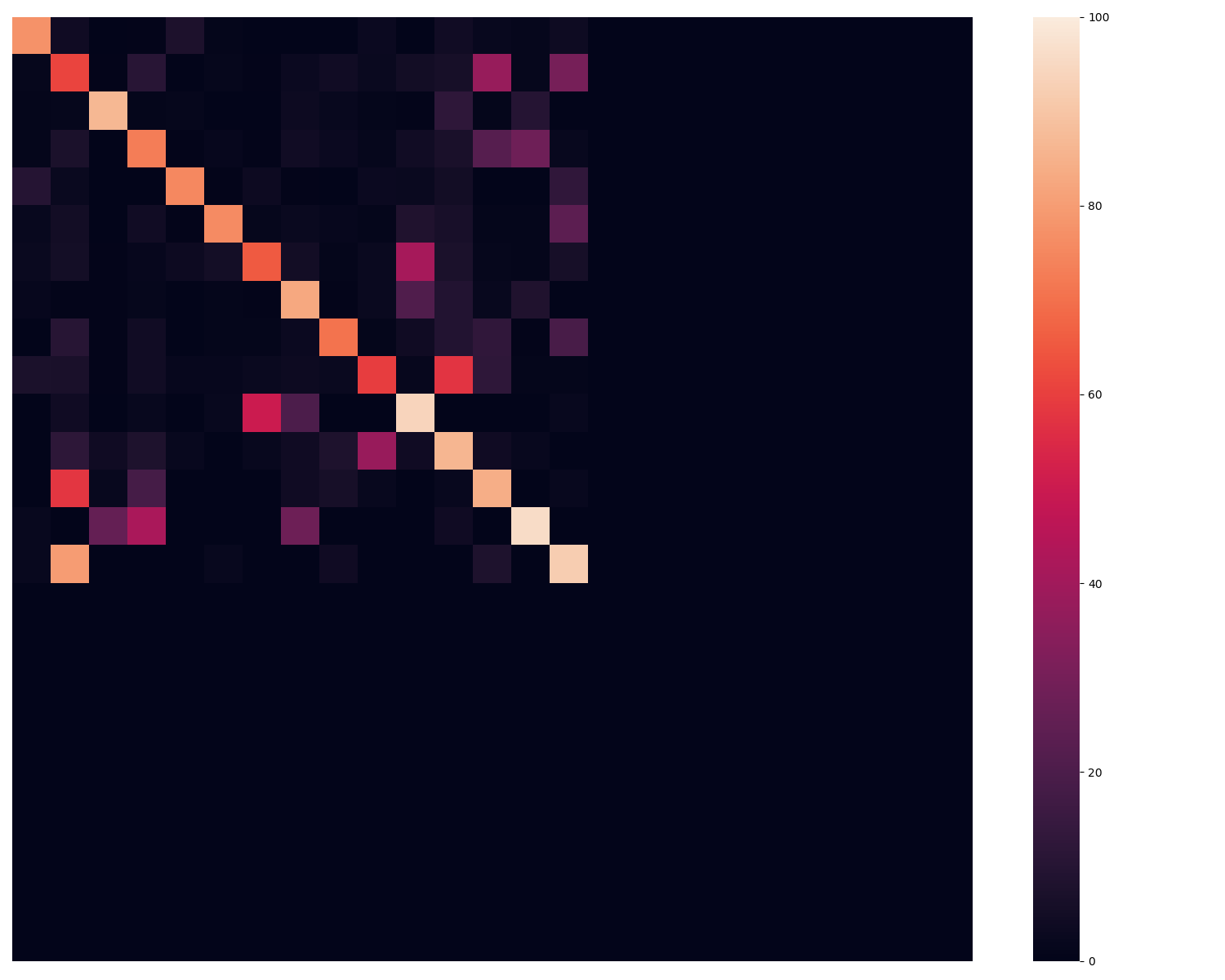}
}
\subfigure[After task 5]{
\includegraphics[width=0.23\linewidth]{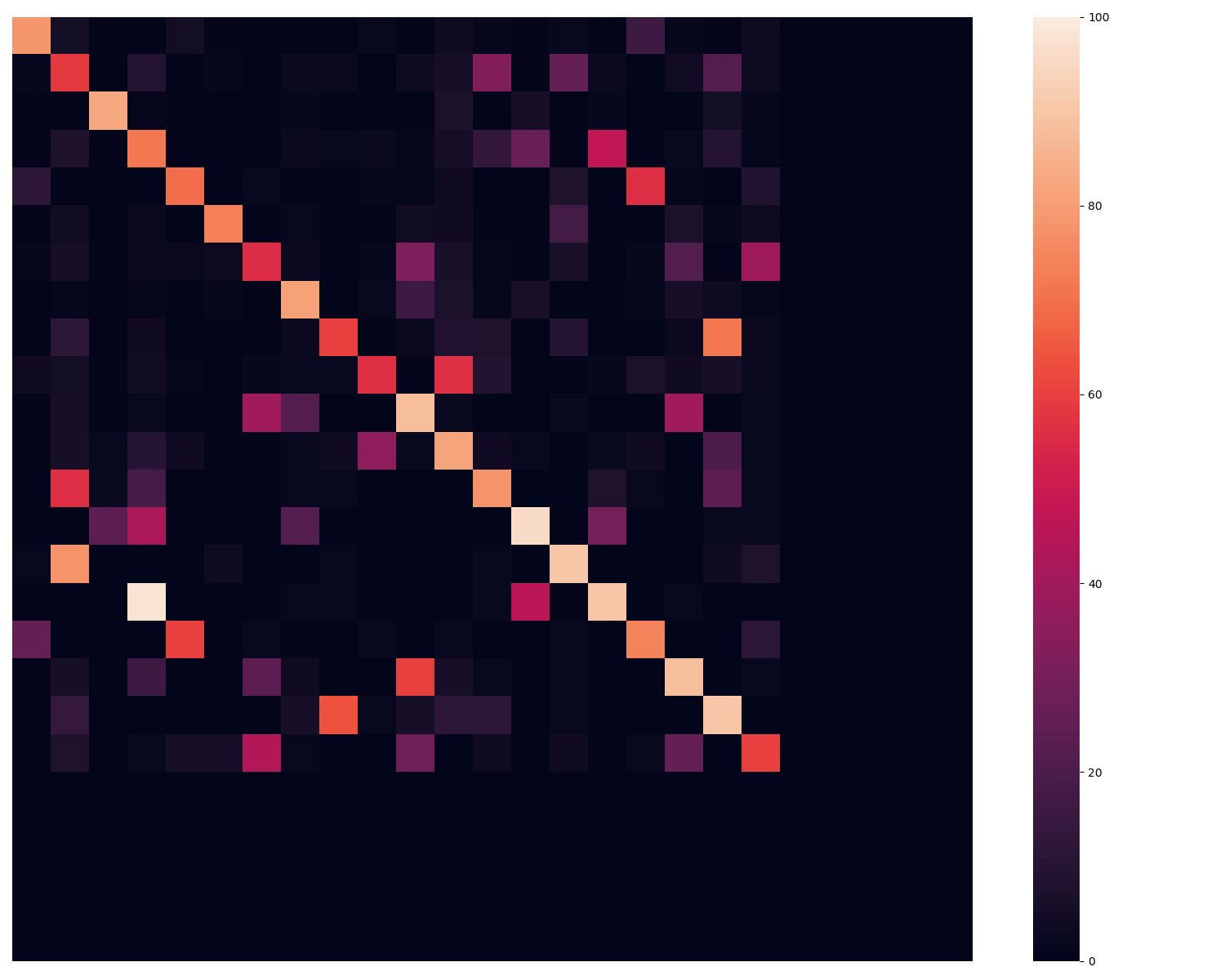}
}
\subfigure[After task 10]{
\includegraphics[width=0.23\linewidth]{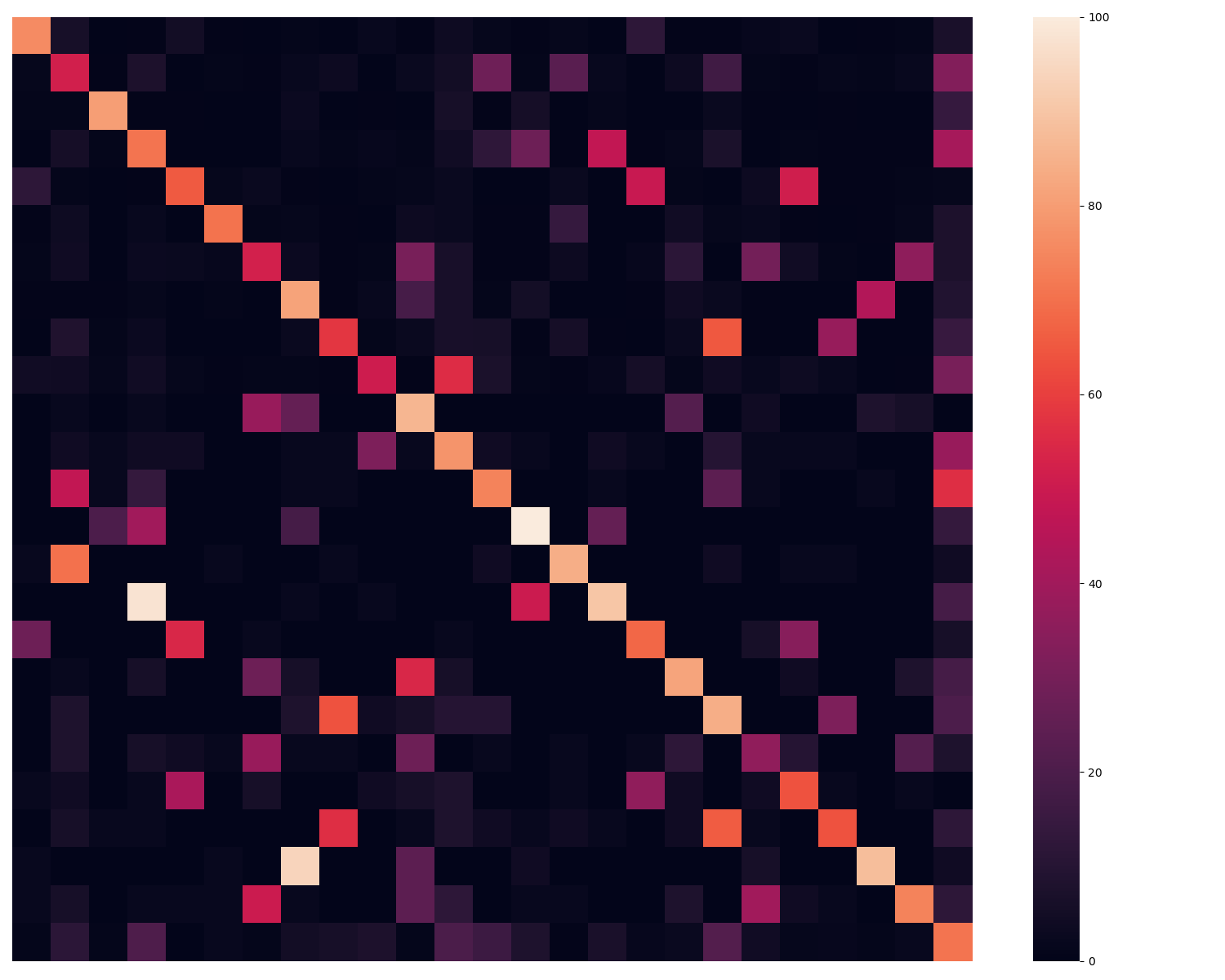}
}
\end{center}
\caption{iCaRL-CNN confusion matrix after introducing tasks 0, 1, 5, 10 of IIRC-CIFAR respectively. The y-axis is the correct label (or one of the correct labels). The x-axis is the model predicted labels. Labels are arranged by their order of introduction. Only 25 labels are shown for better visibility.}
\label{fig:icarl_cnn_confusion_matrix_elaborate}
\end{figure}

\begin{figure}[!ht]
\begin{center}
\subfigure[After task 0]{
\includegraphics[width=0.23\linewidth]{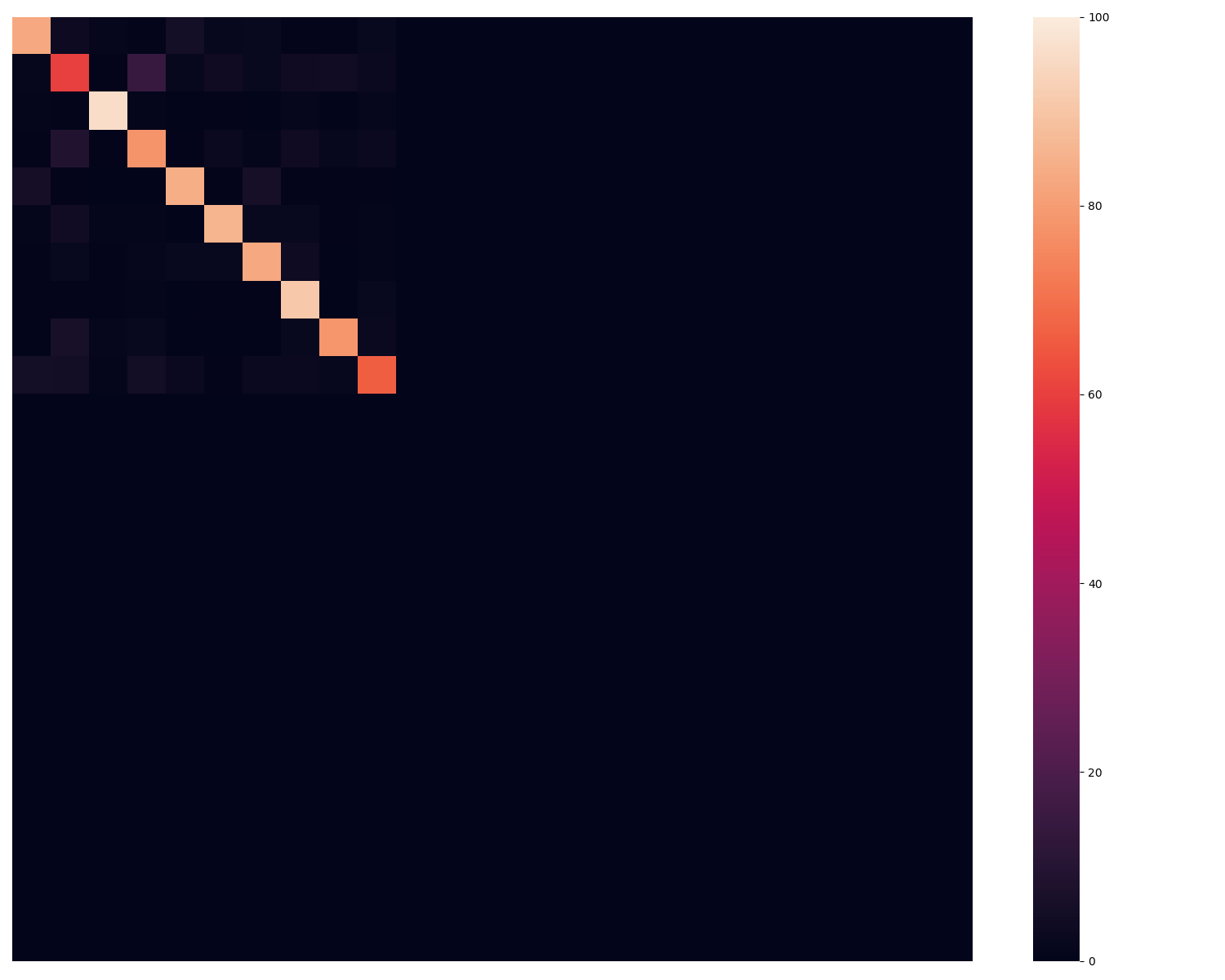}
}
\subfigure[After task 1]{
\includegraphics[width=0.23\linewidth]{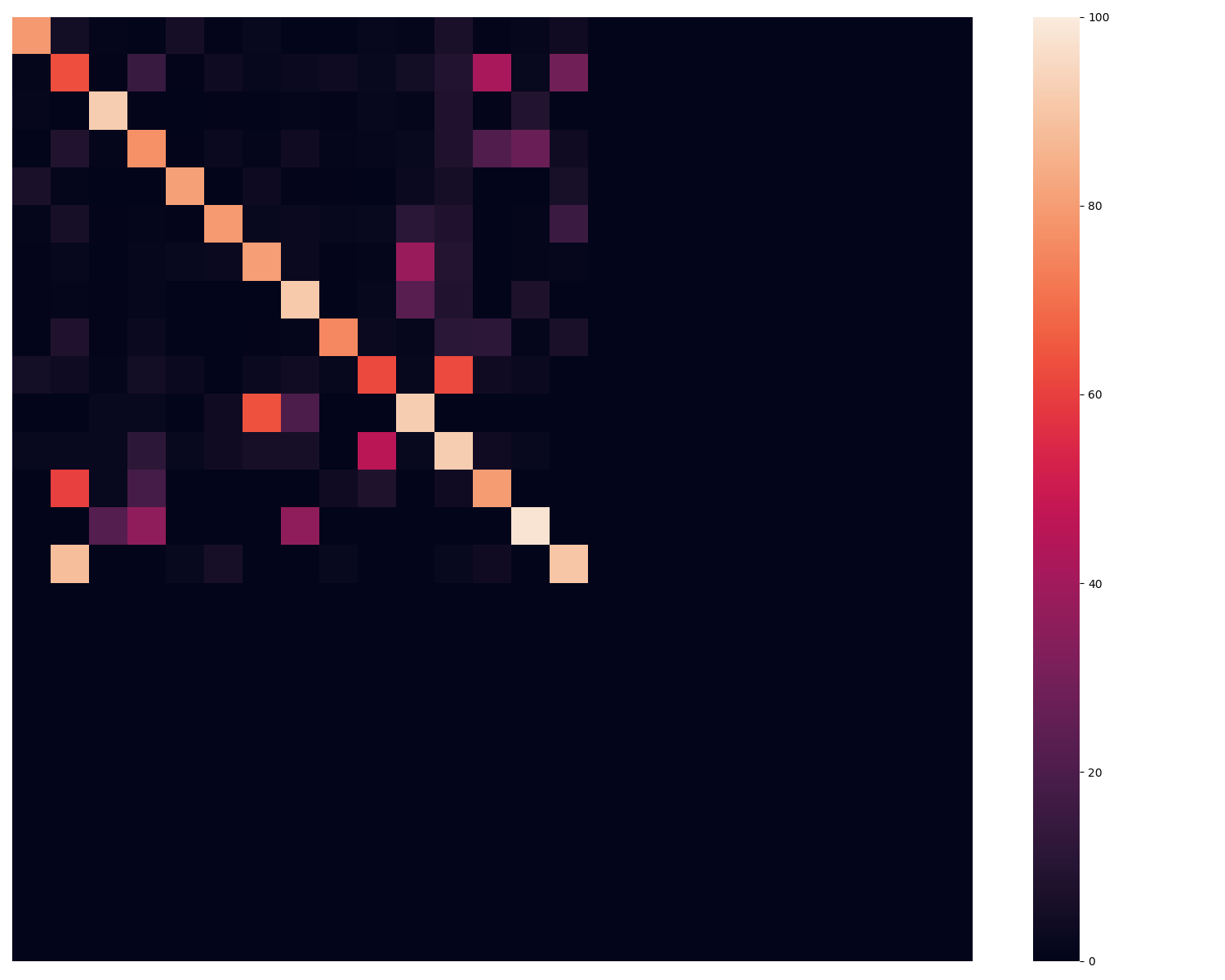}
}
\subfigure[After task 5]{
\includegraphics[width=0.23\linewidth]{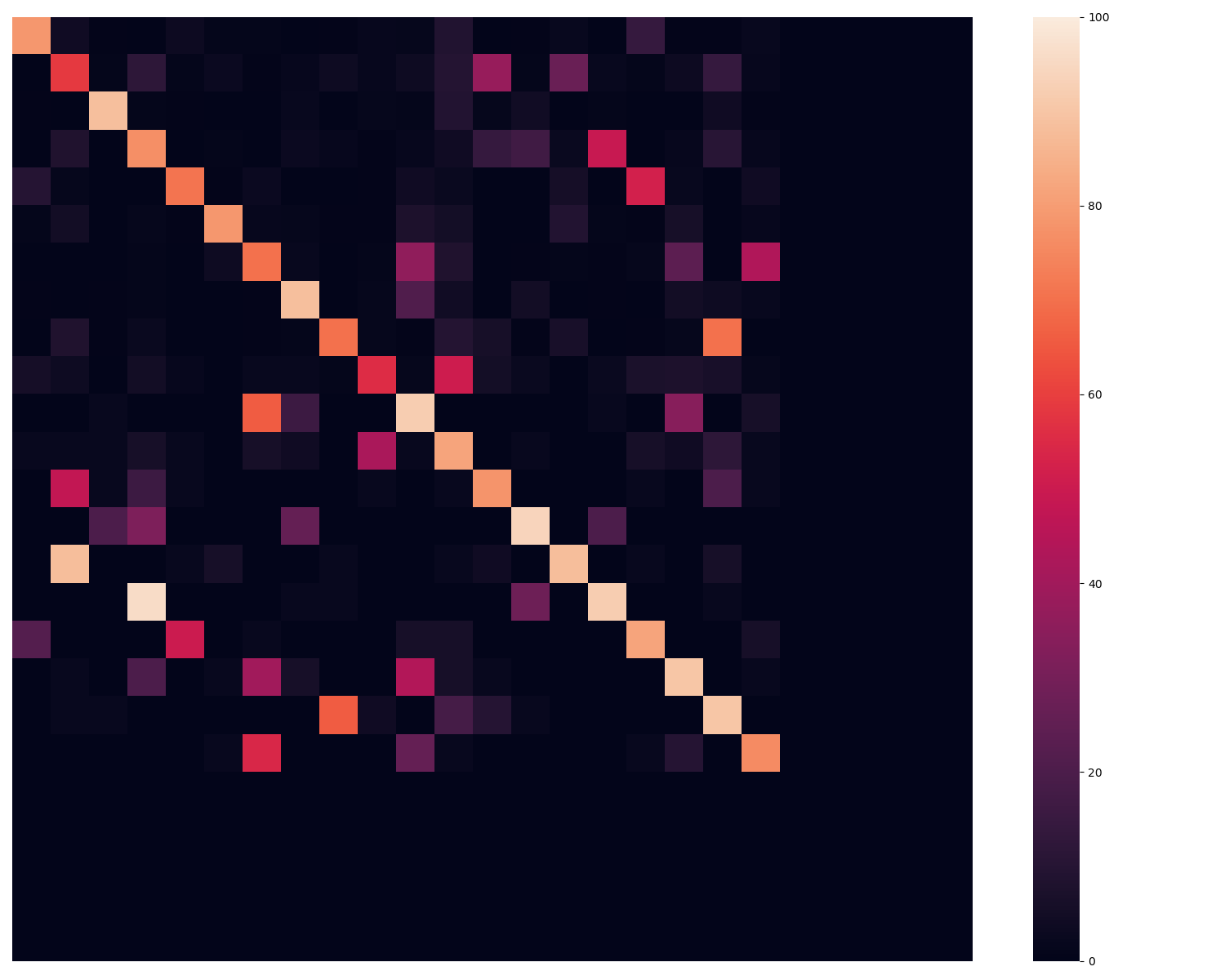}
}
\subfigure[After task 10]{
\includegraphics[width=0.23\linewidth]{latex/figures/confusion_matrices/icarl_cnn_norm_20/task_10_confusion.png}
}
\end{center}
\caption{iCaRL-norm confusion matrix after introducing tasks 0, 1, 5, 10 of IIRC-CIFAR respectively. The y-axis is the correct label (or one of the correct labels). The x-axis is the model predicted labels. Labels are arranged by their order of introduction. Only 25 labels are shown for better visibility.}
\label{fig:icarl_norm_confusion_matrix_elaborate}
\end{figure}

\begin{figure}[!ht]
\begin{center}
\subfigure[After task 0]{
\includegraphics[width=0.23\linewidth]{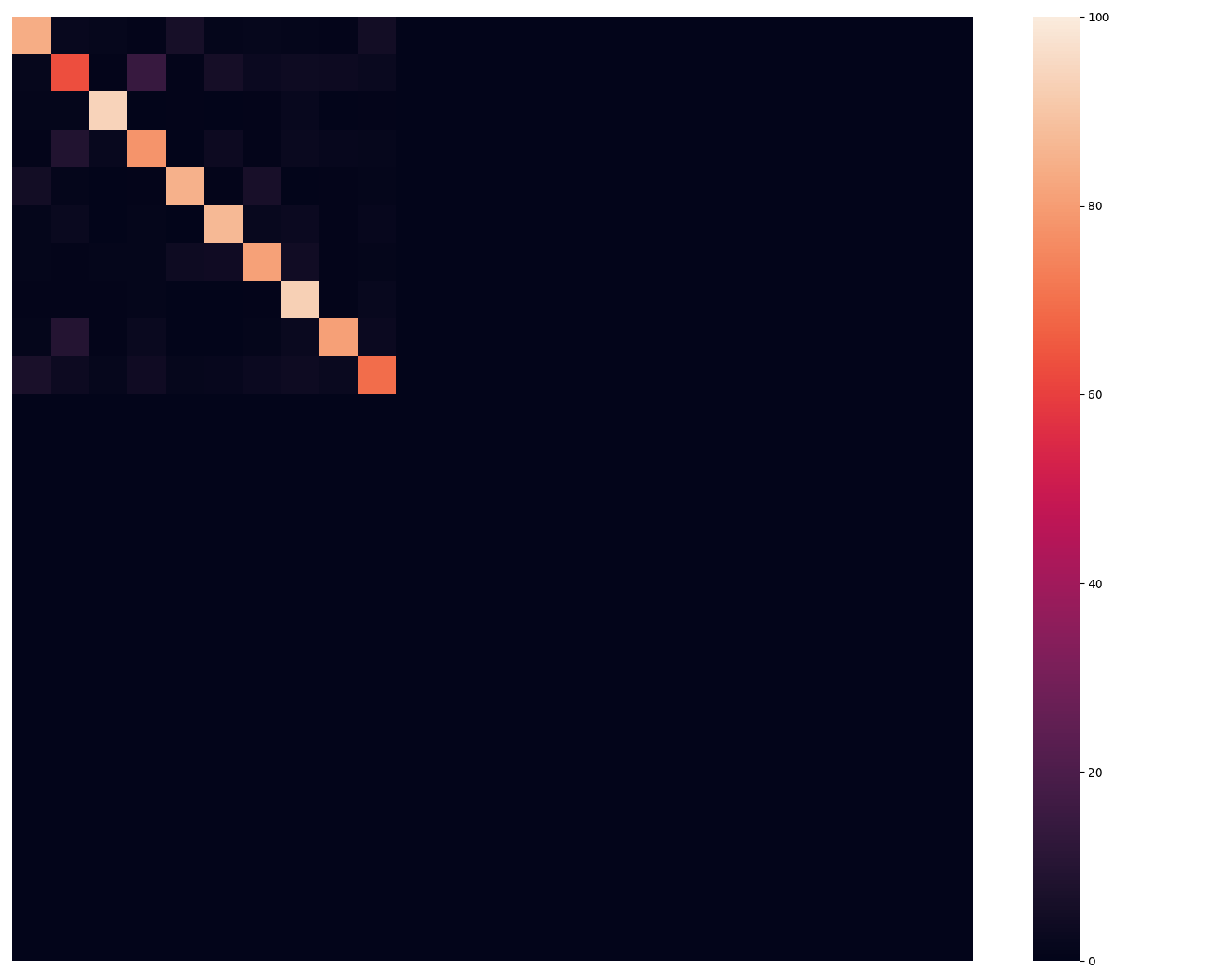}
}
\subfigure[After task 1]{
\includegraphics[width=0.23\linewidth]{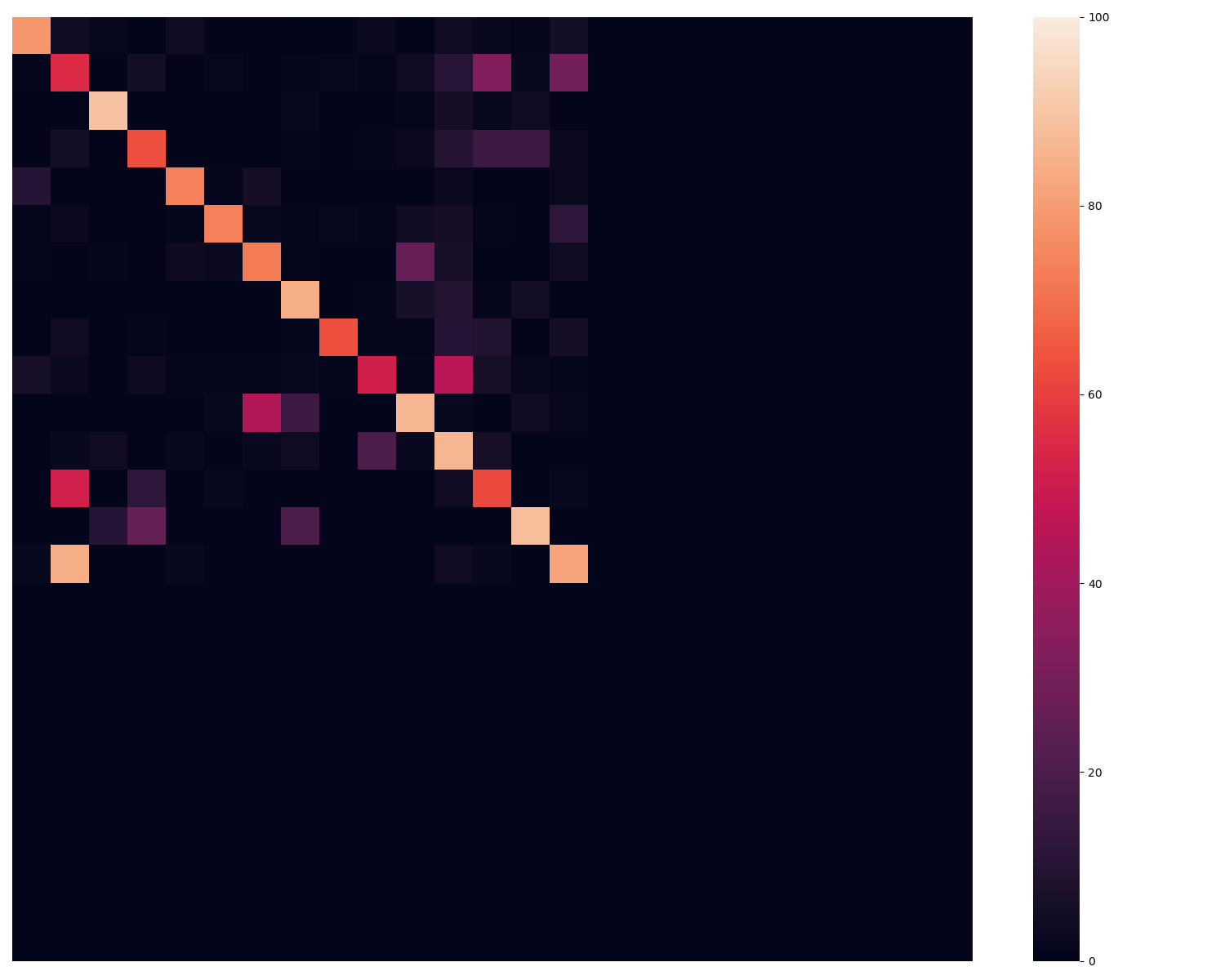}
}
\subfigure[After task 5]{
\includegraphics[width=0.23\linewidth]{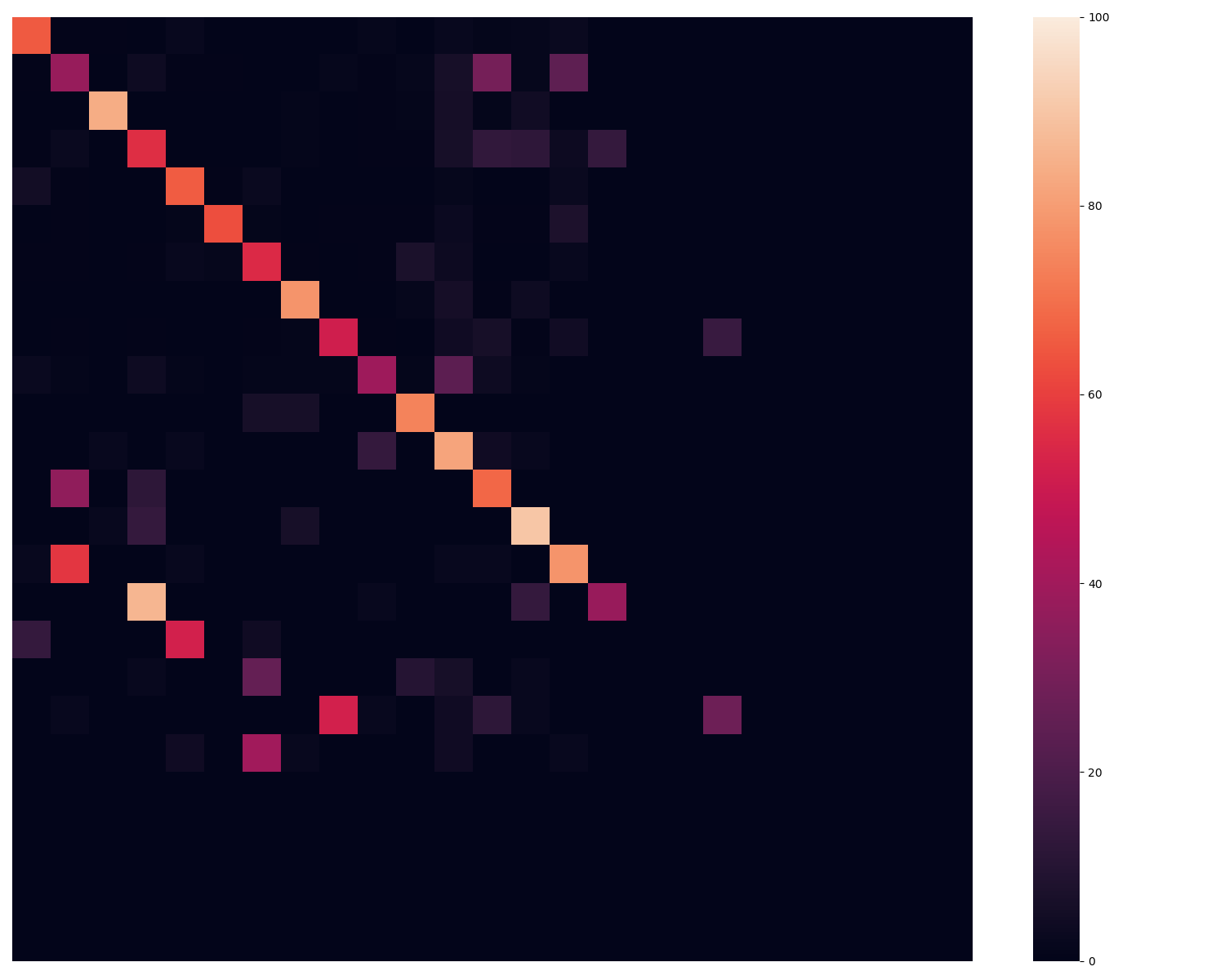}
}
\subfigure[After task 10]{
\includegraphics[width=0.23\linewidth]{latex/figures/confusion_matrices/unified_20/task_10_confusion.png}
}
\end{center}
\caption{LUCIR confusion matrix after introducing tasks 0, 1, 5, 10 of IIRC-CIFAR respectively. The y-axis is the correct label (or one of the correct labels). The x-axis is the model predicted labels. Labels are arranged by their order of introduction. Only 25 labels are shown for better visibility.}
\label{fig:unified_confusion_matrix_elaborate}
\end{figure}

\clearpage
\subsection{Full Resolution Confusion Matrix}
\label{confusion_full_resolution}

\begin{figure}[!ht]
\begin{center}
\includegraphics[width=1.0\linewidth]{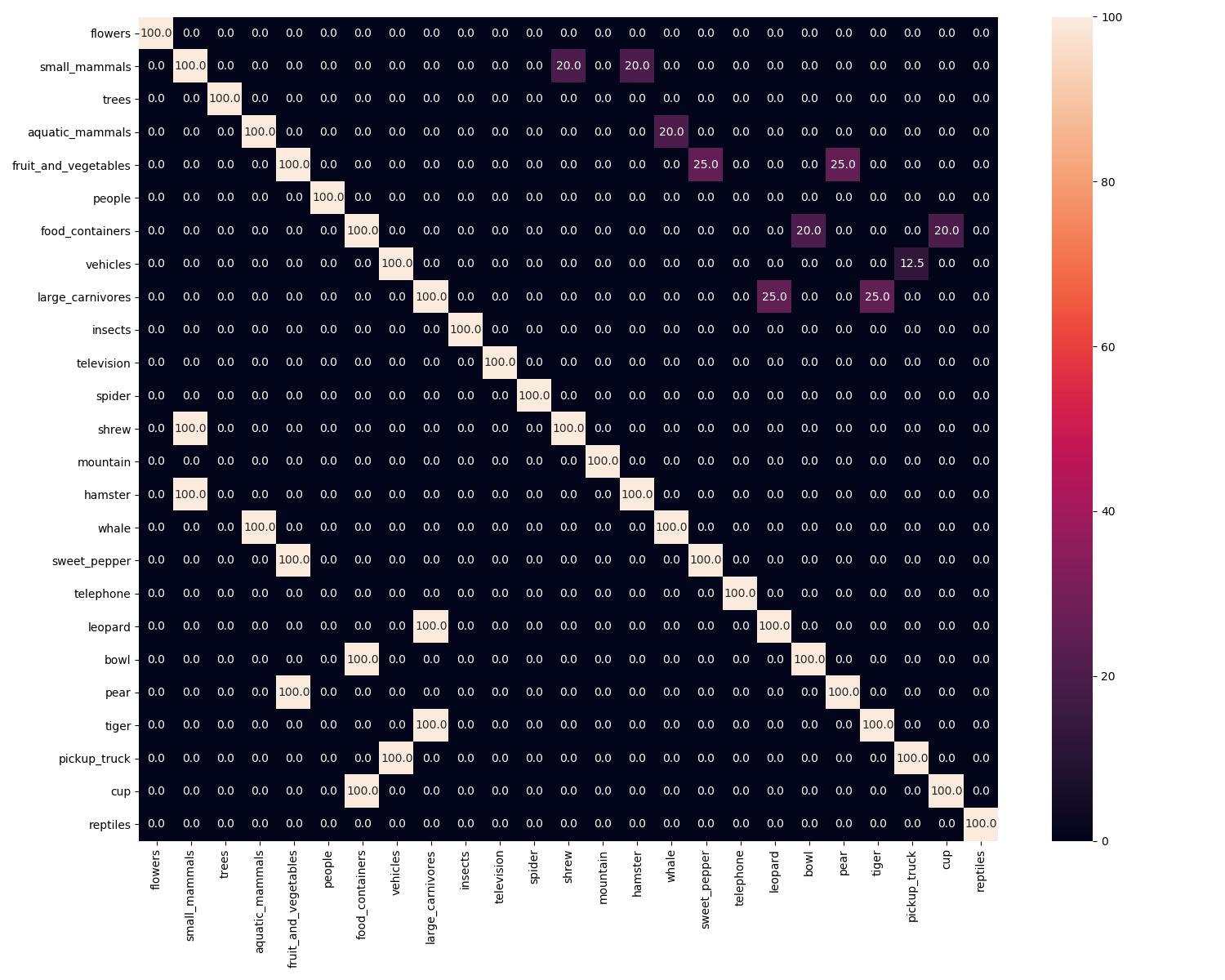}
\end{center}
\caption{Confusion matrix (ground truth) after training on task 10 of IIRC-CIFAR. the y-axis is the correct label (or one of the correct labels), the x-axis is the model predicted labels, The classes are arranged by their order of introduction. Only 25 classes are shown for better visibility. The y-axis represents the true label (or one of the true labels), while the x-axis represents the model predictions.
}
\label{fig:cifar_confusion_gt_full_version}
\end{figure}

\begin{figure}[!ht]
\begin{center}
\includegraphics[width=1.0\linewidth]{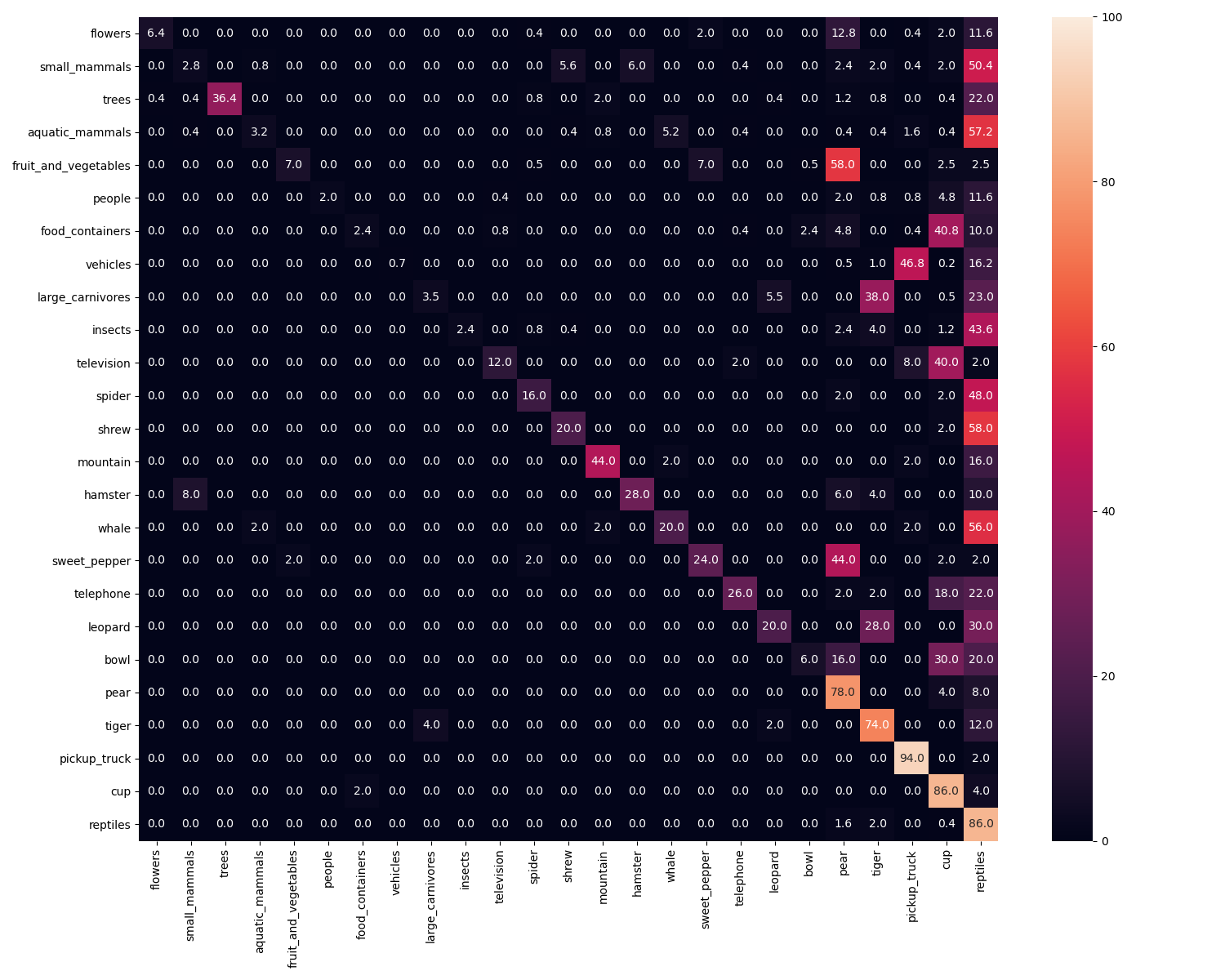}
\end{center}
\caption{Confusion matrix (ER) after training on task 10 of IIRC-CIFAR. the y-axis is the correct label (or one of the correct labels), the x-axis is the model predicted labels, The classes are arranged by their order of introduction. Only 25 classes are shown for better visibility. The y-axis represents the true label (or one of the true labels), while the x-axis represents the model predictions.
}
\label{fig:cifar_confusion_replay_full_version}
\end{figure}

\begin{figure}[!ht]
\begin{center}
\includegraphics[width=1.0\linewidth]{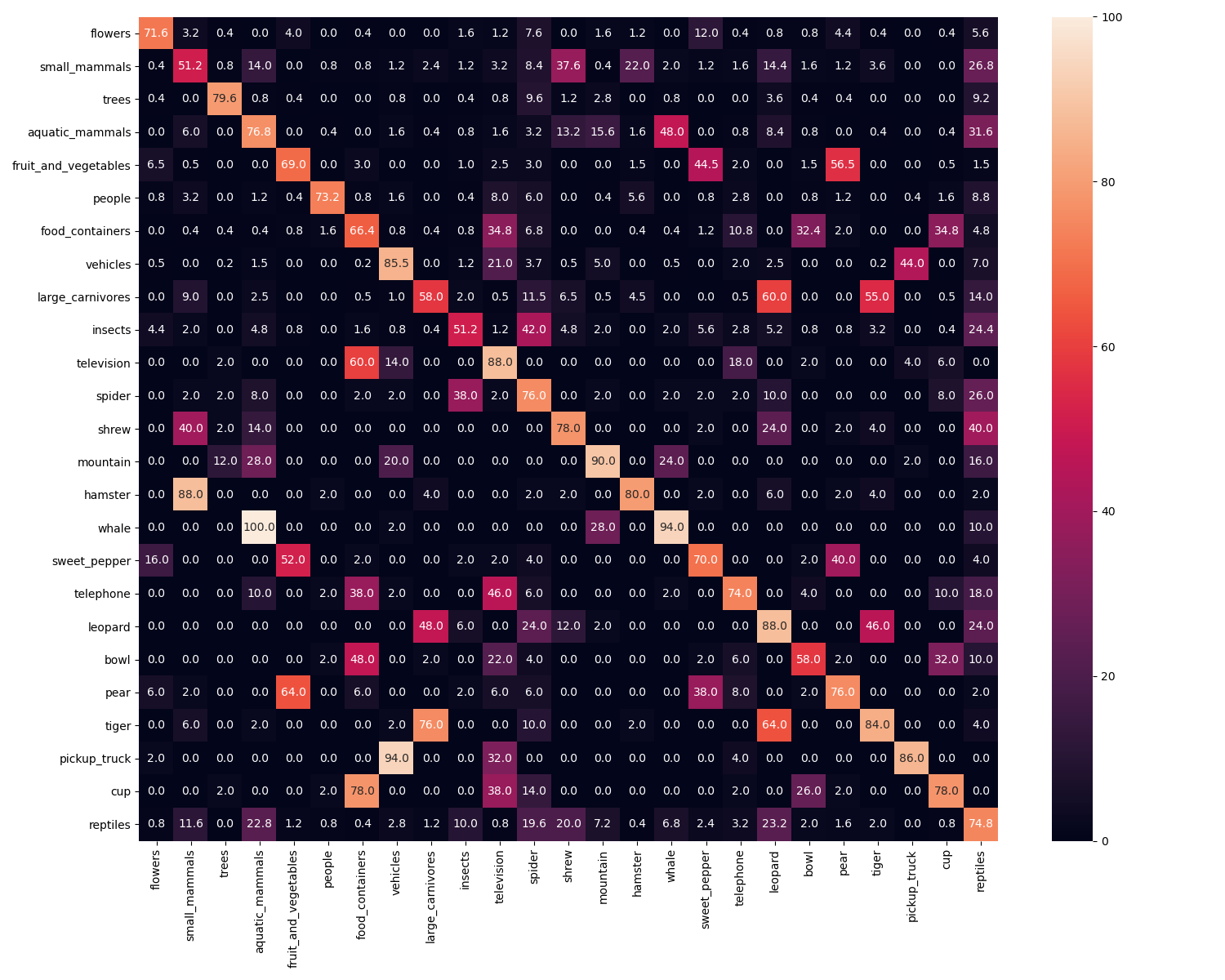}
\end{center}
\caption{Confusion matrix (iCaRL-norm) after training on task 10 of IIRC-CIFAR. The x-axis is the model predicted labels, The classes are arranged by their order of introduction. Only 25 classes are shown for better visibility. The y-axis represents the true label (or one of the true labels), while the x-axis represents the model predictions.
}
\label{fig:cifar_confusion_icarl_full_version}
\end{figure}

\begin{figure}[!ht]
\begin{center}
\includegraphics[width=1.0\linewidth]{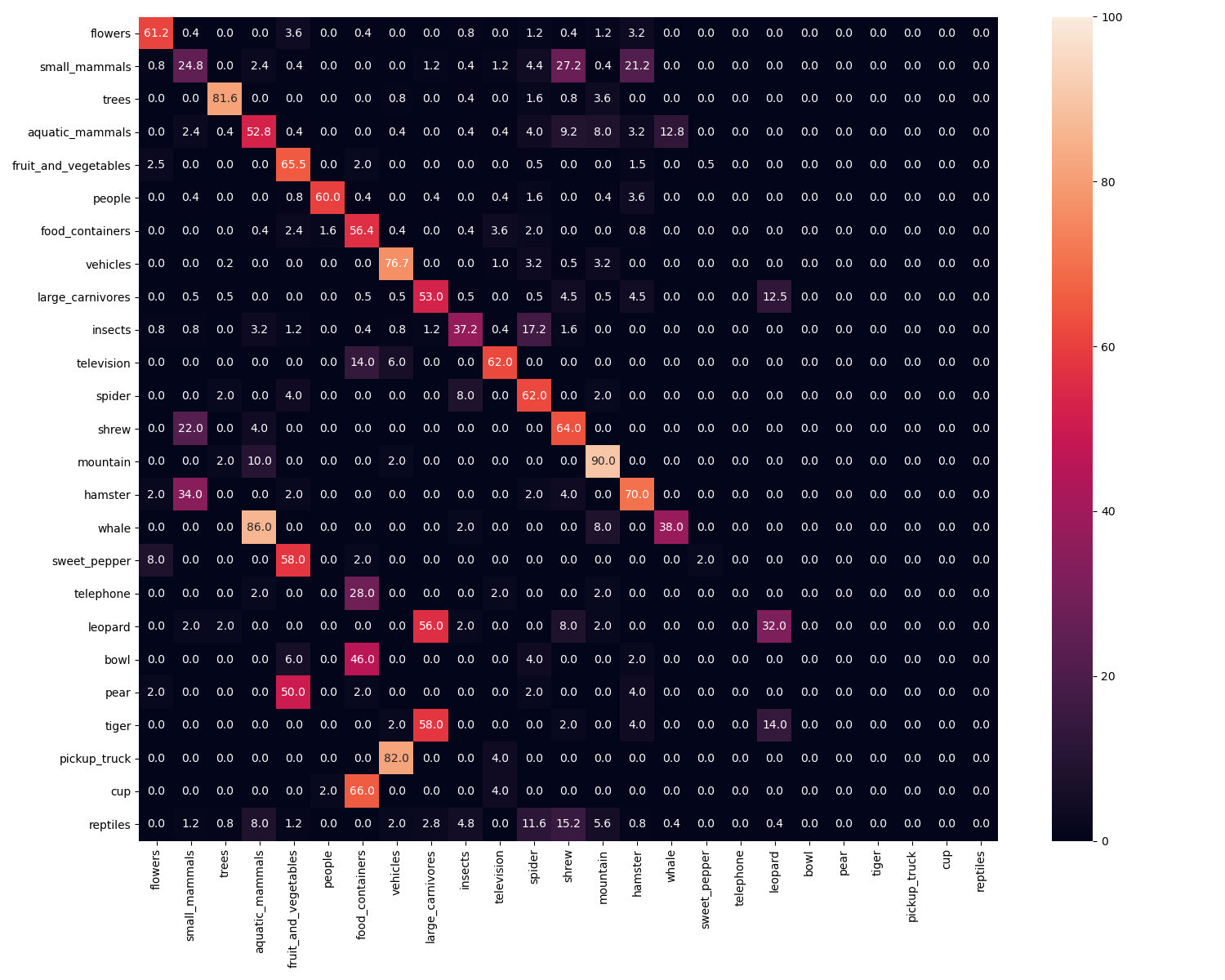}
\end{center}
\caption{Confusion matrix (LUCIR) after training on task 10 of IIRC-CIFAR. The x-axis is the model predicted labels, The classes are arranged by their order of introduction. Only 25 classes are shown for better visibility. The y-axis represents the true label (or one of the true labels), while the x-axis represents the model predictions.
}
\label{fig:cifar_confusion_unified_full_version}
\end{figure}

\clearpage

\section{Effect of Buffer}
\label{buffer_results}

\begin{figure}[h!]
\begin{center}
\includegraphics[clip,trim=3.5cm 1.5cm 4.5cm 2.75cm, width=0.9\linewidth]{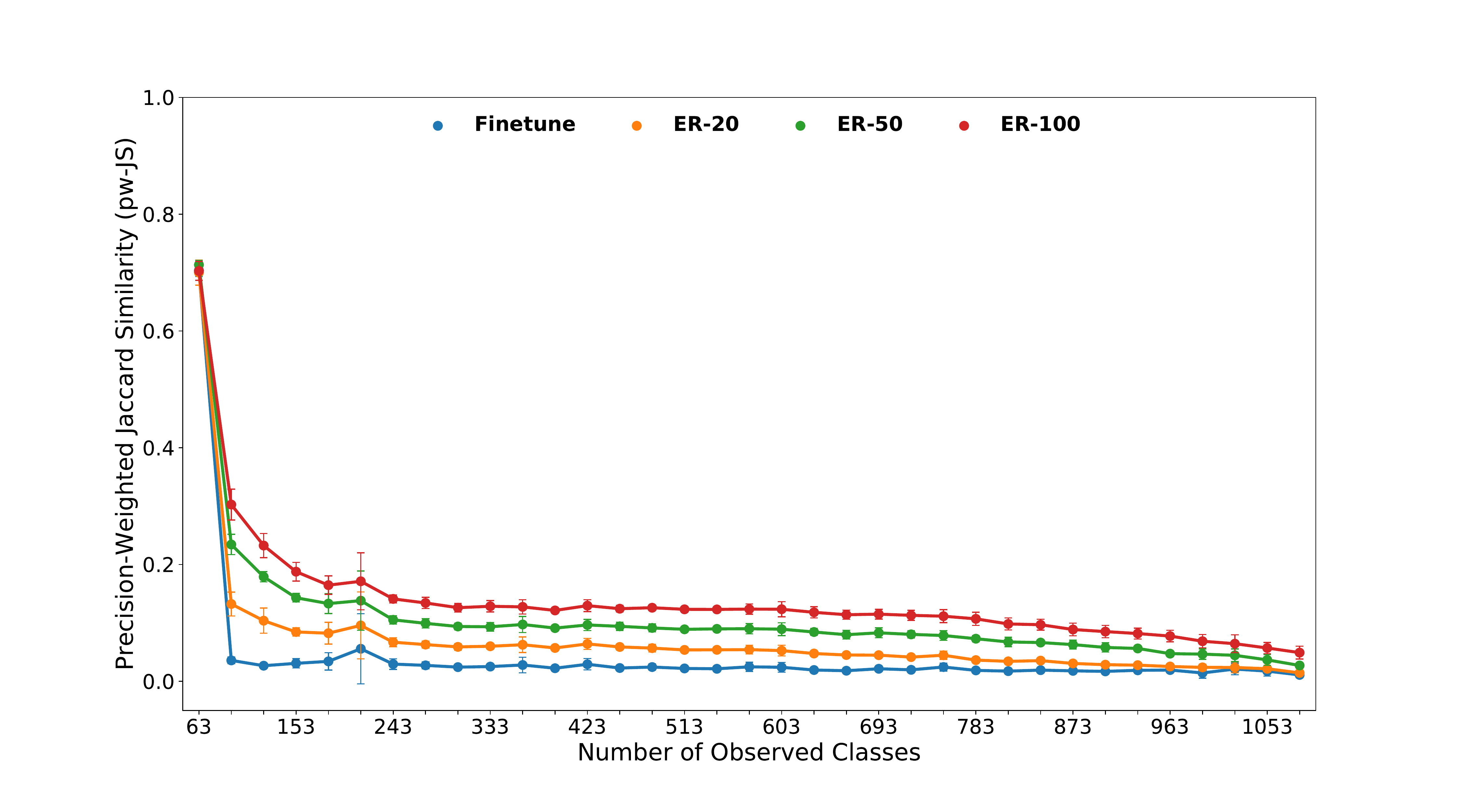}
\end{center}
\caption{Imagenet average performance using different buffer sizes, The number next to ER indicates the number of samples per class used for the replay buffer}
\label{fig:imagenet_buffer}
\end{figure}

\begin{figure}[h!]
\begin{center}
\includegraphics[clip,trim=3.5cm 1.5cm 4.5cm 2.75cm, width=0.9\linewidth]{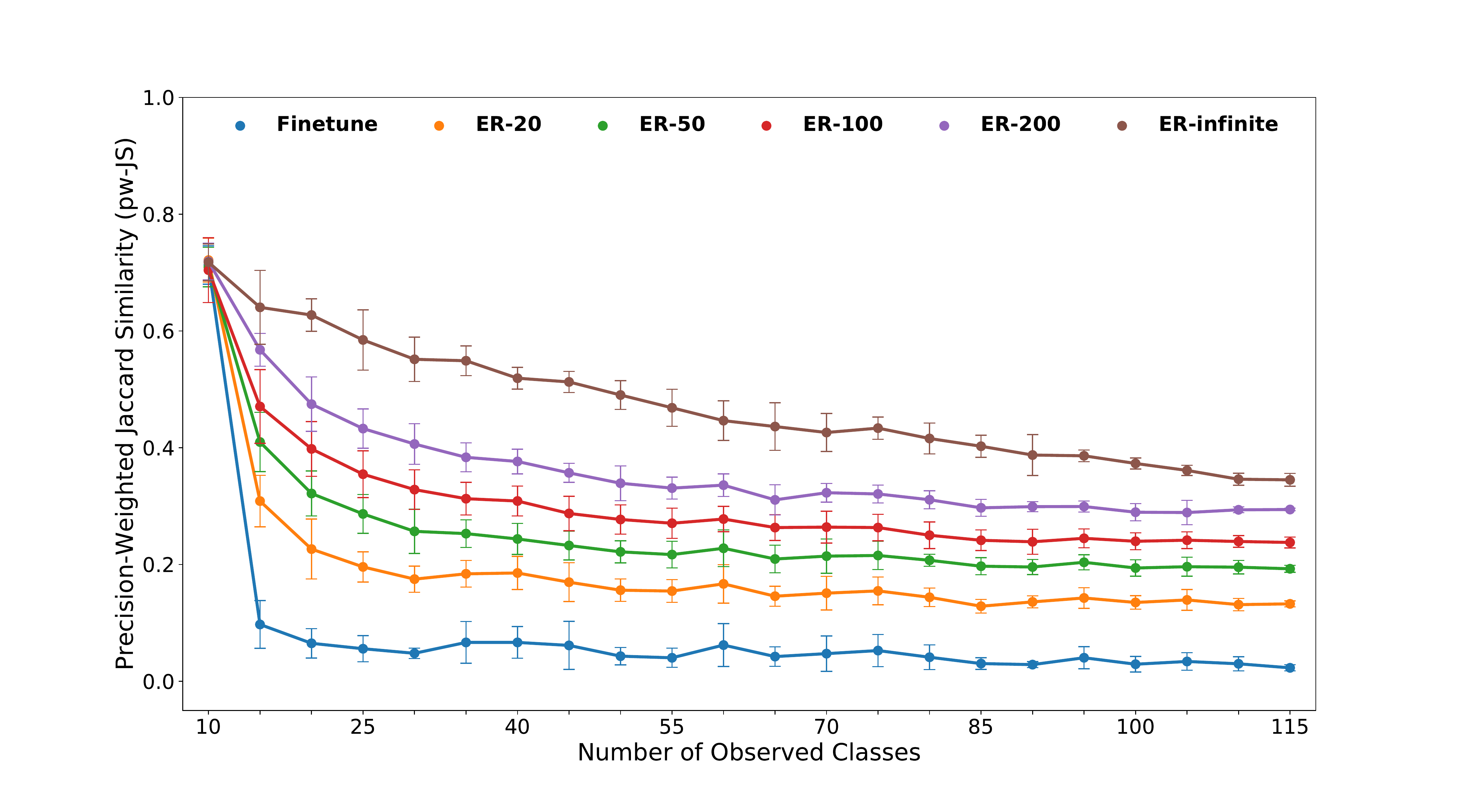}
\end{center}
\caption{CIFAR average performance using different buffer sizes, The number next to ER indicates the number of samples per class used for the replay buffer}
\label{fig:cifar_buffer}
\end{figure}
\clearpage

\section{Pseudo Codes}
\label{appendix:pseudo_codes}
\begin{algorithm}
\SetAlgoLined
\DontPrintSemicolon
\SetKwInOut{Input}{input}\SetKwInOut{Output}{output}\SetKwInOut{Require}{Require}
\Require{$tasks$\qquad // A list of the classes to-be-introduced at each task}
$trainSet,\;validSet_{inTask},\;validSet_{postTask},\;testSet$ $\leftarrow$ LoadDatasets()\;
$model$ $\leftarrow$ CreateModel() \;
\tcc{create an empty buffer}
$buffer$ $\leftarrow$ CreateBuffer()\;
\For {$task$ in $tasks$}{
$model$ $\leftarrow$ TrainOnTask($model,\; buffer,\;trainingSet,\;validSet_{inTask}$)\;
\tcc{add randomly selected samples to buffer}
$buffer$ $\leftarrow$ AddToBuffer ($buffer,\;trainingSet$) \;
PostTaskEvaluate($model,\;validSet_{postTask},\;testSet$)\;
}
\caption{IncrementalTrain\label{MC}}
\end{algorithm}

\begin{algorithm}
\SetAlgoLined
\DontPrintSemicolon
\SetKwInOut{Input}{input}\SetKwInOut{Output}{output}\SetKwInOut{Require}{Require}
\Input{$rawData_{train}$ \qquad // The default single-label full dataset (train)}
\Input{$rawData_{test}$\qquad // The default single-label full dataset (test)}
\Input{$classHierarchy$\qquad // A dictionary that maps each superclass to its constituent subclasses}

$multilabeledData_{train}$ $\leftarrow$ AddSuperclassLabels($rawData_{train},\;classHierarchy$)\;
$multilabeledData_{test}$ $\leftarrow$ AddSuperclassLabels($rawData_{test},\;classHierarchy$)\;
\BlankLine
$multilabeledData_{train},\;multilabeledData_{valid_{inTask}},\;multilabeledData_{valid_{postTask}}$ $\leftarrow$ SplitData($multilabeledData_{train}$)\;
\BlankLine
$trainSet$ $\leftarrow$ IncompleteInfoIncrementalDataset($multilabeledData_{train}$)\;
$validSet_{inTask}$ $\leftarrow$ IncompleteInfoIncrementalDataset($multilabeledData_{valid_{inTask}}$)\;
$validSet_{postTask}$ $\leftarrow$ CompleteInfoIncrementalTestDataset($multilabeledData_{valid_{postTask}}$)\;
$testSet$ $\leftarrow$ CompleteInfoIncrementalTestDataset($multilabeledData_{test}$)\;
\Output{$trainSet$\qquad // The incomplete information incremental learning training set}
\Output{$validSet_{inTask}$\qquad // The incomplete information incremental learning validation set (for in-task performance)}
\Output{$validSet_{postTask}$\qquad // The complete information incremental learning validation set (for post-task performance)}
\Output{$testSet$\qquad // The complete information incremental learning test set}
\caption{LoadDatasets\label{LD}}
\end{algorithm}

\begin{algorithm}
\SetAlgoLined
\DontPrintSemicolon
\SetKwInOut{Input}{input}\SetKwInOut{Output}{output}\SetKwInOut{Require}{Require}
\Input{$multilabelData$\qquad // A list of samples with each sample in the form of ($image$, ($superclassLabel,\;subclassLabel$)) or ($image$, ($subclassLabel$)))}
\Input{$superclassToSubclass$\qquad // a mapping that maps superclasses to their constituent subclasses}
\Input{$tasks$\qquad // The classes to-be-introduced at each task}
\Require{$subclasses$\qquad // All refined subclasses (those who have a superclass as well as those who don't)}
\Output{a dataset object with the data changing along the tasks}
\BlankLine
\textbf{Initialization:}\;
$classToDataIndices$ $\leftarrow$ EmptyDictionary\;
$currentTaskId$ $\leftarrow$ 0\;
$dataIndices_{task}$ $\leftarrow$ []\;

\For {$subclass$ in $subclasses$}{
    \tcc{get the indices of the samples which correspond to this subclass}
    $dataIndices_{subclass}$ $\leftarrow$ GetSamplesIndices($mtultilabelData,\;subclass$)\;
    \If {$subclass$ has superclass}{
        $dataSubsetLength_{superclass}$ $\leftarrow$ 0.4 * Length($dataIndices_{subclass}$)\;
        $dataSubsetLength_{subclass}$ $\leftarrow$ 0.8 * Length($dataIndices_{subclass}$)\;
        $dataIndices_{subclass}$ $\leftarrow$ Shuffle($dataIndices_{subclass}$)\;
        $dataSubsetIndices_{subclass}$ $\leftarrow$ $dataIndices_{subclass}$[:$dataSubsetLength_{subclass}$]\;
        $dataSubsetIndices_{superclass}$ $\leftarrow$ $dataIndices_{subclass}$[-$dataSubsetLength_{superclass}$:]\;
        $classToDataIndices$[$subclass$] $\leftarrow$ $dataIndices_{subclass}$ \;
        $classToDataIndices$[$superclass$] $\leftarrow$ $classToDataIndices$[$superclass$] $\cup$ $dataSubsetIndices_{superclass}$ \;
    }
    \ElseIf {$subclass$ has no superclass} {
        $classToDataIndices$[subclass] $\leftarrow$ $dataIndices_{subclass}$ \;
    }
}
\BlankLine
\textbf{IncrementTask:}\;
$currentTaskId$ $\leftarrow$ $currentTaskId + 1$\;
$dataIndices_{task}$ $\leftarrow$ []\;
\For{$class$ in $tasks$[$currentTaskId$]}{
    $dataIndices_{task}$ $\leftarrow$ $dataIndices_{task}$ $\cup$ $classToDataIndices$[$class$]\;
}
\BlankLine
\textbf{GetItem:}\;
\Require{$classes_{task}$\qquad // The classes present in the current task}
\Input{$index$\qquad // an index in the range of length of $dataIndices_{task}$}
$image,\;labels$ $\leftarrow$ $multilabelData$[$dataIndices_{task}$[$index$]]\;
$label$ $\leftarrow$ $labels \cap classes_{task}$\;
\Output{$image$\qquad // The sample image}
\Output{$label$\qquad // The label corresponding to this image that exists in the current task}
\caption{IncompleteInfoIncrementalDataset\label{iiid}}
\end{algorithm}

\begin{algorithm}
\SetAlgoLined
\DontPrintSemicolon
\SetKwInOut{Input}{input}\SetKwInOut{Output}{output}\SetKwInOut{Require}{Require}
\Input{$multilabelData$\qquad // A list of samples with each sample in the form of ($image$, ($superclassLabel,\;subclassLabel$)) or ($image$, ($subclassLabel$)))}
\Input{$superclassToSubclass$\qquad // a mapping that maps superclasses to their constituent subclasses}
\Input{$tasks$\qquad // The classes available at each task}
\Require{$subclasses$\qquad // All refined subclasses (those who have a superclass as well as those who don't)}
\Output{a test dataset object which keeps collecting data along the tasks}
\BlankLine
\textbf{Initialization:}\;
$classToDataIndices$ $\leftarrow$ empty\_dictionary\;
$classes_{observed}$ $\leftarrow$ []\;
$dataIndices_{accessible} \leftarrow$ []\;
\For {$subclass$ in $subclasses$}{
    \tcc{get the indices of the samples which correspond to this subclass}
    $dataIndices_{subclass}$ $\leftarrow$ GetSamplesIndices($multilabelData,\;subclass$)\;
    $classToDataIndices$[$subclass$] $\leftarrow$ $dataIndices_{subclass}$ \;
    \If {subclass has superclass}{
        $classToDataIndices$[$superclass$] $\leftarrow$ $classToDataIndices$[$superclass$] $\cup$ $classToDataIndices$[$subclass$] \;
    }
}
\BlankLine
\textbf{LoadTask}\;
\Input{$taskId$\qquad // The index of the task to load}
$dataIndices_{accessible}$ $\leftarrow$ []\;
$classes_{observed}$ $\leftarrow$ $classes_{observed}$ $\cup$ $tasks$[$taskId$]\;
\For{$class$ in $tasks$[$taskId$]}{
    $dataIndices_{accessible}$ $\leftarrow$ $dataIndices_{accessible} \cup classToDataIndices$[$class$]
}
\BlankLine
\textbf{LoadAllObservedData}\;
\Require{$classes_{observed}$\qquad // All classes observed till now in all previous tasks}
$dataIndices_{accessible}$ $\leftarrow$ []\;
\For{$class$ in $classes_{observed}$}{
    $dataIndices_{accessible} \leftarrow dataIndices_{accessible} \cup classToDataIndices$[$class$]
}
$dataIndices_{accessible} \leftarrow$ RemoveDuplicates($dataIndices_{accessible})$\;
\BlankLine
\textbf{GetItem}\;
\Require{$classes_{observed}$\qquad // All classes observed till now in all previous tasks}
\Input{$index$\qquad // an index in the range of task\_data\_indices}
$image,\;labels$ $\leftarrow$ $multilabelData$[$dataIndices_{accessible}$[$index$]] \;
$labels$ $\leftarrow$ $labels$ $\cap$ $classes_{observed}$ \;
\Output{$image$\qquad // The sample image}
\Output{$labels$\qquad // The labels corresponding to this image that exist in the $classes_{observed}$}
\caption{CompleteInfoIncrementalTestDataset\label{ciitd}}
\end{algorithm}

\clearpage

\section{IIRC Datasets Hierarchies}
\label{appendix:clusters}
\subsection{IIRC-CIFAR Hierarchy}
\label{appendix:cifar_clusters}
\begin{table}[!htb]
\centering
\begin{tabular}{||c | L ||} 
 \hline
 superclass & subclasses \\ [0.5ex] 
 \hline\hline
aquatic mammals  & beaver, dolphin, otter, seal, whale\\ [0.5ex] \hline
fish  & aquarium fish, flatfish, ray, shark, trout\\ [0.5ex] \hline
flowers  & orchid, poppy, rose, sunflower, tulip\\ [0.5ex] \hline
food containers  & bottle, bowl, can, cup, plate\\ [0.5ex] \hline
fruit and vegetables  & apple, orange, pear, sweet pepper\\ [0.5ex] \hline
household furniture  & bed, chair, couch, table, wardrobe\\ [0.5ex] \hline
insects  & bee, beetle, butterfly, caterpillar, cockroach\\ [0.5ex] \hline
large carnivores  & leopard, lion, tiger, wolf\\ [0.5ex] \hline
large omnivores and herbivores  & bear, camel, cattle, chimpanzee, elephant, kangaroo\\ [0.5ex] \hline
medium sized mammals  & fox, porcupine, possum, raccoon, skunk\\ [0.5ex] \hline
people  & baby, boy, girl, man, woman\\ [0.5ex] \hline
reptiles  & crocodile, dinosaur, lizard, snake, turtle\\ [0.5ex] \hline
small mammals  & hamster, mouse, rabbit, shrew, squirrel\\ [0.5ex] \hline
trees  & maple tree, oak tree, palm tree, pine tree, willow tree\\ [0.5ex] \hline
vehicles  & bicycle, bus, motorcycle, pickup truck, train, streetcar, tank, tractor \\[0.5ex] \hline
  - & mushroom, clock, keyboard, lamp, telephone, television, bridge, castle,
        house, road, skyscraper, cloud, forest, mountain, plain, sea, crab, lobster, snail,
        spider, worm, lawn mower, rocket\\ [1.0ex] 
 \hline
\end{tabular}
\label{table:cifar_hierarchy}
\end{table}
\clearpage

\subsection{IIRC-ImageNet Hierarchy}
\label{appendix:imagenet_clusters}
\begin{longtable}{||c | L ||} 
 \hline
 superclass & subclasses \\ [0.5ex] 
 \hline\hline
        dog  & 
            dalmatian, 
            basenji, 
            pug, 
            Leonberg, 
            Newfoundland, 
            Great Pyrenees, 
            Mexican hairless, 
            Brabancon griffon, 
            Pembroke, 
            Cardigan, 
            Chihuahua, 
            Japanese spaniel, 
            Maltese dog, 
            Pekinese, 
            Shih-Tzu, 
            toy terrier, 
            papillon, 
            Blenheim spaniel, 
            Rhodesian ridgeback, 
            boxer, 
            bull mastiff, 
            Great Dane, 
            Saint Bernard, 
            Eskimo dog, 
            Tibetan mastiff, 
            French bulldog, 
            malamute, 
            Siberian husky, 
            Samoyed, 
            Pomeranian, 
            chow, 
            keeshond, 
            toy poodle, 
            miniature poodle, 
            standard poodle, 
            Afghan hound, 
            basset, 
            beagle, 
            bloodhound, 
            bluetick, 
            redbone, 
            Ibizan hound, 
            Norwegian elkhound, 
            otterhound, 
            Saluki, 
            Scottish deerhound, 
            Weimaraner, 
            black-and-tan coonhound, 
            Walker hound, 
            English foxhound, 
            borzoi, 
            Irish wolfhound, 
            Italian greyhound, 
            whippet, 
            Bedlington terrier, 
            Border terrier, 
            Kerry blue terrier, 
            Irish terrier, 
            Norfolk terrier, 
            Norwich terrier, 
            Yorkshire terrier, 
            Airedale, 
            cairn, 
            Australian terrier, 
            Dandie Dinmont, 
            Boston bull, 
            Scotch terrier, 
            Tibetan terrier, 
            silky terrier, 
            soft-coated wheaten terrier, 
            West Highland white terrier, 
            Lhasa, 
            Staffordshire bullterrier, 
            American Staffordshire terrier, 
            wire-haired fox terrier, 
            Lakeland terrier, 
            Sealyham terrier, 
            German short-haired pointer, 
            vizsla, 
            kuvasz, 
            schipperke, 
            Doberman, 
            miniature pinscher, 
            affenpinscher, 
            Brittany spaniel, 
            clumber, 
            cocker spaniel, 
            Sussex spaniel, 
            English springer, 
            Welsh springer spaniel, 
            Irish water spaniel, 
            English setter, 
            Irish setter, 
            Gordon setter, 
            flat-coated retriever, 
            curly-coated retriever, 
            golden retriever, 
            Labrador retriever, 
            Chesapeake Bay retriever, 
            miniature schnauzer, 
            giant schnauzer, 
            standard schnauzer, 
            Greater Swiss Mountain dog, 
            Bernese mountain dog, 
            Appenzeller, 
            EntleBucher, 
            briard, 
            kelpie, 
            komondor, 
            Old English sheepdog, 
            Shetland sheepdog, 
            collie, 
            Border collie, 
            Bouvier des Flandres, 
            Rottweiler, 
            German shepherd, 
            groenendael, 
            malinois
        \\ [0.5ex]   \hline 
        bird  & 
            cock, 
            hen, 
            ostrich, 
            bee eater, 
            hornbill, 
            hummingbird, 
            jacamar, 
            toucan, 
            coucal, 
            quail, 
            partridge, 
            peacock, 
            black grouse, 
            ptarmigan, 
            ruffed grouse, 
            prairie chicken, 
            water ouzel, 
            robin, 
            bulbul, 
            jay, 
            magpie, 
            chickadee, 
            brambling, 
            goldfinch, 
            house finch, 
            junco, 
            indigo bunting, 
            black swan, 
            European gallinule, 
            goose,
            drake, 
            red-breasted merganser, 
            pelican, 
            albatross, 
            king penguin, 
            spoonbill, 
            flamingo, 
            limpkin, 
            bustard, 
            white stork, 
            black stork, 
            American coot, 
            oystercatcher, 
            red-backed sandpiper, 
            redshank, 
            dowitcher, 
            ruddy turnstone, 
            little blue heron, 
            bittern, 
            American egret, 
            African grey, 
            macaw, 
            sulphur-crested cockatoo, 
            lorikeet, 
            vulture, 
            kite, 
            bald eagle, 
            great grey owl
        \\ [0.5ex]   \hline 
        garment  & 
            suit, 
            abaya, 
            kimono, 
            cardigan, 
            feather boa, 
            stole, 
            jersey, 
            sweatshirt, 
            poncho, 
            brassiere, 
            jean, 
            gown, 
            military uniform, 
            pajama, 
            apron, 
            academic gown, 
            vestment, 
            bow tie, 
            Windsor tie, 
            fur coat, 
            lab coat, 
            trench coat, 
            hoopskirt, 
            miniskirt, 
            overskirt, 
            sarong, 
            cloak
        \\ [0.5ex]   \hline 
        beverage  & 
            espresso, 
            red wine, 
            cup, 
            eggnog
        \\ [0.5ex]   \hline 
        aircraft  & 
            airship, 
            balloon, 
            airliner, 
            warplane, 
            wing, 
            space shuttle
        \\ [0.5ex]   \hline 
        bear  & 
            brown bear, 
            American black bear, 
            ice bear, 
            sloth bear
        \\ [0.5ex]   \hline 
        fox  & 
            red fox, 
            kit fox, 
            Arctic fox, 
            grey fox
        \\ [0.5ex]   \hline 
        wolf  & 
            timber wolf, 
            white wolf, 
            red wolf, 
            coyote
        \\ [0.5ex]   \hline 
        bag  & 
            backpack, 
            mailbag, 
            plastic bag, 
            purse, 
            sleeping bag
        \\ [0.5ex]   \hline 
        footwear  & 
            clog, 
            cowboy boot, 
            Loafer, 
            running shoe, 
            sandal
        \\ [0.5ex]   \hline 
        toiletry  & 
            hair spray, 
            lotion, 
            perfume, 
            face powder, 
            sunscreen, 
            lipstick
        \\ [0.5ex]   \hline 
        box  & 
            carton, 
            chest, 
            crate, 
            mailbox, 
            pencil box, 
            safe
        \\ [0.5ex]   \hline 
        rodent  & 
            hamster, 
            porcupine, 
            marmot, 
            beaver, 
            guinea pig, 
            fox squirrel
        \\ [0.5ex]   \hline 
        bottle  & 
            beer bottle, 
            pill bottle, 
            pop bottle, 
            water bottle, 
            wine bottle, 
            water jug, 
            whiskey jug
        \\ [0.5ex]   \hline 
        fabric  & 
            velvet, 
            wool, 
            bib, 
            dishrag, 
            handkerchief, 
            bath towel, 
            paper towel
        \\ [0.5ex]   \hline 
        cup  & 
            beer glass, 
            goblet, 
            cocktail shaker, 
            measuring cup, 
            pitcher, 
            beaker, 
            coffee mug
        \\ [0.5ex]   \hline 
        fungus  & 
            coral fungus, 
            gyromitra, 
            stinkhorn, 
            earthstar, 
            hen-of-the-woods, 
            bolete, 
            agaric
        \\ [0.5ex]   \hline 
        musteline  & 
            weasel, 
            mink, 
            polecat, 
            black-footed ferret, 
            otter, 
            skunk, 
            badger
        \\ [0.5ex]   \hline 
        truck  & 
            fire engine, 
            garbage truck, 
            pickup, 
            tow truck, 
            trailer truck, 
            moving van, 
            police van, 
            recreational vehicle, 
            forklift, 
            harvester, 
            snowplow, 
            tractor
        \\ [0.5ex]   \hline 
        headdress  & 
            crash helmet, 
            football helmet, 
            bearskin, 
            bonnet, 
            cowboy hat, 
            sombrero, 
            bathing cap, 
            mortarboard, 
            shower cap, 
            pickelhaube
        \\ [0.5ex]   \hline 
        ball  & 
            baseball, 
            basketball, 
            croquet ball, 
            golf ball, 
            ping-pong ball, 
            punching bag, 
            rugby ball, 
            soccer ball, 
            tennis ball, 
            volleyball
        \\ [0.5ex]   \hline 
        car  & 
            ambulance, 
            beach wagon, 
            cab, 
            convertible, 
            jeep, 
            limousine, 
            Model T, 
            racer, 
            sports car, 
            minivan, 
            grille, 
            golfcart
        \\ [0.5ex]   \hline 
        measuring instrument  & 
            barometer, 
            scale, 
            odometer, 
            rule, 
            sundial, 
            digital watch, 
            hourglass, 
            parking meter, 
            stopwatch, 
            analog clock, 
            digital clock, 
            wall clock
        \\ [0.5ex]   \hline 
        tool  & 
            hammer, 
            plunger, 
            screwdriver, 
            shovel, 
            cleaver, 
            letter opener, 
            can opener, 
            corkscrew, 
            hatchet, 
            chain saw, 
            plane, 
            scabbard, 
            power drill, 
            carpenter's kit
        \\ [0.5ex]   \hline 
        watercraft  & 
            schooner, 
            catamaran, 
            trimaran, 
            fireboat, 
            gondola, 
            canoe, 
            yawl, 
            lifeboat, 
            speedboat, 
            pirate, 
            wreck, 
            container ship, 
            liner, 
            aircraft carrier, 
            submarine, 
            amphibian, 
            paddle
        \\ [0.5ex]   \hline 
        dish  & 
            Petri dish, 
            mixing bowl, 
            soup bowl, 
            tray
        \\ [0.5ex]   \hline 
        bus  & 
            minibus, 
            school bus, 
            trolleybus
        \\ [0.5ex]   \hline 
        cart  & 
            horse cart, 
            jinrikisha, 
            oxcart
        \\ [0.5ex]   \hline 
        tracked vehicle  & 
            snowmobile, 
            half track, 
            tank
        \\ [0.5ex]   \hline 
        lamp  & 
            candle, 
            spotlight, 
            jack-o'-lantern, 
            lampshade, 
            table lamp
        \\ [0.5ex]   \hline 
        optical instrument  & 
            binoculars, 
            projector, 
            sunglasses, 
            lens cap, 
            loupe, 
            Polaroid camera, 
            reflex camera
        \\ [0.5ex]   \hline 
        gymnastic apparatus  & 
            balance beam, 
            horizontal bar, 
            parallel bars
        \\ [0.5ex]   \hline 
        swine  & 
            hog, 
            wild boar, 
            warthog
        \\ [0.5ex]   \hline 
        rabbits  & 
            hare, 
            wood rabbit, 
            Angora
        \\ [0.5ex]   \hline 
        echinoderm  & 
            starfish, 
            sea urchin, 
            sea cucumber
        \\ [0.5ex]   \hline 
        wild dog  & 
            dingo, 
            dhole, 
            African hunting dog
        \\ [0.5ex]   \hline 
        pouched mammal  & 
            wombat, 
            wallaby, 
            koala
        \\ [0.5ex]   \hline 
        aquatic mammal  & 
            dugong, 
            grey whale, 
            killer whale, 
            sea lion
        \\ [0.5ex]   \hline 
        person  & 
            ballplayer, 
            scuba diver, 
            groom
        \\ [0.5ex]   \hline 
        mollusk  & 
            chiton, 
            chambered nautilus, 
            conch, 
            snail, 
            slug, 
            sea slug
        \\ [0.5ex]   \hline 
        weapon  & 
            bow, 
            projectile, 
            cannon, 
            missile, 
            rifle, 
            revolver, 
            assault rifle, 
            holster
        \\ [0.5ex]   \hline 
        bovid  & 
            bison, 
            water buffalo, 
            ram, 
            ox, 
            bighorn, 
            ibex, 
            hartebeest, 
            impala, 
            gazelle
        \\ [0.5ex]   \hline 
        salamander  & 
            European fire salamander, 
            common newt, 
            eft, 
            spotted salamander, 
            axolotl
        \\ [0.5ex]   \hline 
        frog  & 
            tree frog, 
            tailed frog, 
            bullfrog
        \\ [0.5ex]   \hline 
        big cat  & 
            leopard, 
            snow leopard, 
            jaguar, 
            lion, 
            tiger, 
            cheetah
        \\ [0.5ex]   \hline 
        domestic cat  & 
            tabby, 
            tiger cat, 
            Persian cat, 
            Siamese cat, 
            Egyptian cat
        \\ [0.5ex]   \hline 
        cooking utensil  & 
            spatula, 
            frying pan, 
            wok, 
            Crock Pot, 
            Dutch oven, 
            caldron, 
            coffeepot, 
            teapot
        \\ [0.5ex]   \hline 
        primate  & 
            Madagascar cat, 
            indri, 
            gibbon, 
            siamang, 
            orangutan, 
            gorilla, 
            chimpanzee, 
            marmoset, 
            capuchin, 
            howler monkey, 
            titi, 
            spider monkey, 
            squirrel monkey, 
            guenon, 
            patas, 
            baboon, 
            macaque, 
            langur, 
            colobus, 
            proboscis monkey
        \\ [0.5ex]   \hline 
        fish  & 
            barracouta, 
            electric ray, 
            stingray, 
            hammerhead, 
            great white shark, 
            tiger shark, 
            sturgeon, 
            gar, 
            puffer, 
            rock beauty, 
            anemone fish, 
            lionfish, 
            eel, 
            tench, 
            goldfish, 
            coho
        \\ [0.5ex]   \hline 
        lizard  & 
            banded gecko, 
            common iguana, 
            American chameleon, 
            whiptail, 
            agama, 
            frilled lizard, 
            alligator lizard, 
            Gila monster, 
            green lizard, 
            African chameleon, 
            Komodo dragon
        \\ [0.5ex]   \hline 
        turtle  & 
            mud turtle, 
            terrapin, 
            box turtle, 
            loggerhead, 
            leatherback turtle
        \\ [0.5ex]   \hline 
        spider  & 
            black and gold garden spider, 
            barn spider, 
            garden spider, 
            black widow, 
            tarantula, 
            wolf spider, 
            spider web
        \\ [0.5ex]   \hline 
        insect  & 
            ringlet, 
            sulphur butterfly, 
            lycaenid, 
            cabbage butterfly, 
            monarch, 
            admiral, 
            dragonfly, 
            damselfly, 
            lacewing, 
            cicada, 
            leafhopper, 
            cockroach, 
            mantis, 
            walking stick, 
            grasshopper, 
            cricket, 
            bee, 
            ant, 
            fly, 
            tiger beetle, 
            ladybug, 
            ground beetle, 
            long-horned beetle, 
            leaf beetle, 
            weevil, 
            dung beetle, 
            rhinoceros beetle
        \\ [0.5ex]   \hline 
        green groceries  & 
            acorn, 
            hip, 
            ear, 
            fig, 
            pineapple, 
            banana, 
            jackfruit, 
            custard apple, 
            pomegranate, 
            strawberry, 
            orange, 
            lemon, 
            Granny Smith, 
            buckeye, 
            rapeseed, 
            corn, 
            cucumber, 
            artichoke, 
            cardoon, 
            mushroom, 
            bell pepper, 
            mashed potato, 
            zucchini, 
            spaghetti squash, 
            acorn squash, 
            butternut squash, 
            broccoli, 
            cauliflower, 
            head cabbage
        \\ [0.5ex]   \hline 
        keyboard instrument  & 
            accordion, 
            organ, 
            grand piano, 
            upright
        \\ [0.5ex]   \hline 
        percussion instrument  & 
            chime, 
            drum, 
            gong, 
            maraca, 
            marimba, 
            steel drum
        \\ [0.5ex]   \hline 
        stringed instrument  & 
            banjo, 
            acoustic guitar, 
            electric guitar, 
            cello, 
            violin, 
            harp
        \\ [0.5ex]   \hline 
        wind instrument  & 
            ocarina, 
            harmonica, 
            flute, 
            panpipe, 
            bassoon, 
            oboe, 
            sax, 
            cornet, 
            French horn, 
            trombone
        \\ [0.5ex]   \hline 
        crustacean  & 
            isopod, 
            crayfish, 
            hermit crab, 
            spiny lobster, 
            American lobster, 
            Dungeness crab, 
            rock crab, 
            fiddler crab, 
            king crab
        \\ [0.5ex]   \hline 
        pen  & 
            ballpoint, 
            fountain pen, 
            quill
        \\ [0.5ex]   \hline 
        display  & 
            desktop computer, 
            laptop, 
            notebook, 
            screen, 
            television, 
            monitor
        \\ [0.5ex]   \hline 
        electronic equipement  & 
            cassette player, 
            CD player, 
            modem, 
            oscilloscope, 
            tape player, 
            iPod, 
            printer, 
            joystick, 
            dial telephone, 
            pay-phone, 
            cellular telephone, 
            mouse, 
            hand-held computer
        \\ [0.5ex]   \hline 
        snake  & 
            sea snake, 
            horned viper, 
            boa constrictor, 
            rock python, 
            Indian cobra, 
            green mamba, 
            diamondback, 
            sidewinder, 
            thunder snake, 
            ringneck snake, 
            hognose snake, 
            green snake, 
            king snake, 
            garter snake, 
            water snake, 
            vine snake, 
            night snake
        \\ [0.5ex]   \hline 
        geological formation  & 
            cliff, 
            geyser, 
            lakeside, 
            seashore, 
            valley, 
            promontory, 
            alp, 
            volcano, 
            coral reef, 
            sandbar
        \\ [0.5ex]   \hline 
        food  & 
            dough, 
            guacamole, 
            chocolate sauce, 
            carbonara, 
            French loaf, 
            bagel, 
            pretzel, 
            plate, 
            trifle, 
            ice cream, 
            ice lolly, 
            pizza, 
            potpie, 
            burrito, 
            consomme, 
            hot pot, 
            hotdog, 
            cheeseburger, 
            meat loaf
        \\ [0.5ex]   \hline 
        white home appliances  & 
            dishwasher, 
            refrigerator, 
            washer, 
            stove
        \\ [0.5ex]   \hline 
        kitchen appliances  & 
            microwave, 
            toaster, 
            waffle iron, 
            espresso maker
        \\ [0.5ex]   \hline 
        wheel  & 
            car wheel, 
            paddlewheel, 
            pinwheel, 
            potter's wheel, 
            reel, 
            disk brake
        \\ [0.5ex]   \hline 
        seat  & 
            toilet seat, 
            studio couch, 
            park bench, 
            barber chair, 
            folding chair, 
            rocking chair, 
            throne
        \\ [0.5ex]   \hline 
        baby bed  & 
            bassinet, 
            cradle, 
            crib
        \\ [0.5ex]   \hline 
        cabinet  & 
            medicine chest, 
            wardrobe, 
            china cabinet, 
            bookcase, 
            chiffonier, 
            file, 
            entertainment center, 
            plate rack
        \\ [0.5ex]   \hline 
        table  & 
            desk, 
            pool table, 
            dining table
        \\ [0.5ex]   \hline 
        bridges  & 
            steel arch bridge, 
            suspension bridge, 
            viaduct
        \\ [0.5ex]   \hline 
        fence  & 
            chainlink fence, 
            picket fence, 
            stone wall, 
            worm fence
        \\ [0.5ex]   \hline 
        long structures  & 
            beacon, 
            obelisk, 
            totem pole
        \\ [0.5ex]   \hline 
        movable homes  & 
            mountain tent, 
            mobile home, 
            yurt
        \\ [0.5ex]   \hline 
        building  & 
            planetarium, 
            barn, 
            cinema, 
            boathouse, 
            palace, 
            monastery, 
            castle, 
            dome, 
            church, 
            mosque, 
            stupa, 
            bell cote, 
            thatch, 
            tile roof, 
            triumphal arch
        \\ [0.5ex]   \hline 
        body armor  & 
            chain mail, 
            cuirass, 
            bulletproof vest, 
            breastplate
        \\ [0.5ex]   \hline 
        mask  & 
            mask, 
            oxygen mask, 
            gasmask, 
            ski mask
        \\ [0.5ex]   \hline 
        curtain-screen  & 
            window shade, 
            shower curtain, 
            theater curtain
        \\ [0.5ex]   \hline 
        bike  & 
            moped, 
            bicycle-built-for-two, 
            tricycle, 
            unicycle, 
            mountain bike, 
            motor scooter
        \\ [0.5ex]   \hline 
        train  & 
            passenger car, 
            freight car, 
            electric locomotive, 
            bullet train, 
            streetcar, 
            steam locomotive
        \\ [0.5ex]   \hline 
        swimsuit  & 
            bikini, 
            maillot, 
            swimming trunks
        \\ [0.5ex]   \hline 
        socks mittens  & 
            Christmas stocking, 
            mitten, 
            sock
        \\ [0.5ex]   \hline 
        keyboard  & 
            computer keyboard, 
            space bar, 
            typewriter keyboard
        \\ [0.5ex]  \hline
  - & African crocodile, 
        American alligator, 
        triceratops, 
        trilobite, 
        harvestman, 
        scorpion, 
        tick, 
        centipede, 
        tusker, 
        echidna, 
        platypus, 
        jellyfish, 
        sea anemone, 
        brain coral, 
        flatworm, 
        nematode, 
        crane, 
        hyena, 
        cougar, 
        lynx, 
        mongoose, 
        meerkat, 
        sorrel, 
        zebra, 
        hippopotamus, 
        Arabian camel, 
        llama, 
        armadillo, 
        three-toed sloth, 
        Indian elephant, 
        African elephant, 
        lesser panda, 
        giant panda, 
        abacus, 
        altar, 
        apiary, 
        ashcan, 
        bakery, 
        Band Aid, 
        bannister, 
        barbell, 
        barbershop, 
        barrel, 
        barrow, 
        bathtub, 
        binder, 
        birdhouse, 
        bobsled, 
        bolo tie, 
        bookshop, 
        bottlecap, 
        brass, 
        breakwater, 
        broom, 
        bucket, 
        buckle, 
        butcher shop, 
        car mirror, 
        carousel, 
        cash machine, 
        cassette, 
        chain, 
        cliff dwelling, 
        coil, 
        combination lock, 
        confectionery, 
        crutch, 
        dam, 
        diaper, 
        dock, 
        dogsled, 
        doormat, 
        drilling platform, 
        drumstick, 
        dumbbell, 
        electric fan, 
        envelope, 
        fire screen, 
        flagpole, 
        fountain, 
        four-poster, 
        gas pump, 
        go-kart, 
        greenhouse, 
        grocery store, 
        guillotine, 
        hair slide, 
        hamper, 
        hand blower, 
        hard disc, 
        home theater, 
        honeycomb, 
        hook, 
        iron, 
        jigsaw puzzle, 
        knee pad, 
        knot, 
        ladle, 
        lawn mower, 
        library, 
        lighter, 
        loudspeaker, 
        lumbermill, 
        magnetic compass, 
        manhole cover, 
        matchstick, 
        maypole, 
        maze, 
        megalith, 
        microphone, 
        milk can, 
        mortar, 
        mosquito net, 
        mousetrap, 
        muzzle, 
        nail, 
        neck brace, 
        necklace, 
        nipple, 
        oil filter, 
        packet, 
        padlock, 
        paintbrush, 
        parachute, 
        patio, 
        pedestal, 
        pencil sharpener, 
        photocopier, 
        pick, 
        pier, 
        piggy bank, 
        pillow, 
        plow, 
        pole, 
        pot, 
        prayer rug, 
        prison, 
        puck, 
        quilt, 
        racket, 
        radiator, 
        radio, 
        radio telescope, 
        rain barrel, 
        remote control, 
        restaurant, 
        rotisserie, 
        rubber eraser, 
        safety pin, 
        saltshaker, 
        scoreboard, 
        screw, 
        seat belt, 
        sewing machine, 
        shield, 
        shoe shop, 
        shoji, 
        shopping basket, 
        shopping cart, 
        ski, 
        slide rule, 
        sliding door, 
        slot, 
        snorkel, 
        soap dispenser, 
        solar dish, 
        space heater, 
        spindle, 
        stage, 
        stethoscope, 
        strainer, 
        stretcher, 
        sunglass, 
        swab, 
        swing, 
        switch, 
        syringe, 
        teddy, 
        thimble, 
        thresher, 
        tobacco shop, 
        torch, 
        toyshop, 
        tripod, 
        tub, 
        turnstile, 
        umbrella, 
        vacuum, 
        vase, 
        vault, 
        vending machine, 
        wallet, 
        washbasin, 
        water tower, 
        whistle, 
        wig, 
        window screen, 
        wooden spoon, 
        web site, 
        comic book, 
        crossword puzzle, 
        street sign, 
        traffic light, 
        book jacket, 
        menu, 
        hay, 
        bubble, 
        daisy, 
        yellow lady's slipper, 
        toilet tissue\\ [1.0ex] \hline
\end{longtable}

\clearpage

\section{Results in Tables}
\label{appendix:results_tables}
\newcolumntype{?}{!{\vrule width 1pt}}

\begin{table}[!htb]
\centering
\begin{tabular}{|| c | c | c | c | c | c ?c | c ||} 
 \hline
 task & \multicolumn{7}{c ||}{model} \\ [0.5ex] \hline
  & ER & iCaRL-CNN & iCaRL-norm & LUCIR & AGEM & incremental joint & ER-infinite \\ [0.5ex] \hline \hline 
  0 & 0.72 (0.039) & 0.71 (0.042) & 0.75 (0.037) & 0.74 (0.036) & 0.72 (0.032) & 0.71 (0.032) & 0.72 (0.033) \\ [0.5ex] \hline 
1 & 0.31 (0.046) & 0.47 (0.032) & 0.5 (0.035) & 0.5 (0.101) & 0.15 (0.024) & 0.7 (0.035) & 0.64 (0.067) \\ [0.5ex] \hline 
2 & 0.23 (0.054) & 0.36 (0.017) & 0.39 (0.024) & 0.35 (0.178) & 0.1 (0.023) & 0.67 (0.029) & 0.63 (0.029) \\ [0.5ex] \hline 
3 & 0.2 (0.027) & 0.3 (0.022) & 0.32 (0.026) & 0.3 (0.162) & 0.07 (0.028) & 0.66 (0.021) & 0.58 (0.054) \\ [0.5ex] \hline 
4 & 0.17 (0.024) & 0.26 (0.022) & 0.27 (0.024) & 0.27 (0.148) & 0.05 (0.01) & 0.66 (0.02) & 0.55 (0.04) \\ [0.5ex] \hline 
5 & 0.18 (0.024) & 0.23 (0.017) & 0.24 (0.027) & 0.25 (0.13) & 0.07 (0.034) & 0.65 (0.019) & 0.55 (0.027) \\ [0.5ex] \hline 
6 & 0.19 (0.03) & 0.21 (0.022) & 0.21 (0.028) & 0.23 (0.133) & 0.07 (0.031) & 0.64 (0.02) & 0.52 (0.02) \\ [0.5ex] \hline 
7 & 0.17 (0.035) & 0.19 (0.021) & 0.19 (0.027) & 0.19 (0.141) & 0.06 (0.045) & 0.63 (0.029) & 0.51 (0.019) \\ [0.5ex] \hline 
8 & 0.16 (0.02) & 0.18 (0.02) & 0.18 (0.026) & 0.18 (0.135) & 0.04 (0.014) & 0.63 (0.022) & 0.49 (0.026) \\ [0.5ex] \hline 
9 & 0.15 (0.021) & 0.17 (0.018) & 0.18 (0.022) & 0.15 (0.134) & 0.04 (0.018) & 0.63 (0.021) & 0.47 (0.033) \\ [0.5ex] \hline 
10 & 0.17 (0.035) & 0.16 (0.017) & 0.17 (0.017) & 0.14 (0.118) & 0.06 (0.039) & 0.62 (0.019) & 0.45 (0.036) \\ [0.5ex] \hline 
11 & 0.15 (0.018) & 0.16 (0.017) & 0.16 (0.018) & 0.13 (0.117) & 0.04 (0.019) & 0.62 (0.023) & 0.44 (0.043) \\ [0.5ex] \hline 
12 & 0.15 (0.03) & 0.16 (0.017) & 0.16 (0.015) & 0.12 (0.109) & 0.05 (0.033) & 0.62 (0.022) & 0.43 (0.035) \\ [0.5ex] \hline 
13 & 0.15 (0.025) & 0.15 (0.016) & 0.16 (0.016) & 0.13 (0.094) & 0.05 (0.025) & 0.62 (0.016) & 0.43 (0.02) \\ [0.5ex] \hline 
14 & 0.14 (0.017) & 0.15 (0.011) & 0.15 (0.014) & 0.11 (0.096) & 0.04 (0.022) & 0.62 (0.014) & 0.42 (0.028) \\ [0.5ex] \hline 
15 & 0.13 (0.012) & 0.15 (0.01) & 0.15 (0.014) & 0.11 (0.094) & 0.03 (0.012) & 0.62 (0.012) & 0.4 (0.02) \\ [0.5ex] \hline 
16 & 0.14 (0.011) & 0.15 (0.013) & 0.15 (0.015) & 0.09 (0.094) & 0.03 (0.005) & 0.63 (0.013) & 0.39 (0.037) \\ [0.5ex] \hline 
17 & 0.14 (0.019) & 0.15 (0.013) & 0.15 (0.013) & 0.07 (0.084) & 0.04 (0.021) & 0.63 (0.017) & 0.39 (0.011) \\ [0.5ex] \hline 
18 & 0.14 (0.012) & 0.15 (0.012) & 0.15 (0.012) & 0.06 (0.081) & 0.03 (0.014) & 0.63 (0.013) & 0.37 (0.01) \\ [0.5ex] \hline 
19 & 0.14 (0.019) & 0.15 (0.009) & 0.14 (0.012) & 0.06 (0.075) & 0.03 (0.012) & 0.63 (0.007) & 0.36 (0.009) \\ [0.5ex] \hline 
20 & 0.13 (0.011) & 0.15 (0.01) & 0.14 (0.011) & 0.06 (0.072) & 0.03 (0.015) & 0.63 (0.007) & 0.35 (0.011) \\ [0.5ex] \hline 
21 & 0.13 (0.005) & 0.15 (0.01) & 0.15 (0.01) & 0.06 (0.071) & 0.02 (0.003) & 0.63 (0.008) & 0.35 (0.012) \\ [0.5ex] \hline 
\end{tabular}
\caption{The average performance on IIRC-CIFAR after each task using the precision-weighted Jaccard Similarity. This table represents the same results as in Figure~\ref{fig:cifar_incremental_mjaccard} with the standard deviation between brackets}
\end{table}

\begin{table}[!htb]
\centering
\begin{tabular}{|| c | c | c | c | c ?c ||} 
 \hline
 task & \multicolumn{5}{c ||}{model} \\ [0.5ex] \hline
  & ER & iCaRL-CNN & iCaRL-norm & LUCIR & incremental joint \\ [0.5ex] \hline \hline 
  0 & 0.7 (0.027) & 0.78 (0.018) & 0.8 (0.019) & 0.76 (0.025) & 0.73 (0.02) \\ [0.5ex] \hline 
1 & 0.13 (0.022) & 0.46 (0.039) & 0.49 (0.034) & 0.17 (0.044) & 0.73 (0.026) \\ [0.5ex] \hline 
2 & 0.12 (0.071) & 0.34 (0.047) & 0.38 (0.041) & 0.15 (0.048) & 0.73 (0.019) \\ [0.5ex] \hline 
3 & 0.08 (0.01) & 0.27 (0.035) & 0.31 (0.024) & 0.14 (0.045) & 0.73 (0.012) \\ [0.5ex] \hline 
4 & 0.08 (0.01) & 0.23 (0.022) & 0.27 (0.015) & 0.1 (0.068) & 0.73 (0.015) \\ [0.5ex] \hline 
5 & 0.07 (0.012) & 0.2 (0.018) & 0.25 (0.014) & 0.1 (0.063) & 0.73 (0.01) \\ [0.5ex] \hline 
6 & 0.07 (0.017) & 0.18 (0.018) & 0.23 (0.017) & 0.06 (0.062) & 0.73 (0.01) \\ [0.5ex] \hline 
7 & 0.06 (0.004) & 0.17 (0.013) & 0.22 (0.013) & 0.04 (0.057) & 0.73 (0.013) \\ [0.5ex] \hline 
8 & 0.07 (0.013) & 0.16 (0.01) & 0.21 (0.01) & 0.03 (0.056) & 0.72 (0.015) \\ [0.5ex] \hline 
9 & 0.06 (0.002) & 0.16 (0.011) & 0.2 (0.01) & 0.03 (0.053) & 0.72 (0.017) \\ [1.0ex] \hline 
\end{tabular}
\caption{The average performance on IIRC-ImageNet-lite after each task using the precision-weighted Jaccard Similarity. This table represents the same results as in Figure~\ref{fig:imagenet_lite_incremental} with the standard deviation between brackets}
\end{table}

\begin{table}[!htb]
\centering
\begin{tabular}{|| c | c | c | c | c ?c ||} 
 \hline
 task & \multicolumn{5}{c ||}{model} \\ [0.5ex] \hline
  & ER & iCaRL-CNN & iCaRL-norm & LUCIR & joint \\ [0.5ex] \hline \hline 0 & 0.7 (0.024) & 0.78 (0.009) & 0.8 (0.009) & 0.75 (0.017) & - \\ [0.5ex] \hline 
1 & 0.13 (0.023) & 0.47 (0.016) & 0.51 (0.014) & 0.15 (0.044) & - \\ [0.5ex] \hline 
2 & 0.1 (0.024) & 0.34 (0.026) & 0.39 (0.02) & 0.14 (0.046) & - \\ [0.5ex] \hline 
3 & 0.08 (0.008) & 0.27 (0.02) & 0.32 (0.014) & 0.13 (0.044) & - \\ [0.5ex] \hline 
4 & 0.08 (0.021) & 0.23 (0.014) & 0.28 (0.008) & 0.12 (0.061) & - \\ [0.5ex] \hline 
5 & 0.1 (0.064) & 0.21 (0.025) & 0.26 (0.011) & 0.11 (0.06) & - \\ [0.5ex] \hline 
6 & 0.07 (0.008) & 0.19 (0.015) & 0.23 (0.011) & 0.1 (0.056) & - \\ [0.5ex] \hline 
7 & 0.06 (0.007) & 0.17 (0.017) & 0.22 (0.013) & 0.11 (0.051) & - \\ [0.5ex] \hline 
8 & 0.06 (0.006) & 0.16 (0.016) & 0.21 (0.011) & 0.1 (0.045) & - \\ [0.5ex] \hline 
9 & 0.06 (0.005) & 0.15 (0.015) & 0.2 (0.009) & 0.09 (0.06) & - \\ [0.5ex] \hline 
10 & 0.06 (0.015) & 0.15 (0.019) & 0.19 (0.014) & 0.08 (0.054) & - \\ [0.5ex] \hline 
11 & 0.06 (0.005) & 0.14 (0.017) & 0.19 (0.012) & 0.08 (0.05) & - \\ [0.5ex] \hline 
12 & 0.06 (0.011) & 0.14 (0.012) & 0.19 (0.008) & 0.08 (0.047) & - \\ [0.5ex] \hline 
13 & 0.06 (0.006) & 0.13 (0.013) & 0.19 (0.005) & 0.07 (0.045) & - \\ [0.5ex] \hline 
14 & 0.06 (0.007) & 0.13 (0.011) & 0.18 (0.006) & 0.07 (0.042) & - \\ [0.5ex] \hline 
15 & 0.05 (0.001) & 0.13 (0.015) & 0.18 (0.005) & 0.07 (0.04) & - \\ [0.5ex] \hline 
16 & 0.05 (0.003) & 0.12 (0.018) & 0.18 (0.007) & 0.06 (0.036) & - \\ [0.5ex] \hline 
17 & 0.05 (0.008) & 0.12 (0.017) & 0.17 (0.005) & 0.06 (0.034) & - \\ [0.5ex] \hline 
18 & 0.05 (0.01) & 0.12 (0.019) & 0.17 (0.008) & 0.05 (0.031) & - \\ [0.5ex] \hline 
19 & 0.05 (0.003) & 0.11 (0.021) & 0.17 (0.009) & 0.05 (0.03) & - \\ [0.5ex] \hline 
20 & 0.05 (0.004) & 0.11 (0.022) & 0.16 (0.009) & 0.05 (0.028) & - \\ [0.5ex] \hline 
21 & 0.04 (0.004) & 0.11 (0.024) & 0.16 (0.01) & 0.04 (0.025) & - \\ [0.5ex] \hline 
22 & 0.04 (0.004) & 0.1 (0.023) & 0.16 (0.009) & 0.04 (0.024) & - \\ [0.5ex] \hline 
23 & 0.04 (0.008) & 0.1 (0.019) & 0.15 (0.008) & 0.04 (0.022) & - \\ [0.5ex] \hline 
24 & 0.04 (0.001) & 0.1 (0.018) & 0.15 (0.007) & 0.03 (0.023) & - \\ [0.5ex] \hline 
25 & 0.03 (0.003) & 0.1 (0.017) & 0.14 (0.007) & 0.02 (0.022) & - \\ [0.5ex] \hline 
26 & 0.04 (0.004) & 0.09 (0.018) & 0.13 (0.007) & 0.02 (0.021) & - \\ [0.5ex] \hline 
27 & 0.03 (0.002) & 0.09 (0.018) & 0.12 (0.004) & 0.02 (0.019) & - \\ [0.5ex] \hline 
28 & 0.03 (0.002) & 0.08 (0.016) & 0.12 (0.003) & 0.02 (0.018) & - \\ [0.5ex] \hline 
29 & 0.03 (0.002) & 0.08 (0.016) & 0.11 (0.003) & 0.02 (0.017) & - \\ [0.5ex] \hline 
30 & 0.03 (0.005) & 0.08 (0.014) & 0.11 (0.003) & 0.02 (0.016) & - \\ [0.5ex] \hline 
31 & 0.02 (0.006) & 0.08 (0.015) & 0.1 (0.01) & 0.02 (0.015) & - \\ [0.5ex] \hline 
32 & 0.02 (0.01) & 0.08 (0.016) & 0.09 (0.007) & 0.02 (0.014) & - \\ [0.5ex] \hline 
33 & 0.02 (0.005) & 0.08 (0.017) & 0.09 (0.011) & 0.02 (0.012) & - \\ [0.5ex] \hline 
34 & 0.01 (0.001) & 0.07 (0.016) & 0.08 (0.01) & 0.01 (0.015) & 0.42 (0.018) \\ [1.0ex] \hline 
\end{tabular}
\caption{The average performance on IIRC-ImageNet-full after each task using the precision-weighted Jaccard Similarity. This table represents the same results as in Figure~\ref{fig:imagenet_full_incremental} with the standard deviation between brackets}
\end{table}

\end{document}